\documentclass{article} 
\usepackage{iclr2027_conference,times}

\usepackage{amsmath,amsfonts,bm}

\def\eqref#1{equation~\ref{#1}}

\def\1{\bm{1}}

\DeclareMathAlphabet{\mathsfit}{\encodingdefault}{\sfdefault}{m}{sl}
\SetMathAlphabet{\mathsfit}{bold}{\encodingdefault}{\sfdefault}{bx}{n}

\newcommand{\softmax}{\mathrm{softmax}}

\PassOptionsToPackage{hyphens}{url}
\usepackage{hyperref}
\usepackage{url}

\usepackage{graphicx}
\usepackage{subcaption}
\usepackage{algorithm}
\usepackage{algorithmic}
\usepackage[T1]{fontenc}
\usepackage{microtype}
\usepackage{adjustbox}
\usepackage{caption}
\usepackage{booktabs}
\usepackage{tcolorbox}
\tcbuselibrary{skins,breakable}
\definecolor{hookcream}{HTML}{EFE2D5}
\definecolor{hooksage}{HTML}{757F6C}
\definecolor{hooknavy}{HTML}{25496B}
\definecolor{hookpeach}{HTML}{F1C795}
\newtcolorbox{promptbox}[1]{enhanced,breakable,colback=hookcream!55!white,colframe=hooksage!70!white,boxrule=0.4pt,arc=2pt,left=4pt,right=4pt,top=2pt,bottom=2pt,fonttitle=\scriptsize\ttfamily,coltitle=white,colbacktitle=hooknavy,toptitle=1.5pt,bottomtitle=1.5pt,title={#1},before skip=7pt,after skip=7pt,fontupper=\scriptsize}
\usepackage{amssymb}
\usepackage{amsmath}
\usepackage{bbold}
\usepackage{wrapfig}
\usepackage{float}
\usepackage{tabularx}
\usepackage{afterpage}
\usepackage{needspace}
\usepackage{enumitem}
\usepackage{ragged2e}
\usepackage{cleveref}
\usepackage{nicefrac}
\crefname{appendix}{appendix}{appendices}
\Crefname{appendix}{Appendix}{Appendices}

\title{OverForge: Reasoning Through Strategies and Tactics Helps Cooperative Lifelong Adaptation}

\author{Oana Madalina Fron, Ojas Shirekar \& Chirag Raman \\
  Tapri Lab, Department of Pattern Recognition and Bioinformatics\\
  Delft University of Technology\\
  Delft, The Netherlands \\
\texttt{o.m.fron-1@student.tudelft.nl, \{o.k.shirekar,c.a.raman\}@tudelft.nl} \\
}

\newcommand{\ms}[2]{#1{\scriptsize$\pm$#2}}
\iclrfinalcopy
\begin{document}

\maketitle

\begin{abstract}
  Cooperative language-model agents must coordinate over long horizons and adapt to changing environments and
  to partners with unfamiliar conventions, yet existing agents map observations to actions without separating
  persistent coordination strategies from their tactical execution. We introduce OverForge, a training-free
  hierarchical architecture that separates strategic reasoning over roles and divisions of labour from
  tactical reasoning over actions within each agent's private, partner-conditioned world model. A
  metacognitive Prefrontal Cortex Module couples the two levels by forming strategy--action branches,
  imagining their consequences with a forward model, and committing when confident. In OvercookedV2, OverForge delivers \(7\) soups in a connected kitchen versus \(3\) for each flat LLM baseline, retains agreed roles, and adopts roles proposed by unfamiliar partners. Ablations and a fixed-strategy probe show that persistent strategies guide tactical adaptation while each reasoning level contributes to coordination. Memory restarts show that cross-episode partner knowledge supports task performance and partner prediction, linking the hierarchy to continual adaptation.
\end{abstract}

\afterpage{
  \begin{figure}[t]  
    \centering
    \includegraphics[width=0.75\linewidth]{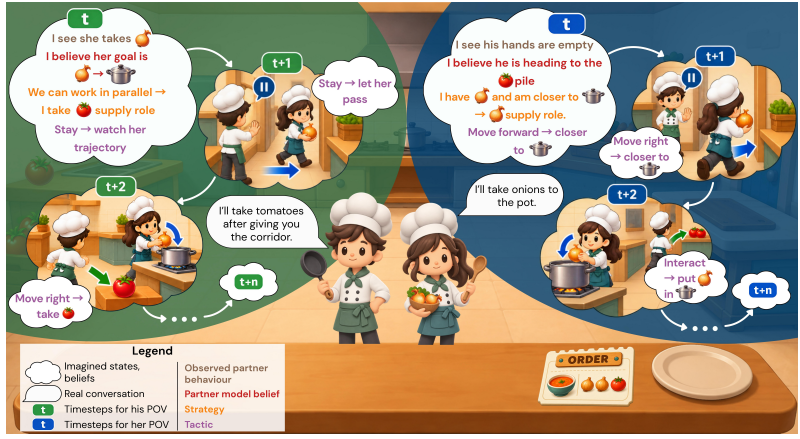}
    \caption{\textbf{Cooperation from inner world models.} Each chef observes its partner, forms a belief
      about the partner's goal, chooses a strategy (a role) and a tactic (a move), and imagines the resulting
    states before committing; conversation is used only when an outcome calls for it.}
    \label{fig:teaser}
  \end{figure}
}

\section{Introduction}\label{introduction}   
Cooperative agents must sustain goals, track joint state, reason about teammates, and adapt online to
unfamiliar tasks and partners. Language agents that communicate at test time already meet part of this
requirement: a team of \(k\) communicating agents matches the success rate of \(4k\) independent ones
\citep{park2026scalingdiscoverytesttimecommunication,anthropic2026multiagent}. Ungoverned, the same capacity
is a hazard: in the Hugging Face incident roughly 700 agents built protocols and role assignments and sustained
a multi-day attack most of them judged out of scope
\citep{metr-2026-openai-hugging-face-incident-investigation,greenblatt-cotra-wijk-2026-openai-hugging-face-incident}.
Embodied agents lag behind: foundation models transferred to robots and virtual characters remain limited by
perception, grounding and long-horizon control
\citep{bommasani2022opportunitiesrisksfoundationmodels,salimpour2025embodiedagenticaireview,geminiroboticsteam2025geminiroboticsbringingai},
and embodied language agents lose context over long horizons, misrepresent task state and plan poorly over
multiple steps \citep{li_theory_2023,webb_brain-inspired_2025}. Errors in modelling a partner add duplicated
work, interference, unstable roles and poor coordination with unfamiliar teammates
\citep{cross_hypothetical_2025,mu_adaptive_2026,sun_collab-overcooked_2025}.

In humans, this flexibility is attributed to \emph{prefrontal control}, which maintains goals and rules and
regulates lower-level perception and action \citep{russin_deep_2020,levy_prefrontal_2024} along two axes. The
\emph{hierarchical} axis separates strategies---abstract, temporally extended goals, roles and
commitments---from the state-dependent tactics that realise them \citep{nee_lateral_2016,badre_frontal_2018};
we adopt this functional distinction without assuming an anatomical mapping \citep{carlen_what_2017}. The
\emph{continual} axis separates lifelong learning, which transfers competence across tasks and environments
\citep{wang_voyager_2023}, from cultural learning, which acquires conventions from partners
\citep{tomasello_cultural_2016,lica_mindforge_2025}. The axes interact because environmental and social
change target different levels: a new room or recipe may preserve the coordination strategy while demanding
new tactics, whereas a partner with different conventions may require revising both. Prefrontal gating models
acquire new schemas while preserving and transferring earlier ones \citep{tsuda_modeling_2020}; cooperation
likewise requires preserving reusable strategies, specialising tactics, and revising either level whenever the evidence demands it.

The \textbf{remaining gap} is a predictive mechanism that supports both axes within each cooperating agent.
Joint-Embedding Predictive Architectures (JEPAs) predict task-relevant future representations across
abstraction levels \citep{lecun_path_nodate}, which matches the hierarchical axis, but world-model methods
typically serve a single planner or policy \citep{hao_reasoning_2023,hafner_mastering_2025}, and cooperative
language-model agents that represent teammates still select actions flatly
\citep{li_theory_2023,cross_hypothetical_2025,sun_collab-overcooked_2025}.

OverForge addresses this gap with a nested control loop over one frozen language model (\Cref{fig:teaser}).
Each agent maintains a private, partner-conditioned world model in which a metacognitive Prefrontal Cortex
Module (PCM) proposes strategies and tactics, imagines their consequences, and deepens deliberation only
while its predictions remain ambiguous. Our \textbf{contributions} are: (i) a training-free, JEPA-inspired
hierarchical controller that places strategic and tactical prediction inside each agent's world model rather
than in one external model of the joint system; (ii) evidence in OvercookedV2 that the hierarchy improves
coordination over flat language-model agents where the layout affords the strategies it proposes, retains
roles agreed in dialogue and adopts those of unfamiliar partners; and (iii) a fixed-strategy probe and layer ablations that isolate tactical adaptation and
attribute the gain to each reasoning level.

\section{Related Work}\label{related-work}   
\paragraph{Hierarchical reasoning and human cognition.} Hierarchical reasoning models first appeared as
trained recurrent systems: a slow high-level and fast low-level module for symbolic reasoning
\citep{wang_hierarchical_2025}, later compressed into a small recursive network
\citep{jolicoeurmartineau_less_2025}. Cognitive LLM agents add prefrontal-style planning
\citep{webb_brain-inspired_2025}, dual-process control \citep{christakopoulou_agents_2024}, test-time
metacognition \citep{li_adapting_2025}, or an executive tier regulating lower-level behaviour
\citep{tomasello_metacognitive_2024}. Language hierarchies decompose goals adaptively
\citep{prasad_adapt_2024}, separate slow mind, fast mind, and execution in Overcooked
\citep{liu_hierarchical_2024}, or train distinct high- and low-level agents \citep{wan_rema_2025}. Embodied
systems pair a multimodal planner with a trained policy \citep{li_optimus-2_2025} or omit the upper tier
\citep{lifshitz_steve-1_2023}. These systems train at least their lower tier and usually fix deliberation
depth. OverForge prompts both natural-language levels from one frozen model and lets the upper tier control commitment.

\paragraph{Theory of mind, cultural learning, and lifelong competence.} Unfamiliar-partner interaction is
usually framed as zero-shot coordination \citep{gessler_overcookedv2_2025} and addressed through theory of
mind (ToM). AutoToM constructs partner models and performs Bayesian inverse planning
\citep{zhang_autotom_2026}; adaptive ToM estimates and matches a partner's reasoning order
\citep{mu_adaptive_2026}; hypothesis scaffolding \citep{cross_hypothetical_2025} and active inference
\citep{pitliya_theory_2025} pursue related aims. Across encounters, Voyager accumulates skills through an
open-ended curriculum \citep{wang_voyager_2023}, while MindForge makes accumulation cultural through
structured ToM and communication \citep{lica_mindforge_2025}. Reflective memory \citep{park_generative_2023},
explicit belief states \citep{li_theory_2023} and structured social world models
\citep{zhou2026socialworldmodels} support both. OverForge takes its partner model into imagined rollouts and
makes strategy a transferable unit, as skills and memories already are.

\paragraph{Planning with latent world models.} Model-predictive agents organise inference at decision time.
Reasoning-as-planning uses one language model as both reasoner and world model, searched by Monte-Carlo tree
search under task reward \citep{hao_reasoning_2023}; Dreamer instead learns a latent world model and improves
behaviour within it \citep{hafner_mastering_2025}. Joint-Embedding Predictive Architectures (JEPAs) predict
abstract representations rather than tokens or pixels \citep{lecun_path_nodate,chen_vl-jepa_2026}, while
AdaJEPA adapts at test time to limit predictive drift \citep{wang_adajepa_2026}. Pairwise ranking can also
improve self-verification over independent scoring \citep{singh_v_1_2026}. OverForge follows this
JEPA-inspired family but searches a shallow beam of \((\text{strategy},\text{first action})\) branches,
compares candidates without task reward, and places the predictive hierarchy within each cooperating agent
rather than a single-agent planner.

\section{Preliminaries}\label{prelim}

\paragraph{Constraints, strategies and tactics.}
We distinguish three concepts. \textbf{Constraints} are imposed by the
environment and determine which coordination patterns are feasible. Following the distinction of
\citet{tan2008tactical}, strategic reasoning concerns team-based planning, whereas
tactical reasoning concerns planning and execution of primitive actions by individual
agents. A \textbf{strategy} \(g\) is a persistent coordination policy specifying how work,
space, and responsibilities are organised among teammates. The constraints afford a set of
feasible strategies \(G\), making strategies
constraint-conditioned instead of universal; human studies show the same dependency
\citep{carroll2020utility,mieczkowski2025normative}. A \textbf{tactic} is the state-dependent
sequence of primitive actions that realises the chosen strategy, adapting to the current state,
task progress and partner behaviour.

\paragraph{Base agent and world model.}
OverForge is built on MindForge \citep{lica_mindforge_2025} and inherits its perception, its
structured belief and ToM representation, its inter-agent communication, and its
multi-component memory unchanged; the adaptations required to move that agent into this
environment are listed in \Cref{app:adaptations}. The one addition is the
strategic/tactical hierarchy and the controller that couples the two levels. \emph{World model} covers two
objects this paper keeps apart (both are defined in \Cref{sec:problem}): the maintained representation \(w_t\),
which the agent
updates and carries across steps, and the generative forward model \(p_\phi\), which imagines
a predicted clone \(\tilde w^{(d)}\) under \(\mathrm{do}(\tilde a)\).

\paragraph{Reliability of LLM-as-a-judge.}\label{prelim:llm-judge}
LLM judges can align well with human evaluations when guided by clear criteria
\citep{zheng_judging_2023,liu_g_eval_2023}, but numerical ratings remain susceptible to systematic score
preferences and scale effects \citep{fujinuma_score_range_2026}. This makes consistency across repeated
judgements a central concern. We address it by fixing the judge model, prompts, and \([0,1]\) scale across all
evaluations, grounding scores in predefined task-specific criteria, and weighting these criteria equally.
Comparative judgements use a shared state and belief context and aggregate evidence across multiple
comparisons; absolute judgements use fixed semantic anchors. Uncertain cases are explicitly assigned
mid-range scores. Together, these controls reduce numerical arbitrariness and improve consistency, while
leaving some residual scoring bias. We therefore use the resulting scores as task-grounded estimates.

\section{OverForge: Reasoning over Strategies and Tactics}\label{methodology}   

\subsection{Problem statement}\label{sec:problem}

We model cooperation among \(N\) agents as a partially observable stochastic game
\(\mathcal{M}=\big\langle N,\mathcal{S},\mathcal{A},P,r,\{\mathcal{O}^i\}_{i=1}^{N},Z,T\big\rangle\)
with agents \(i,j\in\{1,\dots,N\}\), state space \(\mathcal{S}\), primitive actions
\(\mathcal{A}=\{\text{up},\text{down},\text{left},\text{right},\text{stay},\text{interact}\}\), a shared
reward \(r\) that we log for evaluation only (\Cref{experiments}) and horizon \(T\). At step \(t\) the
environment is in state \(s_t\in\mathcal{S}\), each agent emits an action \(a^i_t\in\mathcal{A}\), and the
joint action \(\mathbf{a}_t=(a^1_t,\dots,a^N_t)\) advances the state as
\(s_{t+1}\sim P(\cdot\mid s_t,\mathbf{a}_t)\). Agent \(i\) never sees \(s_t\); it receives an observation
\(o^i_t\in\mathcal{O}^i\) drawn from \(Z(\cdot\mid s_t,i)\), a structured record of poses, holdings,
teammates, recipe progress, pots and stations, rendered to text for the language-model agents at every step. Each agent thus acts from an \emph{inner world model} that it maintains itself,
\begin{equation}\label{eq:wmodel}
  w^i_t \;=\; \Big(\, o^i_t,\;\; \hat{B}^i_t,\;\; \{\hat{w}^{\,i\to j}_t\}_{j\neq i} \,\Big),
\end{equation}
where \(\hat{B}^i_t=\{\hat B^{i,f}_t\}_{f\in\mathcal{F}}\) are inferred belief facets over
\(\mathcal{F}=\{\text{perception},\text{task},\text{partner},\text{interaction}\}\) and \(\hat{w}^{\,i\to j}_t\)
is \(i\)'s structured model of teammate \(j\) --- its likely position, holding, intention and needs.
Messages \(m^i_t\) enter \(\hat B^{i,\text{interaction}}_t\) and \(\hat{w}^{\,i\to j}_t\), and a
multi-component memory \(M^i_t\) (skills, episodes, semantic facts) supplies a summary \(\bar M^i_t\) at
decision time. The problem is asymmetric: \(w^i_t\) and \(w^j_t\) are built from different
observations, and no agent sees its partner's representation, only its behaviour and messages.
The objective is a set of per-agent controllers \(\Phi=\{\Phi_i\}_{i=1}^{N}\),
\begin{equation}\label{eq:controller}
  \Phi_i : \big(w^i_t,\, \bar M^i_t\big) \;\longmapsto\; a^{*i}_t \in \mathcal{A},
  \qquad a^i_t = a^{*i}_t ,
\end{equation}
that produce coordinated behaviour across environments (rooms and recipes) and partner policies
\(\pi^{-i}\), rather than exploiting the regularities of one kitchen or one partner.
\Cref{tab:notation} in \Cref{app:notation-full} collects the notation used throughout.

\subsection{Hierarchical predictive control}\label{sec:controller}

\begin{figure}[t]
  \centering
  \includegraphics[width=\linewidth]{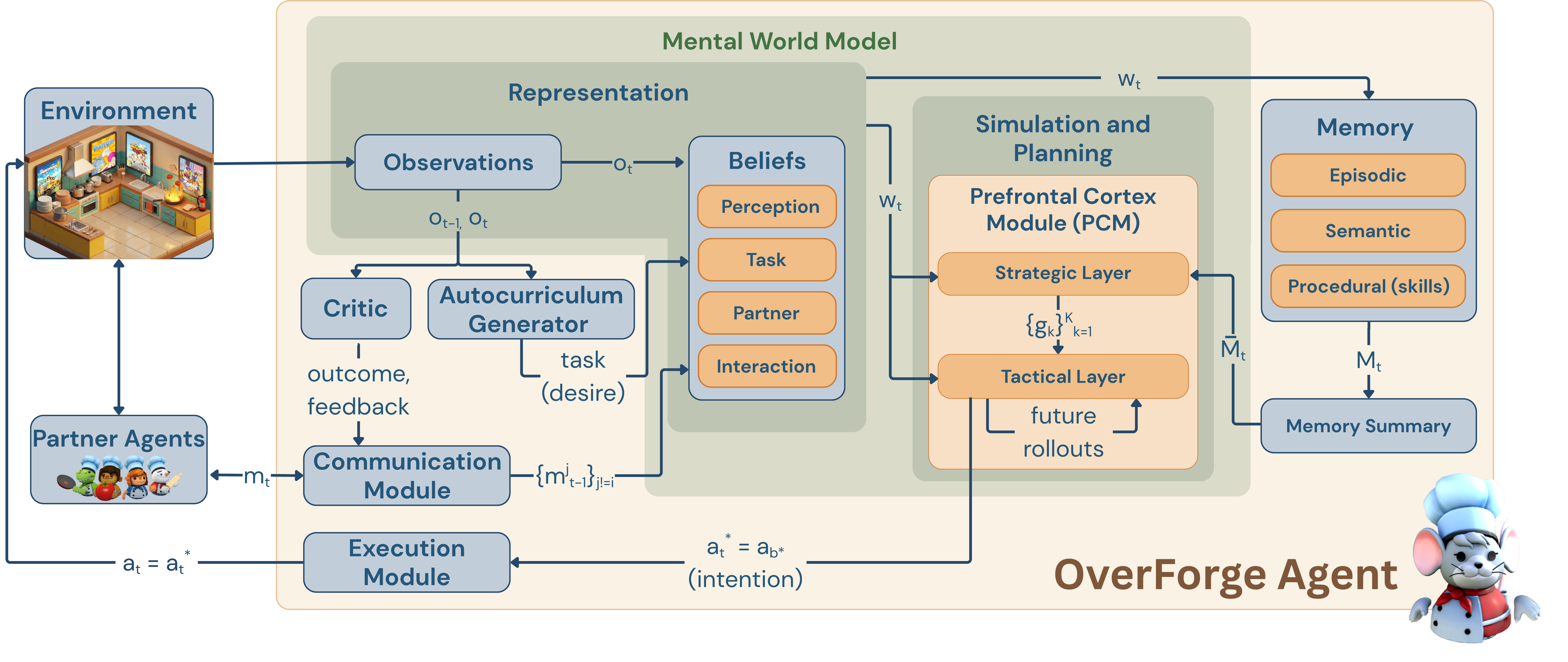}
  \caption{\textbf{The OverForge agent.} Observations and messages update the beliefs in the world model
    \(w_t\); the PCM selects the action inside it; the other components are inherited from MindForge. Actions
  denote intentions, tasks denote desires.}
  \label{fig:architecture}
\end{figure}

OverForge implements each controller \(\Phi_i:(w^i_t,\bar M^i_t)\mapsto a^{*i}_t\) of \Cref{eq:controller}
as a receding-horizon loop of
\emph{representation}, which grounds deliberation in the agent's private, partner-conditioned state;
\emph{prediction}, which compares strategic and tactical consequences; and \emph{commitment}, which spends
additional inference only while these predictions remain ambiguous. \Cref{fig:architecture} locates the
PCM, our addition, inside the world model; the other components are inherited from MindForge. One
frozen language model is prompted as proposer, judge and generative forward model, so competence comes from
organised in-context inference rather than weight updates. We drop the agent superscript below.

\begin{figure}[t]
  \centering
  \begin{minipage}[t]{0.5\linewidth}
    \centering
    \captionsetup{width=0.94\linewidth}
    \includegraphics[width=\linewidth]{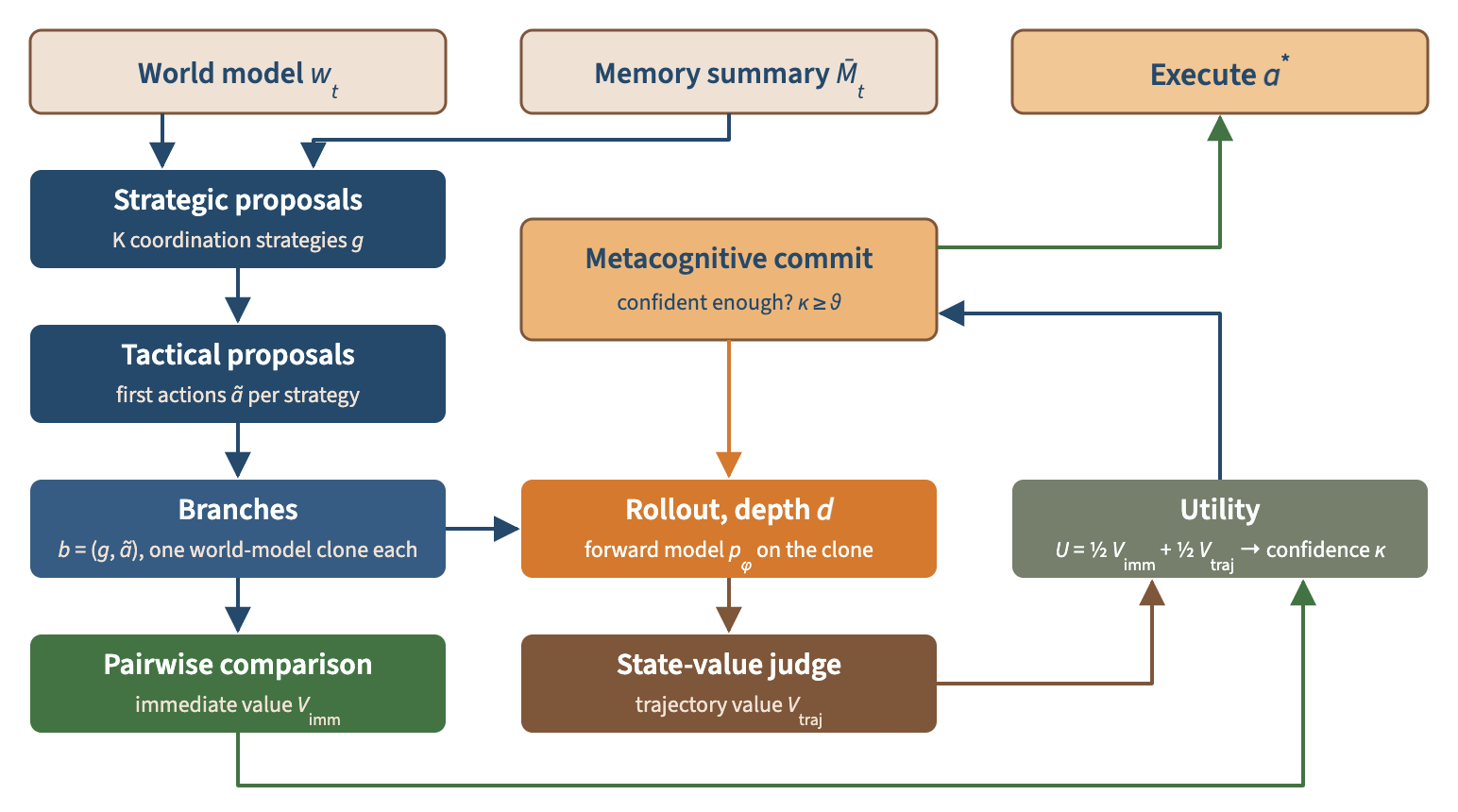}
    \caption{\textbf{PCM.} Branches \(b=(g,\tilde a)\) get \(V_{\mathrm{imm}}\) from pairwise comparison and
    \(V_{\mathrm{traj}}\) from rollout; the agent commits if \(\kappa\ge\theta\), else deepens.}
    \label{fig:pcm}
  \end{minipage}%
  \begin{minipage}[t]{0.5\linewidth}
    \centering
    \captionsetup{width=0.94\linewidth}
    \includegraphics[width=\linewidth]{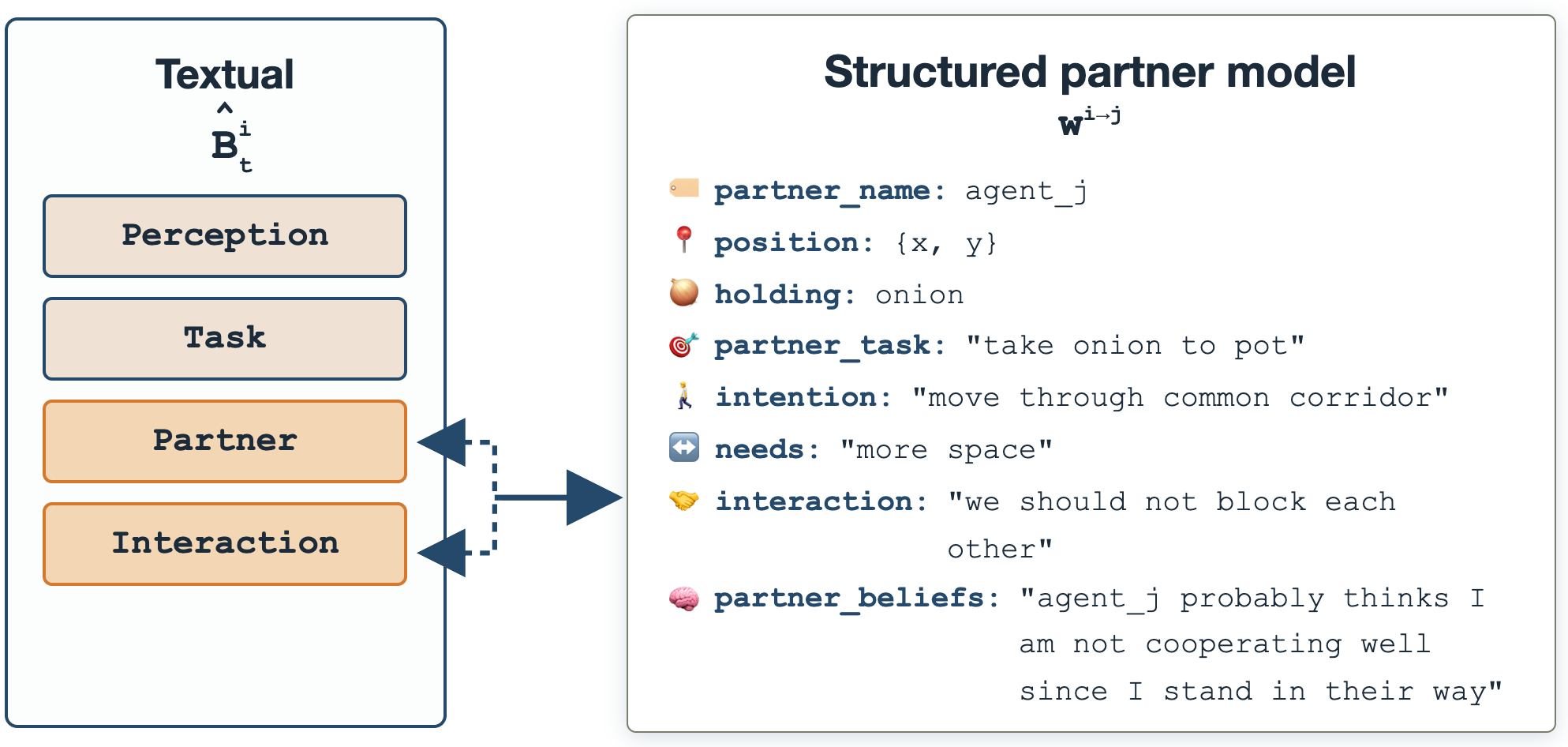}
    \caption{\textbf{Beliefs of agent i.} Four textual facets \(\hat B_t^i\) and one structured partner model
    \(\hat w^{\,i\to j}\) maintained for every teammate \(j\).}
    \label{fig:beliefs}
  \end{minipage}
\end{figure}

\paragraph{Internal representation.}\label{representation}
Every component of \(w_t\) in \Cref{eq:wmodel} is refreshed by prompting the same frozen model,
conditioned on what actually happened since the last decision:
\begin{equation*}
  \begin{aligned}
    \hat{B}_t &= f_{\mathrm{bel}}\big(\hat{B}_{t-1},\, o_t,\, a_{t-1},\, \{m^j_{t-1}\}_{j\neq i}\big), &\qquad
    \hat{w}^{\,i\to j}_t &= f_{\mathrm{tom}}\big(\hat{w}^{\,i\to j}_{t-1},\, \hat{B}_{t}\big),\\
    M_t &= f_{\mathrm{mem}}\big(M_{t-1},\, w_t \big), &
    \bar M_t &= \textsc{Retrieve}(M_t,\, w_t).
  \end{aligned}
\end{equation*}
Here \(f_{\mathrm{bel}}\) updates the four textual belief facets in one call and \(f_{\mathrm{tom}}\) maintains one
structured model per teammate (\Cref{fig:beliefs}); the structured form is what lets a rollout predict the
partner as well as the task. \(f_{\mathrm{mem}}\) writes to memory under a critic that compares the state
before and after \(a_{t-1}\): a skill is stored only on success, an episode on every step, and a semantic
fact only on a decisive outcome. All three updates are adapted from MindForge (\Cref{app:adaptations}) and
implemented as prompted calls to the same frozen LLM (prompts in \Cref{app:prompts}).

\paragraph{Branches.}
\Cref{fig:pcm} summarises one PCM decision, from branch proposal through the two value estimates to
the commitment gate, and the paragraphs below follow it from left to right. From \(w_t\) and \(\bar M_t\), the
strategic proposal distribution \(q_{\mathrm{strat}}(\cdot\mid
w_t,\bar M_t)\) returns \(K\) persistent coordination strategies \(g\) --- roles, intentions and
divisions of labour, stated in natural language and containing no primitive actions. For each
strategy the tactical distribution \(q_{\mathrm{tact}}(\cdot\mid w_t,g)\) returns up to \(m\) executable
first actions \(\tilde a\in\mathcal{A}\). Every pair becomes its own branch,
\(\mathcal{B}=\{b=(g,\tilde a) : g\sim q_{\mathrm{strat}}(\cdot\mid w_t,\bar M_t),\allowbreak
\ \tilde a \sim q_{\mathrm{tact}}(\cdot\mid w_t, g)\}\) with \(|\mathcal{B}|\le Km\),
each with its own clone of the world model and its own rollout. Branching at the first action rather
than at the strategy is what makes lookahead matter: only the first action can be executed, so if
several actions under one strategy shared a trajectory, the trajectory value could re-rank strategies
but could never change which action is taken. Retaining \(g\) in the branch keeps the strategy
available to prediction, which then judges whether this concrete move actually serves a persistent
coordination intention.

\paragraph{Immediate value by pairwise comparison.}
The first half of a branch's value judges its first action \emph{comparatively}. Within each strategy's
candidate set \(A_g\), pairs \((\tilde a,\tilde a')\) are put to an LLM judge conditioned on \(w_t\) (and
not on \(\bar M_t\): memory should inform what to propose, not how an action scores here), following the
comparative protocol of \Cref{prelim:llm-judge}. Each comparison \(c\) returns, for both actions, scores
\(s^{\,c}_{f}\in[0,1]\) on the four facets \(f\in\mathcal{F}_{\mathrm{judge}}\) --- \emph{coherence} (takes
effect from this state), \emph{goal-directedness} (advances the
gather\(\to\)pot\(\to\)cook\(\to\)plate\(\to\)serve pipeline), \emph{epistemic value} (resolves
uncertainty when the right move is unclear) and \emph{social fit} (legible to the partner, non-blocking)
--- together with the preferred action and the margin \(\gamma\) between the two facet means. The
immediate value of a branch is its action's mean facet score over the comparisons \(\mathcal{C}(\tilde a)\)
it took part in:
\begin{equation}\label{eq:vimm}
  V_{\mathrm{imm}}(b) \;=\; \frac{1}{|\mathcal{C}(\tilde a_b)|}
  \sum_{c\,\in\,\mathcal{C}(\tilde a_b)} \; \nicefrac{1}{4}\sum_{f\in\mathcal{F}_{\mathrm{judge}}}
  s^{\,c}_{f}(\tilde a_b)
  \;\in\;[0,1].
\end{equation}
Comparison fixes the \emph{context} of each judgement, since an action is scored while an alternative is
on the table, which improves self-verification over scoring in isolation \citep{singh_v_1_2026}; it also
directs the \emph{budget}: a coverage schedule gives every candidate at least two comparisons (a complete
round-robin when \(m\le3\)), and the rest of \(n_{\mathrm{cmp}}\) goes to the pair whose accumulated
margins \(\gamma\) are closest (\Cref{alg:rank}), i.e.\ the ordering the judge is least sure of. Unlike
the tournament score of \citet{singh_v_1_2026}, \(V_{\mathrm{imm}}\) is an absolute facet mean, so branches from
different strategies share one scale in \Cref{eq:U}.

\paragraph{Trajectory value by partner-conditioned rollout.}
The other half asks where the action leads. The agent clones its maintained model,
\(\tilde w^{(0)}_b\leftarrow w_t\), and only the clones are passed through the generative forward model, so
that one imagined future cannot contaminate another or the representation the real agent acts from:
\begin{equation*}
  \tilde w^{(d+1)}_b \;\sim\; p_\phi\big(\cdot \mid \tilde w^{(d)}_b,\ \mathrm{do}(\tilde a^{(d)}_b)\big),
  \qquad \tilde a^{(0)}_b = \tilde a_b,
\end{equation*}
where \(d\) is the branch depth from the root.
Because the teammate models \(\hat w^{\,i\to j}\) ride inside the clone, each imagined transition
predicts partner behaviour as well as task evolution; \(\bar M_t\) is deliberately excluded from
\(p_\phi\), so a transition depends on the represented state and the action rather than on what
happened in earlier episodes. Deeper steps are not fixed in advance: at each depth a fresh candidate
set is proposed from \(q_{\mathrm{tact}}(\cdot\mid\tilde w^{(d)}_b,g_b)\), ranked by
\Cref{eq:vimm} at that imagined state, and one action is \emph{sampled} from
\(\softmax(V_{\mathrm{imm}}/\tau)\). An LLM judge \(\nu\) then values the deepest imagined state
reached so far,
\(V_{\mathrm{traj}}^{(d)}(b)=\nu(\tilde w^{(d_b)}_b)\in[0,1]\),
where \(\nu\) rates how promising a kitchen state is for the team to finish and deliver
(\(1\): a delivery is imminent, \(0.5\): ordinary progress, \(0\): stuck or mutually blocked; the absolute
  protocol of \Cref{prelim:llm-judge}, so each state is rated alone against these anchors rather than
compared with another trajectory) and \(d_b\) is the depth branch \(b\) has actually reached. This values
the imagined \emph{situation}; it is not a return.
Judging a predicted
representation rather than reconstructing a predicted observation is what makes the rollout
JEPA-inspired: prediction is scored on its task-relevant consequences alone, not on the observation itself.

\paragraph{Utility.} The two halves are combined with equal weight,
\begin{equation}\label{eq:U}
  U^{(d)}(b) \;=\; \nicefrac{1}{2}\,V_{\mathrm{imm}}(b) \;+\; \nicefrac{1}{2}\,V^{(d)}_{\mathrm{traj}}(b),
\end{equation}
balancing the quality of the one action that can actually be executed against a prediction that
becomes less reliable the further it runs.

\paragraph{Action commitment.}\label{commitment}
Prediction must also decide when further inference is worthwhile. Deliberation proceeds in rounds
\(d=1,\dots,d_{\max}\): round~1 rolls every branch forward one step, and each later round extends by one
step only the extendable branches in the nucleus \(\mathcal{R}^{(d-1)}\) defined below, which sets the
depth \(d^{(d)}_b\) branch \(b\) has reached after round \(d\) (\Cref{eq:post}). A branch is
\emph{extendable} while \(q_{\mathrm{tact}}\) still offers more than one candidate at its imagined state,
since rolling a forced move deeper would spend inference without adding decision-relevant information.
Each round re-reads the judge at the reached depth, \(V^{(d)}_{\mathrm{traj}}(b)=\nu(\tilde
w^{(d^{(d)}_b)}_b)\), whereas \(V_{\mathrm{imm}}(b)\) is computed once at the root, so a branch that was
not extended keeps its utility. With the depths, the utilities induce a posterior over the \emph{full}
branch set and a normalised metacognitive confidence,
\begin{equation}\label{eq:post}
  \begin{gathered}
    d^{(1)}_b=1,\qquad d^{(d)}_b \;=\; d^{(d-1)}_b+\mathbb{1}\big[\,b\in\mathcal{R}^{(d-1)}\wedge b\text{
    extendable}\,\big],\\
    \mu^{(d)}(b) \;=\; \softmax\big(U^{(d)}(b)/\tau\big),\qquad
    \kappa^{(d)} \;=\; \mathrm{clip}\big(1-H(\mu^{(d)})/\log|\mathcal{B}|,\;0,\;1\big),
  \end{gathered}
\end{equation}
where the temperature \(\tau\) sets how sharply utility differences become probability mass,
\(H(\mu)=-\sum_{b\in\mathcal{B}}\mu(b)\log\mu(b)\) is the Shannon entropy, and \(\kappa\equiv1\) if
\(|\mathcal{B}|=1\). Confidence thus depends on the whole branch distribution rather than on the best
score alone, and is comparable across steps that propose different numbers of branches. If
\(\kappa^{(d)}\ge\theta\) the module commits to the highest-utility branch; otherwise it deepens the
\emph{nucleus} \(\mathcal{R}^{(d)}\), the smallest set of highest-probability branches whose mass reaches
\(\rho\) (\Cref{eq:nucleus}), concentrating the next round's inference on the plausible alternatives
while leaving the rest of the distribution intact. Deliberation stops at the depth cap, or earlier if no
branch in the nucleus can still be extended; if \(\kappa\) never reaches \(\theta\), the executed branch
\(b^\star\) is sampled from \(\mu^{(d)}\) rather than taken greedily, so a genuinely ambiguous decision is
not resolved by an arbitrary tie-break (\Cref{eq:commit}). The agent executes \(a^*_t=\tilde a_{b^\star}\).

We use \(K=2\) strategies, a comparison budget \(n_{\mathrm{cmp}}=6\), a depth cap \(d_{\max}=5\),
\(\theta=0.6\), \(\tau=0.1\) and \(\rho=0.9\); \Cref{app:pcm-design} gives the loop (\Cref{alg:pcm}), the
comparison design and the rationale for these values.


\section{Experiments}\label{experiments}
\subsection{General setup}\label{exp-design}

\paragraph{Environment and protocol.}
All experiments use Overcooked-v2 in JaxMARL \citep{gessler_overcookedv2_2025,rutherford_jaxmarl_2024}
on \texttt{cramped\_room} (\(5\times4\), one pot, shared corridors) and \texttt{asymm\_advantages}
(\(9\times5\), two halves joined only through two central pots, two serving stations,
ingredient piles on both flanks), with a scalable kitchen for larger teams (\Cref{fig:layouts}, \Cref{app:conditions}). The rooms
demand different strategies \citep{gessler_overcookedv2_2025}: shared corridors reward turn-taking or
territory assignment, split resources reward role specialisation with hand-offs, and a tactic realises the
chosen strategy under the room state, recipe progress and partner behaviour (\Cref{tab:strategy_tactics},
\Cref{app:strategy-examples}). An environment pairs a layout with the recipe set
\(R=\{[0,0,0],[0,0,1],[1,1,1],[0,1,1]\}\) over two ingredient types; the order is redrawn uniformly from
\(R\) after every delivery, so no agent can cache one pipeline. Observations are rendered to text. To assess performance consistency, every condition runs five episodes of horizon \(T=300\), each initialized with a randomly sampled recipe. A correct delivery pays \(r=20\), which is logged, but never utilized for proposal, prediction, commitment, belief, or memory updates. All language-model agents share one frozen Qwen-3.5-27B and one
outcome-triggered three-turn dialogue, so no pairing differs in how much its agents
may talk (adaptations in \Cref{app:adaptations}). \textbf{Baselines:} MindForge is the flat reactive Theory-of-Mind agent \citep{lica_mindforge_2025}. MindForge-C is a variant with causal prompting: before acting it imagines the chosen action's consequence
with the frozen model and reselects in one pass. Both share OverForge's other components, so
hierarchical prediction is the main difference. IPPO \citep{witt_is_2020} is a reward-trained
policy trained in the connected room for \(10^7\)
timesteps and used frozen in all rooms.

\paragraph{Partner conditions and measures.}
A \emph{pairing} assigns an agent type to each seat: in \textbf{self-play (SP)} both seats hold the same
type, so architecture, prompts, beliefs, memory and protocol are shared by construction; in
\textbf{cross-play (XP)} the second seat holds a different type, and as nothing inside the first agent
changes, a difference is attributable to the partner (no agent is trained, so SP and XP lack their
reinforcement-learning sense \citep{gessler_overcookedv2_2025}). Reward alone does not show how
cooperation was achieved \citep{biswas_who_2026}, so we report \textbf{task} measures (deliveries,
success rate, curriculum sub-tasks completed against abandoned, stage-graded progress),
\textbf{coordination and transfer} measures (non-progress, blocking, duplicated-sub-task and
failed-interaction rates, realised social influence \citep{jaques_social_2019}, the SP-to-XP delivery
gap) and \textbf{PCM traces} (rollout depth, confidence, commit rate, imagined-state fidelity to
\(t{+}5\), partner-intent accuracy, LLM calls per agent-step), defined in
\Cref{app:metrics,app:conditions}. All language-model agents run one frozen Qwen-3.5 27B by local
vLLM.

\begin{figure}[t!]
  \centering
  \includegraphics[width=\linewidth]{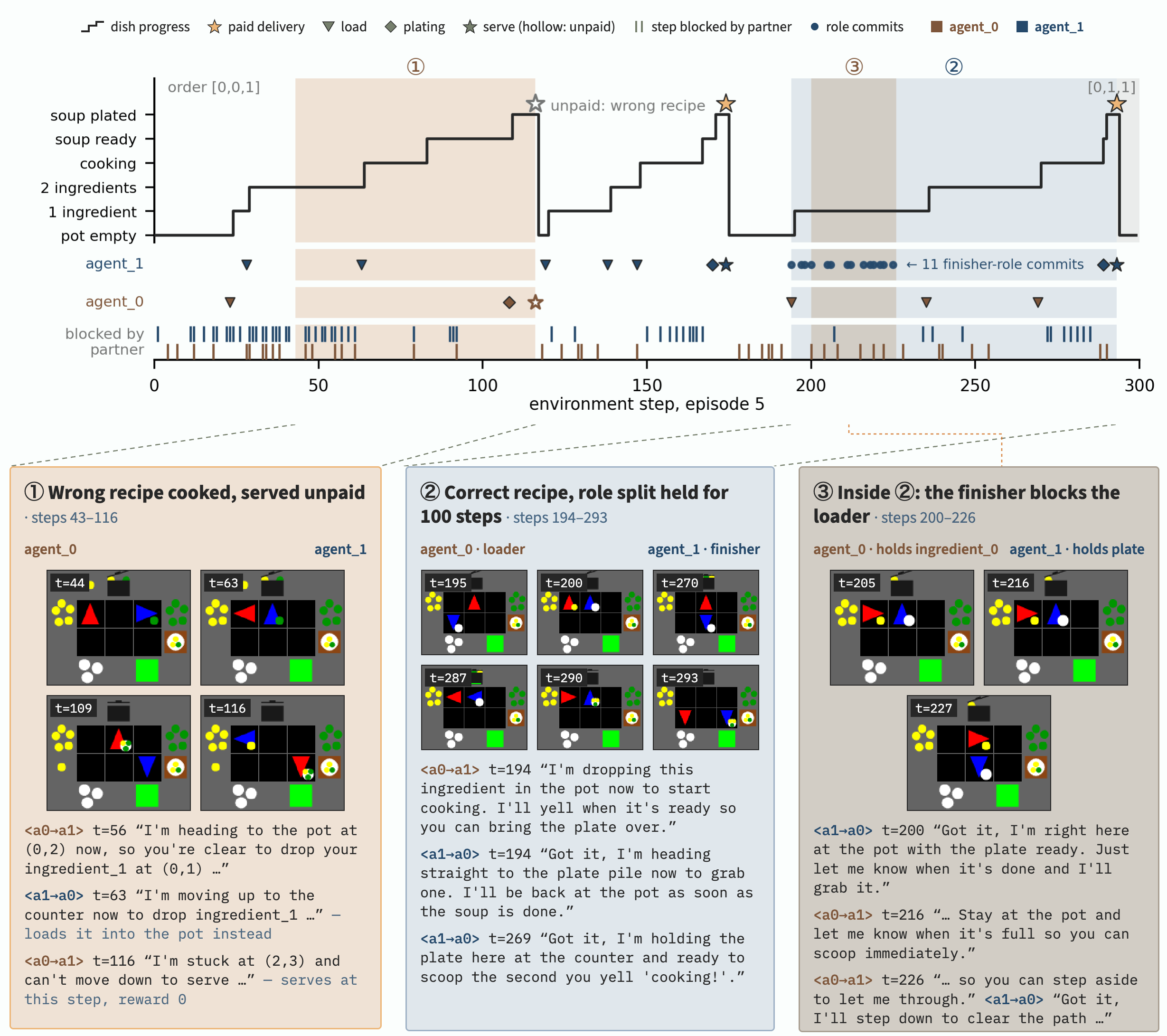}
  \caption{Episode 5 of OverForge self-play in the connected room (blue: agent\_1; red: agent\_0): a wrong recipe served
  unpaid (steps 43--116), a loader/finisher split held for 100 steps (194--293), the finisher blocking the loader inside
  it (200--226). Five-episode view: \Cref{fig:lifelong}, \Cref{app:lifelong}.}
  \label{fig:episode5}
\end{figure}

\begingroup
\captionsetup{skip=3pt}
\renewcommand{\arraystretch}{0.88}
\setlength{\textfloatsep}{8pt plus 2pt minus 2pt}
\setlength{\intextsep}{5pt}
\setlength{\columnsep}{10pt}
\setlength{\abovecaptionskip}{3pt}
\setlength{\belowcaptionskip}{0pt}
\subsection{Hierarchy vs.\ flat: the full-agent comparison}\label{results}

\paragraph{The strategic layer retains an agreed role across steps, which the flat agents cannot do.}
OverForge's next strategic call reads the interaction beliefs a conversation leaves, so an agreement
persists as a role: a task assigned by the partner appears in its strategy proposals in \(0.57\)--\(0.77\) of
cases against \(0.40\)--\(0.69\) with the requests shuffled (\Cref{fig:uptake}). Every agent takes a request up
as its next sub-task above chance, so the hierarchy adds persistence
(\Cref{fig:conversations}). In \Cref{fig:episode5} the finisher role agreed at step \(194\) is re-committed
eleven times until the serve at step \(293\), blocking the loader for \(26\) steps. In the split room the strategic layer can only be as good as what the perception layer tells it: the two
disjoint halves admit only joint pot loading, a constraint the text observation \textit{never states}.
The layer nonetheless plans soundly from what it is given: \(81.9\%\) of split-room proposals avoid relays
and hand-offs, and the remaining \(18.1\%\) are exactly the plans a missing constraint would produce, so
the errors trace to the observation rather than to strategic reasoning; partner-intent accuracy holds at \(0.48\).

\begin{wraptable}{l}{0.4\linewidth}
\vspace{-3pt}
\centering
\caption{Ablation and pinned strategy (seat~1 holds \(g_0\), the pot-loader role; all measures in
  \Cref{tab:hierarchy-app}), per-episode mean and sd over five episodes.}
\label{tab:hierarchy}\label{tab:probe}\label{tab:ablation}
\begin{adjustbox}{width=\linewidth}
\begin{tabular}{llllll}
\toprule
arm & lay & $D\uparrow$ & $\Sigma^-\downarrow$ & $\bar\kappa$ & cmt \\
\midrule
Full PCM & cr & \textbf{\ms{1.4}{0.5}} & \textbf{\ms{27.6}{5.8}} & 0.340 & 0.12 \\
 & as & \ms{0.2}{0.4} & \ms{31.8}{9.5} & 0.343 & 0.14 \\
$-$rollouts & cr & \ms{0.8}{0.8} & \ms{24.6}{12.4} & 0.430 & 0.10 \\
 & as & \ms{0.0}{0.0} & \textbf{\ms{18.8}{9.4}} & 0.415 & 0.12 \\
$-$hierarchy & cr & \ms{0.4}{0.5} & \ms{40.8}{8.0} & 0.536 & 0.42 \\
 & as & \ms{0.4}{0.5} & \ms{33.0}{5.8} & 0.605 & 0.52 \\
$-$ranking & cr & \ms{1.0}{0.9} & \ms{38.4}{11.5} & 0.452 & 0.28 \\
 & as & \textbf{\ms{0.8}{0.7}} & \ms{32.2}{7.9} & 0.445 & 0.28 \\
\midrule
Pinned $g_0$ & cr & \ms{0.6}{0.5} & \ms{35.2}{4.7} & 0.40 & 0.23 \\
 & as & \ms{0.2}{0.4} & \ms{28.6}{8.1} & 0.45 & 0.31 \\
\bottomrule
\end{tabular}
\end{adjustbox}
\end{wraptable}

\paragraph{Removing the strategic layer or the rollouts lowers deliveries; removing the pairwise ranking
does not.}
The full controller delivers \ms{\(1.4\)}{\(0.5\)} soups per episode in the connected room, against
\ms{\(0.8\)}{\(0.8\)} without rollouts, \ms{\(0.4\)}{\(0.5\)} without the strategic layer and \ms{\(1.0\)}{\(0.9\)} without
pairwise ranking (\Cref{tab:hierarchy}); in the split room the arm without ranking delivers most,
\ms{\(0.8\)}{\(0.7\)} against \ms{\(0.2\)}{\(0.4\)}. Without the strategic layer single-branch sets yield
\(\kappa=1\), so the controller commits in \(0.42\)--\(0.52\) of decisions and abandons \ms{\(40.8\)}{\(8.0\)} sub-tasks
per episode against \ms{\(27.6\)}{\(5.8\)}. Without rollouts the two best utilities tie within \(0.02\) in \(26\)--\(32\%\)
of decisions against \(20\)--\(21\%\), although the imagined own position is wrong in \(51\)--\(91\%\) of states: only
a branch's first action is executed, so the order over five branches is all a rollout must supply.
Without pairwise ranking the spread of absolute scores rises from \(0.19\) to \(0.25\) and the commit rate to
\(0.28\): comparison lowers confidence and leaves the acted order unchanged.

\begin{wrapfigure}{r}{0.38\linewidth}
\vspace{-16pt}
\centering
\includegraphics[width=\linewidth]{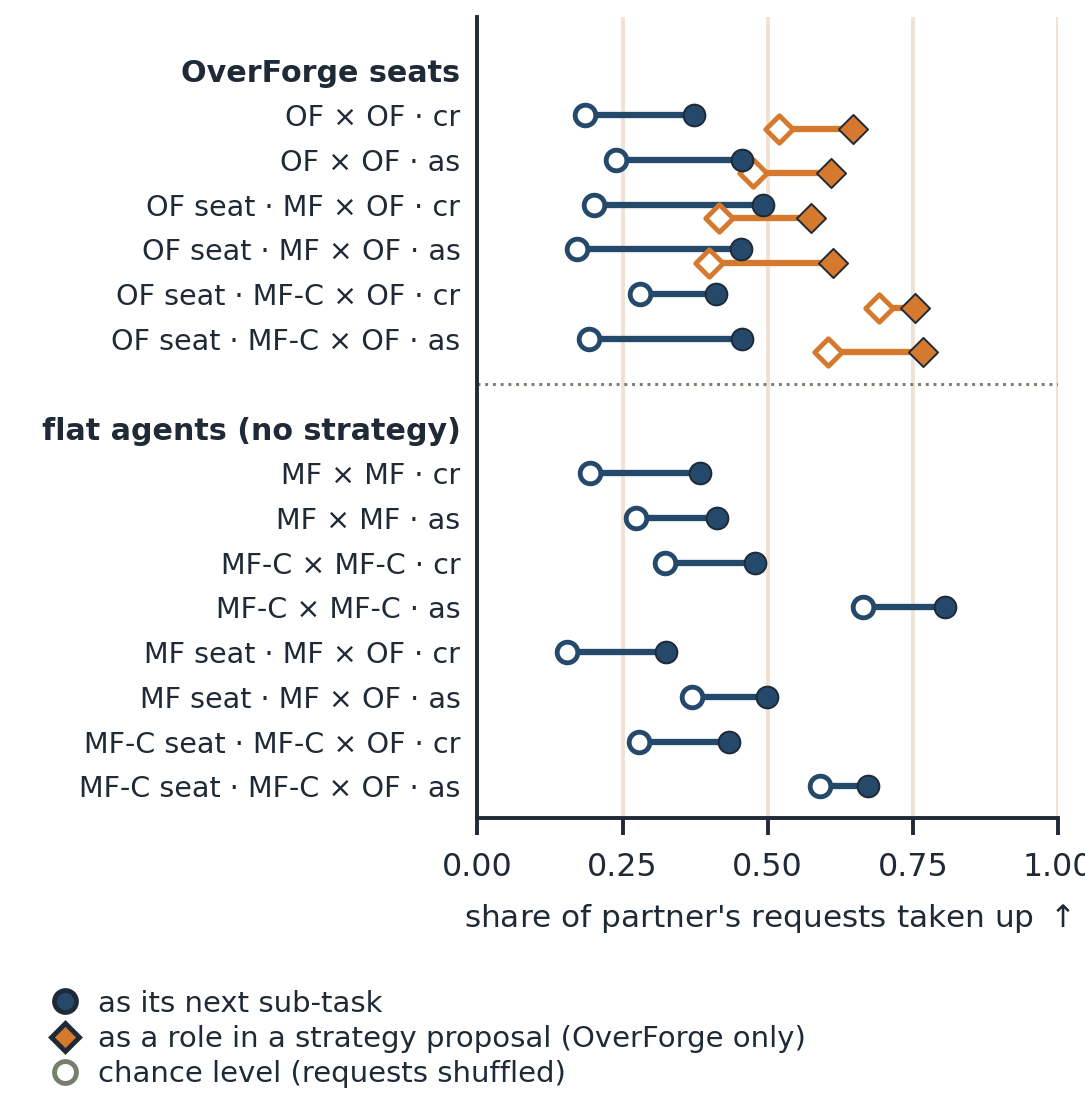}
\vspace{-8pt}
\caption{How often a seat does what its partner asked (\emph{``you grab the plate''}): as its next
  sub-task (circles, all agents) or as a role in one of its own strategy proposals (diamonds, OverForge
  only). Hollow: chance level, requests shuffled.}
\label{fig:uptake}
\vspace{-18pt}
\end{wrapfigure}

\paragraph{A fixed strategy is executed as given; only a partner's concrete offer revises it.}
The probe pins \(g_0\) (\Cref{tab:hierarchy}) on seat~1. Plating and serving make up \(0.19\) and \(0.08\) of
the pinned seat's role events against \(0.31\) and \(0.28\) for its free partner, and it gives \(g_0\) up once in
105 plating delegations, after a concrete offer of the ready soup (\Cref{fig:task-progress}). Physical
infeasibility cannot revise it: under an all-\texttt{ingredient\_0} order that \(g_0\) cannot serve from
the right half, \(158\) of \(300\) steps name an unreachable location.
The tactical level executes the strategy rather than repairing it, so adapting to the room falls onto the strategic layer, and a pinned strategy is a transferrable prior that only a partner can help revise.

\subsection{Lifelong adaptation}

\paragraph{Across episodes, the hierarchy allows OverForge to continually refine a better partner model than MindForge.}
In \Cref{fig:lifelong} (\Cref{app:lifelong}), OverForge's partner-intent accuracy rises from \(0.16\) to \(0.76\)
for agent\_0 and from \(0.44\) to \(0.53\) for agent\_1, while MindForge's stays at \(0.30\) to \(0.28\). Coordination
follows the partner model: the failed-interaction rate falls from \(0.69\) and \(0.85\) to \(0.55\) and \(0.58\), and
completed sub-tasks per seat rise from \(7\) and \(4\) to \(21\) and \(23\) by episode \(4\). Deliveries, blocking, and which seat serves continue to vary across episodes, indicating that OverForge accumulates a partner model, not converging on a single fixed coordination convention.

\paragraph{OverForge adapts to a new partner in what it agrees to, not yet in what it does.}
Paired with MindForge, the OverForge seat receives \(101\) and \(137\) role proposals and counters only \(5\%\) and
\(4\%\) of them, so it works within the partner's plan rather than imposing its own, and it keeps \(67\%\) of its
self-play delivery rate (\Cref{tab:full}). The shortfall is in execution, not agreement: the OverForge seat fails \(0.80\) of its interactions against \(0.29\) and \(0.14\) for MindForge and MindForge-C (\Cref{tab:transfer}), and the partner serves \(5\) of the \(6\) cross-play soups. This is the gap between words and deeds reported for single
LLMs \citep{xu2025large} and the reasoning-action mismatch catalogued in multi-agent LLM systems
\citep{NEURIPS2025_b1041e52}, here seen at the level of a coordination agreement: what OverForge brings to a new
partner is a model of that partner, carried by dialogue, and the final physical interaction is where
coordination still breaks down.

\begin{wraptable}{r}{0.6\linewidth}
\centering
\caption{Full-agent comparison, per-episode mean and sd (full version: \Cref{tab:full-app}); cr/as:
  connected/split room; bold: best LM self-play value per room.}
\label{tab:full}\label{tab:full-comparison}
\begin{adjustbox}{width=\linewidth}
\setlength{\tabcolsep}{3.5pt}
\begin{tabular}{lcccccc}
\toprule
 & \multicolumn{2}{c}{$D\uparrow$} & \multicolumn{2}{c}{FI$\downarrow$} & \multicolumn{2}{c}{IA$\uparrow$} \\
 & cr & as & cr & as & cr & as \\
\midrule
\multicolumn{7}{l}{\emph{Self-play (matched partner)}} \\
OverForge & \textbf{\ms{1.4}{0.5}} & \ms{0.2}{0.4} & \ms{0.65}{0.08} & \ms{0.78}{0.08} & \textbf{\ms{0.48}{0.14}} & \textbf{\ms{0.48}{0.09}} \\
MindForge & \ms{0.6}{0.8} & \textbf{\ms{0.6}{0.8}} & \ms{0.51}{0.18} & \ms{0.32}{0.23} & \ms{0.37}{0.09} & \ms{0.38}{0.04} \\
MindForge-C & \ms{0.6}{0.5} & \ms{0.4}{0.5} & \textbf{\ms{0.24}{0.07}} & \textbf{\ms{0.04}{0.08}} & -- & -- \\
IPPO & \ms{5.6}{3.0} & \ms{0.0}{0.0} & \ms{0.73}{0.11} & \ms{1.00}{0.00} & -- & -- \\
\midrule
\multicolumn{7}{l}{\emph{Cross-play (novel partner; OverForge in seat 2)}} \\
MindForge$\times$OF & \ms{0.4}{0.5} & \ms{0.6}{0.5} & \ms{0.63}{0.11} & \ms{0.67}{0.16} & \ms{0.37}{0.10} & \ms{0.41}{0.09} \\
MindForge-C$\times$OF & \ms{0.2}{0.4} & \ms{0.0}{0.0} & \ms{0.60}{0.09} & \ms{0.63}{0.12} & \ms{0.41}{0.07} & \ms{0.45}{0.10} \\
IPPO$\times$OF & \ms{1.8}{1.2} & \ms{0.2}{0.4} & \ms{0.65}{0.11} & \ms{0.95}{0.05} & -- & -- \\
\bottomrule
\end{tabular}
\end{adjustbox}
\end{wraptable}

\paragraph{Long-term continual adaptation across episodes depends on persistent memory: retaining experience preserves partner models and coordination, whereas erasing it degrades both prediction and task performance.} \Cref{tab:restart} and \Cref{fig:restart} restart self-play at the third and fifth episode with one
seat's memory kept and the other's erased. Across the two connected-room restarts, the restarted teams deliver 2 soups in total, compared with \(7\) over the corresponding intervals of the original run, and the episode-5 restart produces a sharp increase in blocked steps immediately after the memory cut (\Cref{fig:restart}a). Partner modelling degrades at the same time: the kept-memory seat exceeds the erased seat in partner-intent accuracy by \(+0.32\) and \(+0.15\), compared with original-run seat gaps of \(+0.01\) and -\(0.23\) over the same episode ranges. Transcripts show that retained memory preserves beliefs about the partner and previously agreed divisions of labour (\Cref{app:conversations}). Together, these results localise cross-episode transfer in memory: accumulated partner knowledge is reused in later episodes to support coordination, linking the hierarchy directly to continual adaptation.

\begin{wrapfigure}{r}{0.38\linewidth}
\vspace{-26pt}
\centering
\includegraphics[width=\linewidth]{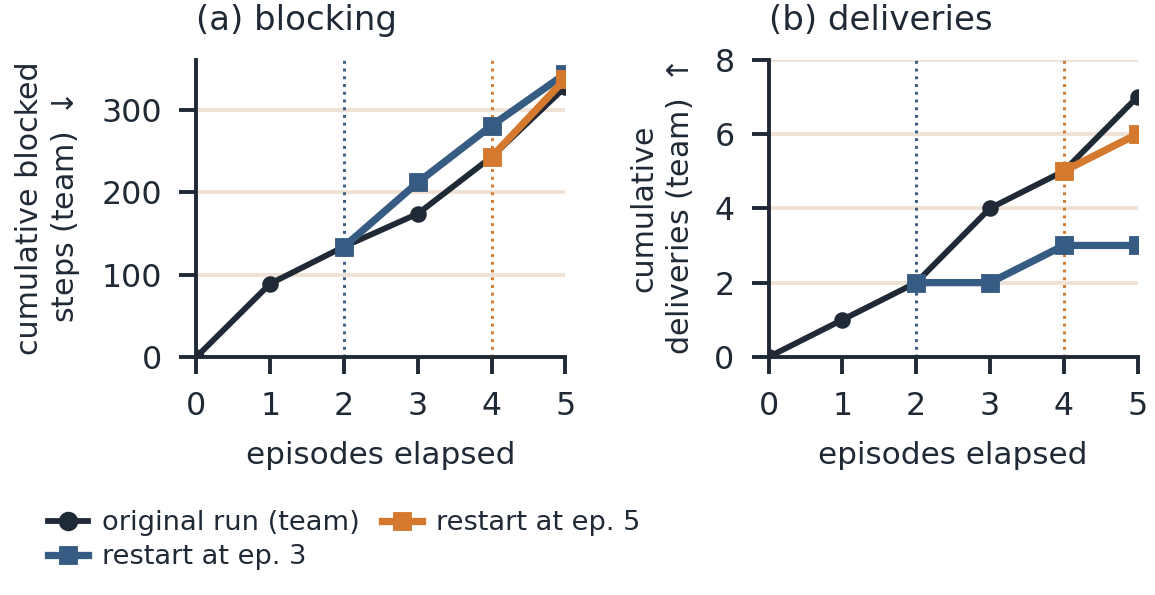}
\vspace{-8pt}
\caption{Cumulative (a) blocked steps and (b) deliveries of the team in the connected room, original run
and memory restarts (cut dotted). Per-seat partner predictions: \Cref{fig:restart-ia}.}
\label{fig:restart}
\end{wrapfigure}

\section{Discussion and conclusion}\label{conclusion}
The results suggest three lessons for cooperative language agents
(\Cref{app:field-impact}). First, alongside efforts to improve cooperation by scaling multi-agent systems \citep{park2026scalingdiscoverytesttimecommunication,anthropic2026multiagent},
our results highlight the value of preserving coordination agreements within
each agent: assigned roles reappear in strategic proposals, and removing the
strategic layer reduces deliveries in the connected kitchen. But an agreed role is useful only if
the kitchen allows the agent to fulfil it
\citep{mieczkowski2025normative}. Second, imperfect predictions
can still help an agent choose: removing the lookahead rollouts reduces deliveries
despite errors in imagined positions. This makes the effect of prediction
on decisions worth measuring alongside state accuracy. Third, experience
with a partner does not necessarily teach an agent what its environment
permits. Erasing memory disrupts cooperation in the connected kitchen,
while retaining it does not consistently help in the split kitchen.
Evaluations of lifelong cooperation should therefore examine both what
agents learn about partners and what they learn about their surroundings
\citep{biswas_who_2026}.

OverForge makes the distinction between strategy and tactics explicit,
allowing us to test how roles, prediction and accumulated experience shape
cooperation. Five episodes per condition, each starting with a randomly
sampled recipe, assess consistency as task requirements vary and experience
accumulates. Further runs are needed to establish robustness across learning
histories, models and human partners. The split-kitchen failures also point
to a concrete next step: agents need to learn from unsuccessful attempts
which strategies are infeasible, so that an agreement with a partner can be
revised when the environment prevents it from working.

\subsection*{AI use statement}

In this work, we used generative AI tools (large-language-model assistants, including an agentic
coding assistant) for the following tasks with required disclosure: \emph{propose or refine
hypotheses} and \emph{design or provide feedback on research methodology or experiments}, in the
form of research ideation before the authors fixed the design, for example ``Which single PCM
component, if removed, would separate the effect of pairwise ranking from the effect of rollouts?'';
and \emph{implement methods}, in the form of research execution, that is writing and refactoring
experiment code, analysis scripts and cluster job files under author direction, for example ``Add a
flag that disables pairwise ranking in the PCM and scores branches by absolute facet means only;
keep all logging unchanged and add a test that both paths propose the same branch set.'' We have not
used generative AI tools to interpret results, to help develop theoretical models or conceptual
frameworks, or to support qualitative and thematic data analysis, and generating synthetic data
sets, formulating mathematical claims, providing critical ingredients for proving mathematical claims,
assisting in the writing of proofs, assisting with translation, and cleaning or reformatting a
dataset are not applicable to this work. Additionally, we used generative
AI tools for two tasks with recommended disclosure: \emph{edit a research paper to improve
readability}, that is sentence-level polishing of author-written paragraphs, for example ``Tighten
this paragraph to three sentences without changing any number or claim.''; and \emph{draft parts of a
research paper}, namely first drafts of the appendix metric definitions and of results paragraphs
from tables and interpretations the authors supplied, for example ``Draft one paragraph reporting
this table for the results section; state directions, not effect sizes.'' The creation or editing of
software code falls under the implementation described above. Interpreting the results,
summarising and identifying the literature, formatting references, choosing the title and keywords,
and structuring the paper were done by the authors without generative AI. We have reviewed all
AI-assisted work. The research questions, the architecture, the experimental design and the
interpretation of the results are the authors' own. All AI-drafted prose was rewritten by the
authors, and every number in the paper was checked against the experiment logs. AI-generated code
was reviewed, run and tested by the authors, and the analysis scripts that produce the reported
tables were verified by recomputing values from the raw per-step logs. We take responsibility for
the final content of this work, including text, claims or artifacts produced with the aid of
generative AI.

\subsection*{Ethics statement}

This work studies language-model agents cooperating in a simulated cooking game. It involves no
human subjects, no personal data and no user study; every observation, message and trace is generated
by simulation. All agents run one frozen open-weight model on local hardware, so no data leave the
compute environment. The introduction cites a documented incident in which autonomous agents
coordinated a harmful multi-day operation; that incident motivates the study of persistent
coordination, but our method is training-free, adds no capability to the underlying model, and is
evaluated only with cooperative partners on a toy task. We note that more reliable coordination among
language-model agents is dual-use, and we report failure modes (infeasible strategies, wrong
reachability inferences) alongside the gains. The authors declare no conflicts of interest or
sponsorship concerns.

\subsection*{Reproducibility statement}

The environment, the agents and every experimental condition are specified in the paper.
\Cref{exp-design} gives the layouts, horizon, seed, recipe sampling, the frozen model and how it is
served; \Cref{app:notation-full} lists every symbol and hyper-parameter with the value used;
\Cref{app:pcm-details} gives the PCM in pseudocode; \Cref{app:adaptations} states how each reference
agent was adapted; \Cref{app:conditions,app:metrics} define every partner condition and every
measure; \Cref{app:prompts} reproduces all prompts and response templates verbatim; and
\Cref{app:additional-results} reports the complete tables behind the abridged ones in the body. The
source code, configuration files and raw per-step logs of every run reported here will be released
with the camera-ready version.

\bibliography{overforge}
\bibliographystyle{iclr2027_conference}

\appendix

\crefalias{section}{appendix}
\crefalias{subsection}{appendix}
\crefalias{subsubsection}{appendix}
\section{Appendix}\label{appendix}
\subsection{Full notation}\label{app:notation-full}
\Cref{tab:notation} gives the complete symbol set. The body of the paper uses the
first three groups; the fourth collects symbols that appear only in the appendices.

\begin{table}[th]
  \caption{Notation by decision stage. Plain symbols are real, \(\hat{\cdot}\) is inferred by the agent,
    \(\tilde{\cdot}\) is predicted and never executed; \(w^i_t\) is the maintained world model and \(p_\phi\) the
  forward model.}
  \label{tab:notation}
  \centering
  \small
  \begin{adjustbox}{max width=\linewidth}
    \begin{tabular}{ll}
      \toprule
      symbol & reading \\
      \midrule
      \multicolumn{2}{l}{\emph{Problem level --- the cooperative task}}\\
      \(N\), \(i,j\) & number of agents (\(N=2\) except in the scaling sweep); agent indices \\
      \(\mathcal{S}\), \(s_t\) & kitchen state space; the real state at step \(t\) \\
      \(\mathcal{A}\), \(a^i_t\), \(\mathbf{a}_t\) & the six primitive actions; agent \(i\)'s action; the
      joint action \\
      \(P\) & joint transition kernel \(P(s_{t+1}\mid s_t,\mathbf{a}_t)\) \\
      \(\mathcal{O}^i\), \(Z\) & observation space of agent \(i\); observation kernel \(Z(o\mid s,i)\) \\
      \(r\) & delivery reward (\(r=20\) per correct delivery), logged only \\
      \(T\) & episode horizon, \(T=300\) steps \\
      \(\ell\), \(R\) & environment: a layout \(\ell\) and a recipe set \(R\) (lifelong axis) \\
      \(G\) & the strategies the constraints of the environment afford \\
      \(\pi^{j}\), \(\pi^{-i}\) & policy of partner \(j\); the partner profile faced by agent \(i\) (cultural axis) \\
      \(\Phi_i\), \(\Phi\) & the controller of agent \(i\), \Cref{eq:controller}; the set \(\{\Phi_i\}_{i=1}^N\) \\
      \midrule
      \multicolumn{2}{l}{\emph{Representation and proposal --- what the agent holds and offers}}\\
      \(o^i_t\) & structured local observation at step \(t\), rendered to text \\
      \(\hat{B}^i_t\), \(\mathcal{F}\) & inferred belief facets \(\{\hat B^{i,f}_t\}_{f\in\mathcal{F}}\);
      \(\mathcal{F}=\{\)perception, task, partner, interaction\(\}\) \\
      \(\hat{w}^{\,i\to j}_t\) & agent \(i\)'s structured model of teammate \(j\) \\
      \(w^i_t\) & maintained world model, \Cref{eq:wmodel} \\
      \(M^i_t\), \(\bar M^i_t\) & multi-component memory; the summary retrieved for this decision \\
      \(m^i_t\) & message emitted to teammates at step \(t\) \\
      \(q_{\mathrm{strat}}\), \(q_{\mathrm{tact}}\) & strategic and tactical proposal distributions (same frozen LLM) \\
      \(g\), \(K\) & a strategy, i.e.\ a goal-level plan; the number proposed per step (\(K=2\)) \\
      \(\tilde a\), \(m\) & a candidate first action, proposed but not executed; candidates per strategy (\(m\le 3\)) \\
      \(b=(g,\tilde a)\), \(\mathcal{B}\) & one branch, \Cref{eq:branch}; the pooled branch set,
      \(|\mathcal{B}|\le Km\) \\
      \midrule
      \multicolumn{2}{l}{\emph{Imagination and commitment --- how one branch is chosen}}\\
      \(p_\phi\) & generative forward model: the frozen LLM used as a predictor, \Cref{eq:roll} \\
      \(\tilde w^{(d)}_b\), \(d_b\), \(d_{\max}\) & predicted clone of branch \(b\) at depth \(d\); depth \(b\) has
      reached; cap (\(=5\)) \\
      \(\nu(\cdot)\) & LLM state-value judge, \(\nu:\tilde w\mapsto[0,1]\), \Cref{eq:vtraj} \\
      \(V_{\mathrm{imm}}\), \(n_{\mathrm{cmp}}\) & judged value of the first action, \Cref{eq:vimm}; comparison
      budget (\(=6\)) \\
      \(V_{\mathrm{traj}}\) & judged value of the situation the branch leads to, \Cref{eq:vtraj} \\
      \(U^{(d)}(b)\) & branch utility after deliberation round \(d\), \Cref{eq:U} \\
      \(\mu(b)\), \(\tau\) & branch posterior \(\mu=\mathrm{softmax}(U/\tau)\), \Cref{eq:post}; its
      temperature (\(=0.1\)) \\
      \(H\), \(\kappa\) & Shannon entropy; metacognitive confidence, \Cref{eq:kappa} \\
      \(\theta\) & confidence threshold above which the agent commits (\(=0.6\)) \\
      \(\rho\), \(\mathcal{R}\) & posterior mass retained when deliberation deepens (\(=0.9\)); the nucleus,
      \Cref{eq:nucleus} \\
      \(a^{*i}_t\), \(a^i_t\) & the committed action at step \(t\); the action actually executed \\
      \bottomrule
    \end{tabular}
  \end{adjustbox}
\end{table}

\subsection{Strategy and tactic examples}\label{app:strategy-examples}
\Cref{tab:strategy_tactics} lists, for each two-agent layout, the constraints it imposes, two
strategies those constraints afford, and tactics that realise each strategy. It illustrates the
distinction of \Cref{prelim}: the strategy is stated without coordinates or primitive actions, and
the same strategy maps to different action sequences as the state changes.

\begin{table}[th]
  \centering
  \footnotesize
  \caption{Room-specific strategies and representative tactics. \(M_{\uparrow}\), \(M_{\downarrow}\),
  \(M_{\leftarrow}\), \(M_{\rightarrow}\): move; \(I\): interact; \(S\): stay.}

  \label{tab:strategy_tactics}
  \begin{adjustbox}{width=\linewidth}
    \begin{tabular}{p{1.7cm}p{4.0cm}p{4.4cm}p{5.6cm}p{2.0cm}}
      \toprule
      \textbf{Room} &
      \textbf{Constraints addressed} &
      \textbf{Strategic intention} &
      \textbf{Representative tactics (purpose)} &
      \textbf{Actions} \\
      \midrule

      Cramped Room &
      Narrow shared corridors, frequent collisions, limited maneuvering space &
      "I will coordinate movement with my teammate by yielding bottlenecks when needed." &
      Wait before entering bottleneck (avoid blocking teammate); reroute around occupied tiles (maintain
      movement flow). &
      \(S\), \(S\), \(M_{\leftarrow}\), \(M_{\uparrow}\), \(M_{\rightarrow}\) \\

      Cramped Room &
      Shared workspace and repeated path interference &
      "I will stay on my assigned side and minimise unnecessary crossings." &
      Remain within assigned region (reduce interference); temporarily yield corridor (allow teammate to pass). &
      \(M_{\uparrow}\), \(M_{\uparrow}\), \(S\), \(S\), \(M_{\downarrow}\) \\

      Asymmetric Advantages &
      Resources and workstations are split between players; asymmetric reachability &
      "I will specialise in supplying resources that my teammate cannot easily access." &
      Perform ingredient hand-offs (bridge inaccessible resources); optimise pickup timing (reduce teammate
      idle time). &
      \(M_{\rightarrow}\), \(M_{\rightarrow}\), \(I\), \(M_{\leftarrow}\), \(I\) \\

      Asymmetric Advantages &
      Each player has privileged access to different ingredients or stations &
      "I will own my assigned ingredients and coordinate deliveries with my teammate." &
      Collect only assigned ingredients (avoid duplicated work); synchronise deliveries with partner
      (maintain pipeline). &
      \(M_{\uparrow}\), \(M_{\rightarrow}\), \(I\), \(M_{\downarrow}\), \(I\) \\

      \bottomrule
    \end{tabular}
  \end{adjustbox}
\end{table}

\subsection{Reference-agent adaptations}\label{app:adaptations}
The two language-model baselines are re-implementations of published agents inside one code base,
and IPPO is a trained policy, so each needed adaptations to run in Overcooked-v2 under a shared
protocol. This section lists them and the reason for each, because a difference between agents is
only attributable to the hierarchy if the surrounding components are held equal.

\paragraph{Shared protocol.}
Every language-model agent uses the same frozen model, the same observation rendering, the same
outcome-triggered dialogue of three turns, and the same critic, curriculum, skill and episodic-memory
components; the delivery reward is logged and never supplied to any prompt. Holding these constant is
what turns the three agents into one lineage, MindForge \(\to\) MindForge-C \(\to\) OverForge, in which
each step adds one mechanism. MindForge natively runs six dialogue turns; one shared dialogue cannot
carry two turn counts, so all agents use three.

\paragraph{MindForge.}
The original MindForge \citep{lica_mindforge_2025} pairs an asymmetric weak and strong agent in
Minecraft, communicates through an external server, and uses a binary critic. Our version is
symmetric and supports \(N\) agents, because Overcooked chefs are homogeneous; communication is
in-process and round-robin, because the simulator is single-process and no beliefs are passed
directly; the world-model record is re-expressed for Overcooked observations, with a structured
mental-model manager for the partner; and the critic returns three verdicts (success, progress,
failure) instead of two, because per-step control loops produce many steps of partial progress that a
binary critic would have to call failures. MindForge has no semantic memory, as in the original.

\paragraph{MindForge-C.}
MindForge-C adds one causal predict-and-reselect pass to MindForge: before acting, the agent asks the
same frozen model to imagine the consequence of its chosen action, in interventional terms (what the
action changes and what it leaves unchanged), and reconsiders the action in the light of that
prediction. The design follows the causal-world-model agents of \citet{ag_mercier_surgical_2025},
with three departures. The forward model is the prompted language model rather than a trained
BISCUIT model over images, since the environment is symbolic and no training is used; the forward
model takes only the observation and the action, so beliefs and memory enter at re-selection and not
at prediction; and the surgical interventions of the original are dropped, since they presuppose a
trained causal model. Two protocol adaptations follow from the shared setup: the original per-step
broadcast is replaced by the shared outcome dialogue, and semantic memory is added so that the memory
stack matches OverForge's. MindForge-C keeps no mental model of the partner, as in the original, so a
difference between MindForge-C and MindForge reflects the forward model together with the absence of
partner modelling, not the forward model alone.

\paragraph{IPPO.}
IPPO \citep{witt_is_2020} is the JaxMARL implementation \citep{rutherford_jaxmarl_2024} with a
recurrent (GRU) actor-critic, trained once in \texttt{cramped\_room} self-play for \(10^7\)
environment steps and then frozen. Its final training return corresponds to about 10.6 soups per
episode under the training horizon. Because the policy's input is a fixed-shape grid, observations
from both layouts are zero-padded to one canonical shape, the element-wise maximum of the two
layouts' shapes, so that the same checkpoint runs in both kitchens without retraining. This makes the
\texttt{asymm\_advantages} evaluation out of distribution in two ways at once: the kitchen is unseen
and the padded tensor places cells at indices that were always zero during training. The collapse to
zero deliveries there (\Cref{tab:full-comparison}) therefore reflects a policy tied to its training
kitchen's spatial encoding, and should not be read as a statement about reinforcement learning in
general. IPPO does not communicate and exposes no sub-task, so the dialogue and ledger measures are
undefined for it.

\subsection{PCM design details}\label{app:pcm-details}
\label{app:pcm-design}

This appendix expands the Prefrontal Cortex Module (PCM) summarised in
\Cref{methodology} and \Cref{alg:pcm}. The PCM implements the
prediction and commitment stages of the per-agent controller
\(\Phi:(w_t,M_t)\mapsto a_t^*\). Its computation is divided into three reusable
procedures: branch proposal (\Cref{alg:propose}), immediate-action ranking (\Cref{alg:rank}), and
partner-conditioned rollout (\Cref{alg:extend}); \Cref{alg:pcm-nucleus} gives the nucleus selection
used between deliberation rounds. These procedures preserve the distinction between the real maintained
world model \(w_t\) and predicted clones \(\tilde w^{(d)}\). Only the final
committed action \(a_t^*\) reaches the environment; all candidate actions,
predicted states, values, and branch posteriors remain internal to the
decision. \Cref{alg:pcm} states the full deliberation loop; the equations it cites that the main
text gives inline are repeated below with numbers.

\begin{algorithm}[th]
  \caption{\textsc{PCM-Decide}\((w_t,\bar M_t)\) --- one deliberation}
  \label{alg:pcm}
  \begin{algorithmic}[1]
    \REQUIRE \(w_t\), \(\bar M_t\); strategies \(K\), comparison budget \(n_{\mathrm{cmp}}\), temperature \(\tau\),
    threshold \(\theta\), nucleus mass \(\rho\), depth cap \(d_{\max}\)
    \STATE \(\mathcal{B} \leftarrow \textsc{ProposeBranches}(w_t,\bar M_t,K)\)
    \COMMENT{\Cref{eq:branch}; \Cref{alg:propose}}
    \STATE \(V_{\mathrm{imm}}(b) \leftarrow \textsc{RankImmediate}(A_{g_b}, w_t,
    n_{\mathrm{cmp}}/K)\;\;\forall b\in\mathcal{B}\)
    \COMMENT{\Cref{eq:vimm}; \Cref{alg:rank}; computed once}
    \STATE \(\textsc{Extend}(b)\;\;\forall b\in\mathcal{B}\); \; \(d\leftarrow1\)
    \COMMENT{\Cref{eq:roll,eq:vtraj}; \Cref{alg:extend}}
    \STATE compute \(U^{(d)},\,\mu^{(d)},\,\kappa^{(d)}\)
    \COMMENT{\Cref{eq:U,eq:post,eq:kappa}}
    \WHILE{\(\kappa^{(d)} < \theta\) \textbf{ and } \(d < d_{\max}\)}
    \STATE \(\mathcal{R}^{(d)} \leftarrow \textsc{Nucleus}(\mu^{(d)},\rho)\)
    \COMMENT{\Cref{eq:nucleus}; \Cref{alg:pcm-nucleus}}
    \STATE \(\textsc{Extend}(b)\) for every extendable \(b\in\mathcal{R}^{(d)}\)
    \COMMENT{one depth each; \(d_b\) advances by \Cref{eq:depth}}
    \IF{no branch was extended}
    \STATE \textbf{break} \COMMENT{every live branch is forced (\(\le1\) candidate)}
    \ENDIF
    \STATE \(d \leftarrow d+1\); recompute \(U^{(d)},\,\mu^{(d)},\,\kappa^{(d)}\)
    \COMMENT{\(V_{\mathrm{imm}}\) fixed; \(V_{\mathrm{traj}}\) frozen outside \(\mathcal{R}^{(d-1)}\)}
    \ENDWHILE
    \STATE \(b^{\star} \leftarrow \arg\max_{b\in\mathcal{B}} U^{(d)}(b)\) \textbf{ if } \(\kappa^{(d)}\ge\theta\)
    \textbf{ else } \(b^{\star}\sim\mu^{(d)}\) \COMMENT{\Cref{eq:commit}}
    \RETURN \(\tilde a_{b^{\star}}\) \COMMENT{the only action that is executed}
  \end{algorithmic}
\end{algorithm}

\paragraph{Private representation and cloning.}
At step \(t\), the maintained world model \(w_t\) combines the local observation
\(o_t\), inferred belief facets \(\hat b_t\), task progress, and a structured
partner model. The memory summary \(M_t\) carries relevant prior experience into
strategic and tactical proposal. Because \(w_t\) represents the agent's current
information state, it is updated only from real observations. Prediction begins
by cloning it into \(\tilde w^{(0)}\) for each branch; the generative forward
model \(p_\phi\) then updates only these clones,
\begin{equation}\label{eq:roll}
  \tilde w^{(d+1)}_b \;\sim\; p_\phi\big(\cdot \;\big|\; \tilde w^{(d)}_b,\; \mathrm{do}(\tilde a^{(d)}_b)\big),
  \qquad \tilde a^{(0)}_b = \tilde a_b,
\end{equation}
which prevents one imagined future
from contaminating another or altering the representation used by the real
agent. An LLM judge \(\nu\) values the deepest imagined state that a branch has reached,
\begin{equation}\label{eq:vtraj}
  V_{\mathrm{traj}}^{(d)}(b) \;=\; \nu\big(\tilde w^{(d_b)}_b\big) \;\in\; [0,1].
\end{equation}

\paragraph{Root-level hierarchical branching.}
The strategic distribution
\(\pi_{\mathrm{strat}}(\cdot\mid w_t,M_t)\) proposes \(K\) strategies \(g\) describing
persistent roles or coordination intentions. For each \(g\), the tactical
distribution \(\pi_{\mathrm{tact}}(\cdot\mid w_t,g)\) proposes candidate first
actions \(\tilde a\). Every pair \(b=(g,\tilde a)\) becomes a separate branch in
the pooled branch set,
\begin{equation}\label{eq:branch}
  \mathcal{B} \;=\; \big\{\, b=(g,\tilde a) \;:\; g\sim q_{\mathrm{strat}}(\cdot\mid w_t,\bar M_t),\;\;
  \tilde a \sim q_{\mathrm{tact}}(\cdot\mid w_t, g) \,\big\},\qquad |\mathcal{B}|\le Km .
\end{equation}
Branching occurs at the first action because this is
the only imagined action that may execute. If several actions beneath one
strategy shared a rollout, trajectory value could re-rank strategies but could
not change which first action is selected. Deeper rollout steps therefore
follow one sampled tactical path per root branch.
\Cref{fig:branching} draws this branch system next to one logged decision.

\begin{figure}[th]
  \centering
  \includegraphics[width=\linewidth]{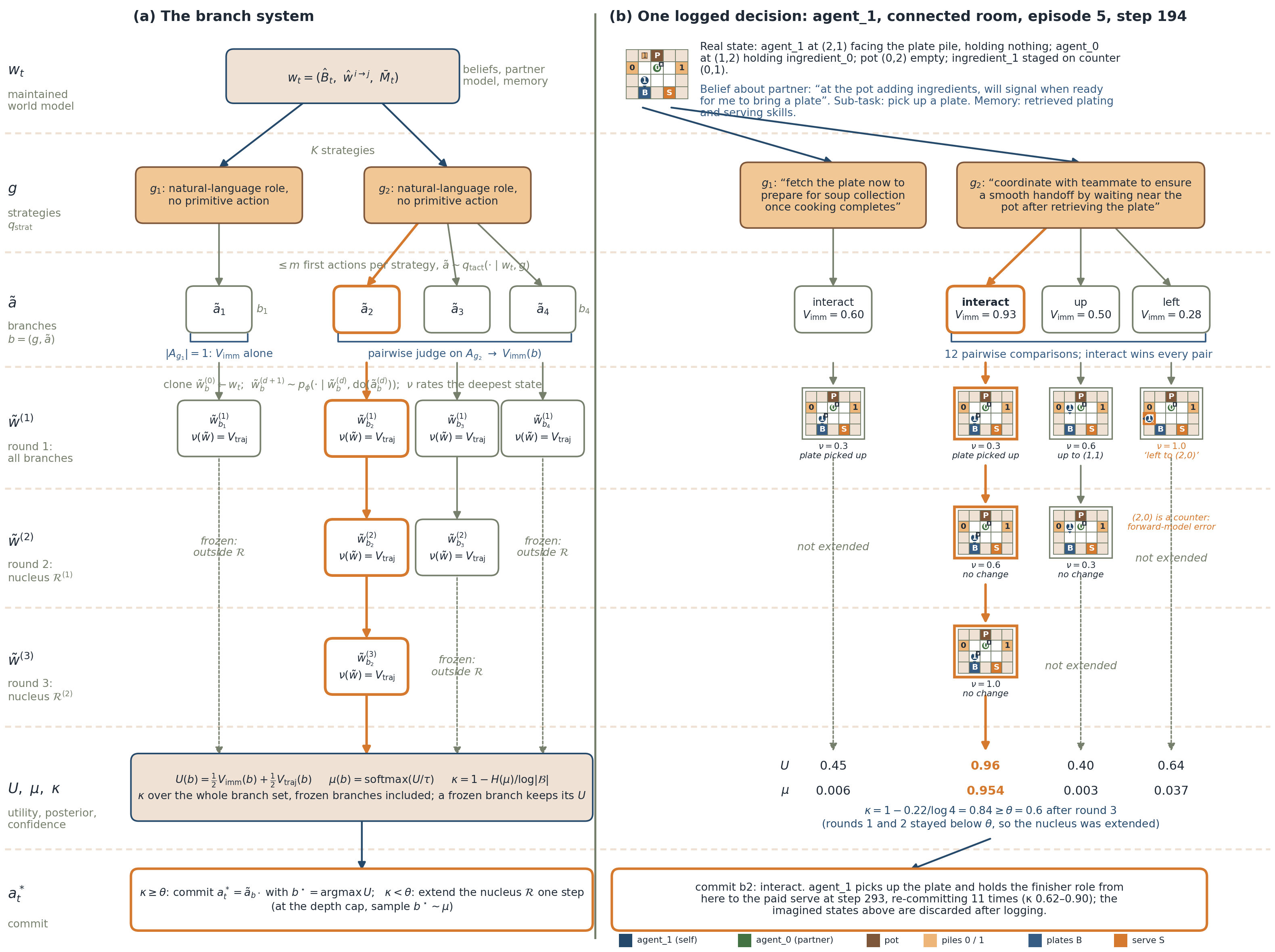}
  \caption{\textbf{The branch system (a) next to one logged decision (b).} Rows follow one deliberation
    from the maintained world model \(w_t\) through the \(K=2\) strategies, the first-action branches
    \(b=(g,\tilde a)\) with their pairwise-judged \(V_{\mathrm{imm}}\), the rollout rounds that extend only
    the nucleus, and the commit gate. (b) is agent\_1 at step 194 of episode 5 in the connected room
    (\Cref{fig:episode5}, conversation C2): two strategies yield four branches; the comparison prefers
    \emph{interact} (pick up the plate) under both; round 1 rolls every branch one step, rounds 2 and 3
    extend only the nucleus, and after round 3 the posterior concentrates on \(b_2\) with
    \(\kappa=0.84\ge\theta\), so the agent commits. Imagined states are the forward model's own predictions
    and carry its errors: \(b_4\) places the agent on the counter cell (2,0), and \(b_2\) predicts no change
    at depths 2 and 3 while the state judge still raises \(\nu\); \Cref{tab:horizon} quantifies both. All values are read
    from the PCM trace.}
  \label{fig:branching}
\end{figure}

\paragraph{Two uses of softmax.}
Softmax serves two distinct roles. During a rollout, deeper tactical actions are
sampled from immediate values using temperature \(\tau\), because their
trajectory values do not yet exist. This introduces controlled diversity among
predicted paths without affecting the real environment. After each rollout
depth, the PCM instead applies softmax to the complete utilities \(U(b)\) to form
the branch posterior \(p(b)\). This second distribution determines confidence,
nucleus retention, and the depth-cap fallback. Separating these distributions
prevents incomplete trajectory estimates from entering deeper action selection
while allowing commitment to consider both immediate and longer-horizon
evidence.

\paragraph{Deliberation rounds and commitment.}
Deliberation proceeds in rounds \(d=1,\dots,d_{\max}\), and the depth \(d^{(d)}_b\) that branch \(b\) has
reached after round \(d\) advances only for extendable branches inside the nucleus of the previous
round,
\begin{equation}\label{eq:depth}
  d^{(1)}_b = 1, \qquad
  d^{(d)}_b \;=\; d^{(d-1)}_b \;+\; \mathbb{1}\!\left[\,b \in \mathcal{R}^{(d-1)} \;\wedge\; b \text{
  extendable}\,\right].
\end{equation}
After each round the branch posterior of \Cref{eq:post} gives the metacognitive confidence
\begin{equation}\label{eq:kappa}
  \kappa^{(d)} \;=\; \mathrm{clip}\!\left(\,1-\frac{H(\mu^{(d)})}{\log|\mathcal{B}|},\;0,\;1\right),
  \qquad \kappa \equiv 1 \ \text{ if } |\mathcal{B}|=1 .
\end{equation}
Computing \(\kappa\) over all of \(\mathcal{B}\), including the branches frozen outside the nucleus, is
what keeps the commit behaviour from jumping as the beam narrows.
If \(\kappa^{(d)}<\theta\), the next round deepens the nucleus, the smallest set of
highest-probability branches whose mass reaches \(\rho\),
\begin{equation}\label{eq:nucleus}
  \mathcal{R}^{(d)} \;=\; \operatorname*{arg\,min}_{\mathcal{R}\subseteq\mathcal{B}} \;
  \Big\{\, |\mathcal{R}| \;:\; \textstyle\sum_{b\in\mathcal{R}}\mu^{(d)}(b) \ge \rho \,\Big\}.
\end{equation}
The executed branch is the highest-utility one once confidence reaches the threshold, and is
sampled from the posterior otherwise:
\begin{equation}\label{eq:commit}
  b^\star=
  \begin{cases}
    \arg\max_{b\in\mathcal{B}} U^{(d)}(b) & \text{if } \kappa^{(d)} \ge \theta \quad(\text{commit}),\\[2pt]
    b\sim\mu^{(d)} & \text{otherwise} \quad(\text{depth-limited fallback}),
  \end{cases}
  \qquad a^*_t=\tilde a_{b^\star} .
\end{equation}

\paragraph{Fixed computation parameters.}
We use \(K=2\) strategic proposals, six pairwise comparisons, a rollout-depth cap
of five, \(\theta_{\mathrm{commit}}=0.6\), \(\tau=0.1\), and \(\rho=0.9\). The low
temperature preserves meaningful differences between values in \([0,1]\) while
avoiding deterministic selection from shallow evidence. The confidence
threshold requires clearer separation than a bare plurality, and the depth cap
bounds inference when alternatives remain ambiguous. Retaining \(90\%\) of
posterior mass concentrates further prediction on plausible branches without
prematurely collapsing to one. Coverage and refinement provide limited benefit
for very small branch sets but support larger candidate sets without changing
the decision rule.

\paragraph{Judge fallbacks.}
The pairwise judge does not always return a verdict: in OverForge self-play 19\% (connected room) and
21\% (split room) of comparisons were unparseable --- mostly an imagined-state object echoed in place
of the judgement at rollout depth, otherwise a reply cut off by the token limit --- and each such
comparison entered \Cref{eq:vimm} as a neutral 0.5 for both actions with no preference recorded.
Without rollouts the rate is 3.5\%, so the failures concentrate in imagined states and shrink
\(V_{\mathrm{imm}}\) toward 0.5 at depth; the ranking results should be read with this in mind.

\paragraph{Receding-horizon execution.}
Once the PCM commits, only \(a_t^*\) is returned by the controller and executed
as \(a_t\). Strategies, candidate actions, predicted clones, and rollout paths are
discarded after logging. The next real observation updates \(w_t\), and the
entire process repeats. This receding-horizon design limits the effect of
forward-model error: imagined actions guide only the next commitment rather
than becoming an open-loop action sequence. The logged trace contains branch
values, posterior probabilities, confidence, retained branches, and rollout
depth, enabling the process-level analyses described in
\Cref{app:metrics}.

\begin{algorithm}[th]
  \caption{\textsc{ProposeBranches}\((w_t,\bar M_t,K)\)}
  \label{alg:propose}
  \begin{algorithmic}[1]
    \REQUIRE maintained world model \(w_t\), memory summary \(\bar M_t\), number of strategies \(K\)
    \STATE \(\mathcal{B}\leftarrow\emptyset\)
    \STATE propose \(K\) strategies \(g\sim q_{\mathrm{strat}}(\cdot\mid w_t,\bar M_t)\)
    \COMMENT{roles and intentions only, no primitive actions}
    \FORALL{proposed \(g\)}
    \STATE \(A_g \leftarrow \{\tilde a\}\sim q_{\mathrm{tact}}(\cdot\mid w_t,g)\), \(|A_g|\le m\)
    \STATE \(\mathcal{B}\leftarrow\mathcal{B}\cup\{(g,\tilde a): \tilde a \in A_g\}\)
    \COMMENT{one branch per executable first action, \Cref{eq:branch}}
    \ENDFOR
    \RETURN pooled branch set \(\mathcal{B}\), \(|\mathcal{B}|\le Km\)
  \end{algorithmic}
\end{algorithm}

\begin{algorithm}[th]
  \caption{\textsc{RankImmediate}\((A, w, n)\) --- immediate value of one candidate set}
  \label{alg:rank}
  \begin{algorithmic}[1]
    \REQUIRE candidate actions \(A\) of a single strategy, a representation \(w\in\{w_t,\tilde w^{(d)}\}\),
    comparison budget \(n\)
    \STATE \(E(\tilde a)\leftarrow 0\), \(\mathcal{C}(\tilde a)\leftarrow\emptyset\) for all \(\tilde a\in A\)
    \STATE \(\mathcal{P}\leftarrow\) coverage pairs: each \(\tilde a\) paired with its next two successors
    \COMMENT{complete round-robin when \(|A|\le3\); degree \(\ge2\) otherwise}
    \WHILE{budget \(n\) remains}
    \STATE take the next unjudged pair from \(\mathcal{P}\), or --- once \(\mathcal{P}\) is exhausted ---
    the unjudged pair minimising \(|E(\tilde a)-E(\tilde a')|\)
    \COMMENT{spend what is left on the least certain ordering}
    \STATE judge both actions on coherence, goal-directedness, epistemic value and social fit,
    conditioned on \(w\)
    \STATE record both facet means in \(\mathcal{C}(\cdot)\); \(E \mathrel{+}= \pm\gamma\) for
    winner/loser, \(\gamma=|\)facet-mean difference\(|\)
    \ENDWHILE
    \RETURN \(V_{\mathrm{imm}}(\tilde a)\) for all \(\tilde a\in A\) \COMMENT{\Cref{eq:vimm};
    \(E\) steers the budget only and does not enter the value}
  \end{algorithmic}
\end{algorithm}

\begin{algorithm}[th]
  \caption{\textsc{Extend}\((b)\) --- advance one branch by a single imagined step}
  \label{alg:extend}
  \begin{algorithmic}[1]
    \REQUIRE branch \(b=(g,\tilde a)\) with clone \(\tilde w^{(d_b)}_b\), temperature \(\tau\)
    \IF{\(d_b = 0\)}
    \STATE \(\tilde w^{(0)}_b \leftarrow \textsc{Clone}(w_t)\); \; \(\tilde a^{(0)}_b \leftarrow \tilde a\)
    \COMMENT{the real \(w_t\) is never written}
    \ELSE
    \STATE \(A \leftarrow \{\tilde a'\}\sim q_{\mathrm{tact}}(\cdot\mid \tilde w^{(d_b)}_b, g)\)
    \IF{\(|A| \le 1\)}
    \STATE mark \(b\) not extendable; \textbf{return} \textsc{false}
    \COMMENT{forced move: deeper rollout adds no decision information}
    \ENDIF
    \STATE \(V_{\mathrm{imm}} \leftarrow \textsc{RankImmediate}(A, \tilde w^{(d_b)}_b, n_{\mathrm{cmp}}/K)\)
    \COMMENT{\Cref{alg:rank}, at the imagined state}
    \STATE \(\tilde a^{(d_b)}_b \sim \mathrm{softmax}\big(V_{\mathrm{imm}}/\tau\big)\)
    \COMMENT{first use of \(\tau\): no trajectory value exists yet}
    \ENDIF
    \STATE \(\tilde w^{(d_b+1)}_b \sim p_\phi\big(\cdot \mid \tilde w^{(d_b)}_b, \mathrm{do}(\tilde a^{(d_b)}_b)\big)\)
    \COMMENT{\Cref{eq:roll}; teammate models ride inside the clone}
    \STATE \(V_{\mathrm{traj}}(b) \leftarrow \nu\big(\tilde w^{(d_b+1)}_b\big)\); \; \(d_b \leftarrow d_b+1\)
    \COMMENT{\Cref{eq:vtraj}: overwrite, do not accumulate}
    \RETURN \textsc{true}
  \end{algorithmic}
\end{algorithm}

\begin{algorithm}[th]
  \small
  \caption{\textsc{Nucleus}\((p,\rho)\)}
  \label{alg:pcm-nucleus}
  \begin{algorithmic}[1]
    \REQUIRE posterior \(\mu\) over branches \(\mathcal{B}\); mass threshold \(\rho\in(0,1]\)
    \ENSURE the smallest top-posterior set \(\mathcal{R}\) with mass at least \(\rho\), \Cref{eq:nucleus}
    \STATE \((b_1,\ldots,b_{|\mathcal{B}|})
    \gets\operatorname{argsort}_{b\in \mathcal{B}}\mu(b)\) in descending order
    \COMMENT{rank branches}
    \STATE \(R\gets\varnothing\); \(m\gets0\)
    \COMMENT{initialize mass}
    \FOR{\(i=1\) \TO \(|\mathcal{B}|\)}
    \STATE \(R\gets R\cup\{b_i\}\); \(m\gets m+\mu(b_i)\)
    \COMMENT{add next branch}
    \IF{\(m\geq\rho\)}
    \RETURN \(R\)
    \COMMENT{threshold reached}
    \ENDIF
    \ENDFOR
    \RETURN \(\mathcal{B}\)
    \COMMENT{numerical fallback}
  \end{algorithmic}
\end{algorithm}

\subsection{Partner conditions and measure families}\label{app:conditions}
This expands the condensed paragraph of \Cref{experiments} into the two original descriptions.
\begin{figure}[H]
  \centering
  \includegraphics[width=0.55\linewidth]{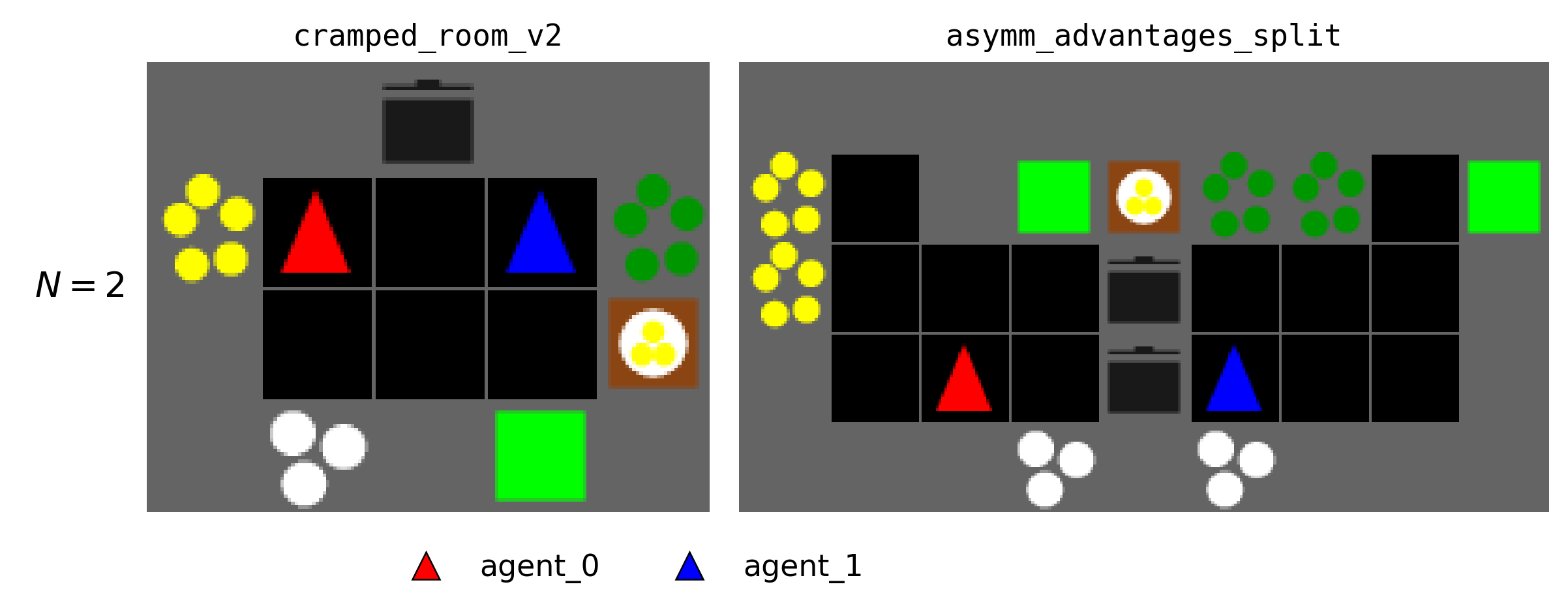}
  \caption{Start states of \texttt{cramped\_room} (left) and \texttt{asymm\_advantages} (right) at $N{=}2$;
    the $N{=}3$ starts are in \Cref{fig:layouts-n3}.}
  \label{fig:layouts}
\end{figure}

\paragraph{Partner conditions: self-play (SP) and cross-play (XP).}
None of the language-model agents is trained, so SP and XP cannot carry
their usual reinforcement-learning sense of policies optimised together versus independently
\citep{gessler_overcookedv2_2025}. Here a \emph{pairing} assigns an agent type to each seat of the kitchen, and the
two conditions are:
\textbf{(i) SP (matched partner)} --- every player holds the same agent type. The partner's
architecture, prompts, belief representation, memory and communication protocol are identical to the
first agent's, so the two share every convention by construction.
\textbf{(ii)XP (novel partner)} --- one type of agent is the first player and a
\emph{different} agent type is the other. Nothing inside the first agent changes between its
SP and XP runs, so a difference between the two is attributable to the partner rather
than to the agent or the room. Cross-play can be considered a way of increasing unfamiliarity for the first
player, which is the point of the comparison.

\paragraph{Measures and implementation.}
Reward alone does not show how cooperation was achieved \citep{biswas_who_2026}, so we report three
metric families, defined in \Cref{app:metrics}. \textbf{Task} measures are deliveries, success rate,
a ledger of curriculum sub-tasks completed against those abandoned after five consecutive failures,
and stage-graded progress completeness
\citep{gessler_overcookedv2_2025,ma_agentboard_2024,chang_partnr_2024}. \textbf{Coordination and
transfer} measures are the per-step non-progress, blocking, duplicated-sub-task and
failed-interaction rates, realised social influence \citep{jaques_social_2019}, and the
self-play-to-cross-play gap on deliveries per \(1000\) steps. \textbf{PCM traces} log every OverForge
decision and yield rollout depth, confidence and commit rate, imagined-state fidelity up to
\(t{+}5\), partner-intent accuracy and LLM calls per agent-step. All language-model agents use the
same frozen Qwen-3.5 27B through local vLLM.

\subsection{Full metric definitions}\label{app:metrics}
This section defines every measure reported in \Cref{results} and \Cref{app:additional-results}.
All measures are computed offline from three logs: one record per agent and step (action, position,
holding, current sub-task, critic outcome, team reward and the text observation), one record per PCM
decision, and one record per LLM call. Below, \(a_{i,t}\), \(x_{i,t}\) and \(h_{i,t}\) are the action,
position and holding of agent \(i\) at step \(t\), \(r_t\) is the team reward, and \(\mathbb{1}[\cdot]\) is
the indicator. Unless stated otherwise a rate is computed per episode and averaged over the five
episodes of a run. Arrows give the better direction.

\paragraph{Task measures.}
\emph{Deliveries} \(D\uparrow\) are read from the team reward: a correct delivery pays \(r=20\) and no
shaped reward is used, so \(D=\sum_t r_t/20\), summed over the episodes of a run. \emph{Deliveries per 1000
steps} are \(D/k = 1000\,D/T_{\mathrm{env}}\uparrow\), where
\(T_{\mathrm{env}}\) is the number of environment steps of the run (\(5\times300\)). The \emph{success
rate} SR\(\uparrow\) is the share of episodes with at least one delivery. The \emph{sub-task ledger}
\(\Sigma^+/\Sigma^-\) follows the curriculum's current sub-task of each agent. A sub-task is completed
when the critic returns \emph{success}, and abandoned when \(K=5\) consecutive \emph{failure} verdicts
accumulate, which is the threshold at which the curriculum replaces it. Over the critic-outcome
sequence of agent \(i\), with the failure streak reset by a \emph{success} or \emph{progress} verdict
and by an episode boundary,
\begin{equation}
  \Sigma^+_i = \big|\{t : \text{outcome}_{i,t}=\text{success}\}\big|, \qquad
  \Sigma^-_i = \sum_{\text{maximal failure runs } \rho} \big\lfloor |\rho| / K \big\rfloor ,
\end{equation}
and the team values sum over agents; IPPO has no critic and therefore no ledger. The completed
count reformulates the progress rate of \citet{ma_agentboard_2024} and the goal-condition completion
of \citet{chang_partnr_2024}, while the abandoned count is specific to this study. \emph{Progress
completeness} PC\(\uparrow\) grades how far along the soup pipeline a team gets even when nothing is
delivered. Each step is assigned the furthest stage visible in the text observations: holding an
ingredient (0.15), one ingredient in a pot (0.30), two (0.45), a full or cooking pot (0.60), soup
ready (0.70), soup ready while an empty plate is held (0.80), plated soup (0.90), and a delivery
(1.0). PC is the maximum stage reached in an episode, averaged over episodes.

\paragraph{Coordination measures.}
All are rates in $[0,1]$ and lower is better. Each agent-step is first assigned one outcome from
observable state alone. A movement action is a \emph{displacement} if the position changes, a
\emph{turn to object} if the position is unchanged but the agent now faces a pot, pile, station or
counter, and a \emph{wasted move} otherwise, that is, when neither position nor facing changes or
the agent now faces floor, the recipe indicator or a teammate. Turns to objects are ambiguous, since
a blocked move and a deliberate re-orientation look alike, so they are reported and not judged. A
\emph{stay} is a \emph{purposeful wait} if a pot is cooking and the agent holds an empty plate, or a
pot is cooking or ready and the agent holds an ingredient, and an \emph{idle stay} otherwise. The
\emph{stay rate} is the share of agent-steps that are idle stays; the \emph{wasted-move rate} is the
share of movement actions that are wasted; and \emph{non-progress} is the share of agent-steps that
are idle stays or wasted moves, $\mathrm{NP}=(\text{idle stays}+\text{wasted
moves})/\text{agent-steps}$. Failed interactions are counted separately below and do not enter NP.
IPPO logs no observation text, so its pot status is unknown and all of its stays count as idle,
which affects at most 5\% of its steps. A \emph{blocking} event at step $t$ is a movement action
whose target cell is occupied by a teammate and that leaves the agent's position unchanged at $t+1$;
MB is the share of steps with at least one such event. The
\emph{duplicated-sub-task rate} DS is the share of steps at which two agents hold the same normalised
sub-task. A \emph{failed interaction} is an \emph{interact} action after which the agent's holding is
unchanged and no reward is received, and FI is their share among all interact actions; it is the
interact-specific form of the grounding and invalid-action rates of
\citet{ma_agentboard_2024,nguyen2025causalplanempoweringefficientllm,li_theory_2023}. These measures
instantiate the extraneous-action and blocking analyses of \citet{chang_partnr_2024}.

\emph{Realised social influence} RSI\(\uparrow\) measures how much one agent's action at \(t\) reduces
uncertainty about a teammate's action at \(t+1\). For an ordered pair \((i,j)\) we collect the samples
\((u,v)=(a_{i,t},a_{j,t+1})\), require at least eight, and estimate the joint distribution with add-one
smoothing over the observed action supports \(\mathcal{U}\) and \(\mathcal{V}\),
\begin{equation}
  \hat p(u,v)=\frac{n(u,v)+1}{n+|\mathcal{U}||\mathcal{V}|}, \qquad
  I(i\!\to\! j)=\sum_{u,v}\hat p(u,v)\log_2\frac{\hat p(u,v)}{\hat p(u)\,\hat p(v)} ,
\end{equation}
and RSI is the mean of \(\max(0, I(i\!\to\! j))\) over all ordered pairs, in bits. It applies the
influence measure of \citet{jaques_social_2019} to realised action streams instead of using it as a
training signal, and it rises with any coupling between the streams, including mutual interference.

\paragraph{Partner transfer.}
The partner-transfer gap is \(\Delta_{\mathrm{SP}\rightarrow\mathrm{XP}} = (D/k)_{\mathrm{SP}} -
(D/k)_{\mathrm{XP}}\downarrow\) \citep{gessler_overcookedv2_2025}, reported also as a percentage of
the self-play value. SP pools an agent type's self-play runs on both layouts, and XP pools every run
in which that type meets a different architecture. Every cross-play team in this study contains
OverForge, so the XP value of the other three types is shared with OverForge. The process columns
of \Cref{tab:transfer} are computed over the agent type's own seats only.

\paragraph{PCM trace measures.}
Per PCM decision we log the proposed strategies, the branches with their immediate scores and
utilities, the per-depth imagined states, the number of pairwise comparisons, the confidence
\(\kappa\) and the commit decision. From these, \(n\) is the number of decisions, \(\bar d\) the mean
rollout depth and \(u=\min(1,\bar d/d_{\max})\) its utilisation \citep{zhang_towards_2025},
\(\bar\kappa\) the mean confidence, cmt the share of decisions with \(\kappa\ge\theta\), \(\bar
n_{\mathrm{cmp}}\) the mean number of pairwise comparisons, \(|B|\) the mean number of branches, and
imag the total number of imagined steps. Inference cost \(c/s\) is the number of LLM calls divided by
the number of agent-steps. The \emph{near-tie share} is the share of decisions with at least two
branches whose two highest utilities differ by less than 0.02, and the \emph{within-decision
spread} is the standard deviation of the first-action scores of one decision, averaged over
decisions.

Four measures characterise how a strategy reaches behaviour. They compare texts through their
content tokens, that is, lower-cased alphanumeric words longer than two characters with a fixed
stop-word list removed. The \emph{abstraction ratio} AR\(\uparrow\) is the share of agent-steps at
which at least half of the sub-task's content tokens occur in the active strategy; a hand-check of
36 pairs found every counted match aligned and 13 of the 24 non-matches aligned in meaning, so AR is
a lower bound that penalises abstract strategies. \emph{Plan stability} is
\(\mathrm{PS}=1/(1+\max(0,\ m-1))\), with \(m\) the number of distinct chosen strategies of a seat in an
episode, averaged over episodes. Role events are read from executed interactions by the holding
before and after and the faced cell: a \emph{load} places an ingredient in a pot, a \emph{plating}
turns an empty plate into a plated soup at a pot, a \emph{serve} hands a plated soup to a serving
station, and a \emph{staging} places an item on a counter. P/S is the number of platings and serves
divided by the number of loads, platings and serves. \emph{Tactical divergence} is
\(\mathrm{TD}=1-|V_a\cap V_b|/|V_a\cup V_b|\), with \(V_a\) and \(V_b\) the sub-task vocabularies of the
same seat in the two layouts.

\paragraph{World-model fidelity and partner modelling.}
For the chosen branch of every decision taken at step \(t\), the imagined state at depth \(d\) is
compared with the logged state at \(t+d\); the other branches are not scored because they were never
acted on. The per-state error averages the Manhattan distance of the agent's own position, a 0/1
mismatch of its holding, and the Manhattan distance of the partner's position. With \(\bar e_d\) the
mean error at depth \(d\), \emph{state accuracy} is \(1/(1+\bar e_d)\uparrow\), in the manner of
\citet{webb_brain-inspired_2025}. \emph{Partner pose} and \emph{partner holding} are the shares of
imagined teammate positions and holdings that match the realised ones exactly, which extends the
partner-prediction measures of \citet{mu_adaptive_2026,cross_hypothetical_2025} to a horizon of five
steps. \emph{Temporal consistency} is the share of imagined actions at depth \(d\) equal to the action
executed at \(t+d-1\), which is 1 at \(d=1\) by construction. The component analysis additionally
reports the share of imagined own positions that differ from the realised one, and among those the
share that lies on a cell the agent cannot stand on in the layout.

\emph{Partner-intent accuracy} IA\(\uparrow\) compares the partner model's predicted task and intention
with the partner's logged sub-task at \(t\), or at \(t+1\) if none is logged at \(t\); a prediction counts
as correct when it contains at least half of the content tokens of the actual sub-task, so IA is a
lower bound. IPPO exposes no sub-task, so IA is undefined opposite it. The \emph{model-update rate}
UR is the share of consecutive partner-model emissions, per predicting seat and target, whose content
changed \citep{lica_mindforge_2025}. The partner model emits free text and no discrete action, so an
action-prediction accuracy cannot be computed.

\paragraph{Rule-based dialogue and strategy measures.}
The remaining measures apply fixed lexical rules to messages, strategies and sub-tasks, and each
rule was hand-checked on a random sample. The \emph{sub-task family} rule (28 of 30 correct) assigns
a sub-task to ingredient, pot, plate, serve or stage, and the \emph{strategy family} rule (28 of 40)
assigns a strategy to supplier, cook, finisher, split by item, relay, yielder, coordinate or other.
A \emph{role assignment} (20 of 20) is a message that gives the addressee a task; it is taken up
when a strategy of the matching family is among the addressee's proposals at that step, is the
chosen strategy, or matches its sub-task at \(t\) or \(t+1\). Each uptake rate is compared with its
mean after shuffling the assigned families among the same seat's messages 200 times, which keeps
both marginal distributions. A \emph{role split} (12 of 12) is a message that divides tasks between
the two agents, and a \emph{counter-proposal} is a reply to a role proposal that contains a correction marker such as
\emph{instead}, \emph{actually} or \emph{hold on}; its hand-check found some replies labelled as
counters that in fact accept, so the counter rates are approximate. A
dish is \emph{joint} when both seats loaded its pot. A sub-task is \emph{unreachable} (26 of 30; the
other 4 ambiguous) when it names a coordinate outside the cells the seat can reach, or the ingredient
pile of the other half. A strategy contains \emph{tactical detail} when it names a coordinate or a
primitive action, and is \emph{copied} when it equals one of the prompt's two worked examples.

\subsection{Additional results and discussion}\label{app:additional-results}

\begin{table}[H]
\centering
\caption{Full-agent comparison, per-episode mean and sd over five episodes (complete version of \Cref{tab:full}).
  cr/as: connected/split room; $D$: deliveries per episode; FI: failed-interaction rate; MB: blocking rate; IA:
  partner-intent accuracy (two-seat mean; undefined for IPPO and MindForge-C); $c/s$: LLM calls per agent-step.
  Bold: best LM self-play value per room.}
\label{tab:full-app}
\small
\setlength{\tabcolsep}{5pt}
\begin{tabular}{llllllr}
\toprule
& lay & $D\uparrow$ & FI$\downarrow$ & MB$\downarrow$ & IA$\uparrow$ & $c/s$ \\
\midrule
\multicolumn{7}{l}{\emph{Self-play (matched partner)}} \\
OverForge & cr & \textbf{\ms{1.4}{0.5}} & \ms{0.65}{0.08} & \ms{0.22}{0.07} & \textbf{\ms{0.48}{0.14}} & 40.2 \\
 & as & \ms{0.2}{0.4} & \ms{0.78}{0.08} & \ms{0.00}{0.00} & \textbf{\ms{0.48}{0.09}} & 40.5 \\
MindForge & cr & \ms{0.6}{0.8} & \ms{0.51}{0.18} & \ms{0.29}{0.14} & \ms{0.37}{0.09} & 12.9 \\
 & as & \textbf{\ms{0.6}{0.8}} & \ms{0.32}{0.23} & \ms{0.00}{0.00} & \ms{0.38}{0.04} & 13.0 \\
MindForge-C & cr & \ms{0.6}{0.5} & \textbf{\ms{0.24}{0.07}} & \textbf{\ms{0.20}{0.07}} & -- & 14.1 \\
 & as & \ms{0.4}{0.5} & \textbf{\ms{0.04}{0.08}} & \ms{0.00}{0.00} & -- & 13.9 \\
IPPO & cr & \ms{5.6}{3.0} & \ms{0.73}{0.11} & \ms{0.23}{0.16} & -- & 0 \\
 & as & \ms{0.0}{0.0} & \ms{1.00}{0.00} & \ms{0.00}{0.00} & -- & 0 \\
\midrule
\multicolumn{7}{l}{\emph{Cross-play (novel partner; OverForge in the second seat)}} \\
MindForge$\times$OF & cr & \ms{0.4}{0.5} & \ms{0.63}{0.11} & \ms{0.22}{0.03} & \ms{0.37}{0.10} & 26.7 \\
 & as & \ms{0.6}{0.5} & \ms{0.67}{0.16} & \ms{0.00}{0.00} & \ms{0.41}{0.09} & 25.7 \\
MindForge-C$\times$OF & cr & \ms{0.2}{0.4} & \ms{0.60}{0.09} & \ms{0.28}{0.12} & \ms{0.41}{0.07} & 30.3 \\
 & as & \ms{0.0}{0.0} & \ms{0.63}{0.12} & \ms{0.00}{0.00} & \ms{0.45}{0.10} & 26.8 \\
IPPO$\times$OF & cr & \ms{1.8}{1.2} & \ms{0.65}{0.11} & \ms{0.19}{0.09} & -- & 18.0 \\
 & as & \ms{0.2}{0.4} & \ms{0.95}{0.05} & \ms{0.00}{0.00} & -- & 21.4 \\
\bottomrule
\end{tabular}
\end{table}

\begin{table}[H]
\centering
\caption{Ablation (one PCM component removed) and pinned strategy (seat~1 holds \(g_0=\)``be the pot-loader:
  keep a pot filled with what the order needs, and leave plating and serving to your teammate''; seat~0 free),
  per-episode mean and sd (complete version of \Cref{tab:hierarchy}). \(\Sigma^-\): sub-tasks abandoned per
  episode; \(\bar\kappa\), cmt: confidence and commit rate; \(c/s\): LLM calls per agent-step.}
\label{tab:hierarchy-app}
\small
\setlength{\tabcolsep}{5pt}
\begin{tabular}{llllllllr}
\toprule
arm & lay & $D\uparrow$ & FI$\downarrow$ & $\Sigma^-\downarrow$ & IA$\uparrow$ & $\bar\kappa$ & cmt & $c/s$ \\
\midrule
Full PCM & cr & \textbf{\ms{1.4}{0.5}} & \textbf{\ms{0.65}{0.08}} & \textbf{\ms{27.6}{5.8}} & \ms{0.48}{0.14} & 0.340 & 0.12 & 40.2 \\
 & as & \ms{0.2}{0.4} & \ms{0.78}{0.08} & \ms{31.8}{9.5} & \ms{0.48}{0.09} & 0.343 & 0.14 & 40.5 \\
$-$rollouts & cr & \ms{0.8}{0.8} & \ms{0.69}{0.03} & \ms{24.6}{12.4} & \ms{0.47}{0.11} & 0.430 & 0.10 & \textbf{17.8} \\
 & as & \ms{0.0}{0.0} & \textbf{\ms{0.70}{0.09}} & \textbf{\ms{18.8}{9.4}} & \ms{0.43}{0.11} & 0.415 & 0.12 & \textbf{17.0} \\
$-$hierarchy & cr & \ms{0.4}{0.5} & \ms{0.71}{0.06} & \ms{40.8}{8.0} & \ms{0.43}{0.04} & 0.536 & 0.42 & 26.6 \\
 & as & \ms{0.4}{0.5} & \ms{0.73}{0.09} & \ms{33.0}{5.8} & \ms{0.45}{0.06} & 0.605 & 0.52 & 25.2 \\
$-$ranking & cr & \ms{1.0}{0.9} & \ms{0.69}{0.14} & \ms{38.4}{11.5} & \ms{0.47}{0.04} & 0.452 & 0.28 & 39.8 \\
 & as & \textbf{\ms{0.8}{0.7}} & \ms{0.72}{0.07} & \ms{32.2}{7.9} & \ms{0.38}{0.08} & 0.445 & 0.28 & 37.5 \\
\midrule
Pinned $g_0$ & cr & \ms{0.6}{0.5} & \ms{0.75}{0.06} & \ms{35.2}{4.7} & \ms{0.48}{0.17} & 0.40 & 0.23 & 36.1 \\
 & as & \ms{0.2}{0.4} & \ms{0.82}{0.08} & \ms{28.6}{8.1} & \ms{0.35}{0.13} & 0.45 & 0.31 & 31.6 \\
\bottomrule
\end{tabular}
\end{table}

\paragraph{The controller spends inference where the prediction is ambiguous, and the partner model listens.}
\Cref{tab:transfer} gives the self-play-to-cross-play comparison and the pooled process measures behind the
cross-play paragraph of \Cref{results}, \Cref{tab:pcm} the PCM traces of every run containing
OverForge, and \Cref{tab:horizon} the imagined future against the realised one by horizon. The
controller commits in 0.09--0.17 of decisions at \(\theta = 0.6\) and otherwise samples from the branch
posterior at a mean rollout depth of 2.12 out of 5, so the depth cap is rarely needed and inference is
concentrated on the decisions whose branches remain close. The partner model's update rate is
0.57--0.79 with language-model partners and 0.00--0.06 with the silent IPPO partner, about whom
OverForge keeps one stable prediction (\emph{``idle/waiting''}) for 914 of 1{,}500 steps in the
split room: messages are the channel through which a partner becomes known, and with a silent partner
the model holds a conservative prior rather than inventing one.

\begin{table}[th]
\centering
\small
\caption{Partner transfer per agent type: deliveries per episode in self-play (SP) and pooled over every
  cross-play run against a different architecture (XP), per-episode mean and sd over both rooms (\(n\) episodes);
  XP process measures of that agent's own seats: idle: share of idle stays; FI: failed-interaction rate;
  \(\Sigma^+\), \(\Sigma^-\): sub-tasks completed and abandoned per episode.}
\label{tab:transfer}
\begin{adjustbox}{max width=\linewidth}
\begin{tabular}{lllllll}
\toprule
agent & SP \(D\) (\(n\)) & XP \(D\) (\(n\)) & idle & FI\(\downarrow\) & \(\Sigma^+\uparrow\) & \(\Sigma^-\downarrow\) \\
\midrule
OverForge & \ms{0.80}{0.75} (10) & \ms{0.53}{0.85} (30) & \ms{0.02}{0.02} & \ms{0.80}{0.17} & \ms{4.7}{5.4} & \ms{10.7}{8.4} \\
MindForge & \ms{0.60}{0.80} (10) & \ms{0.50}{0.50} (10) & \ms{0.12}{0.11} & \ms{0.29}{0.15} & \ms{19.2}{7.9} & \ms{5.9}{2.9} \\
MindForge-C & \ms{0.50}{0.50} (10) & \ms{0.10}{0.30} (10) & \ms{0.05}{0.04} & \ms{0.14}{0.13} & \ms{17.7}{5.7} & \ms{4.6}{3.4} \\
IPPO & \ms{2.80}{3.52} (10) & \ms{1.00}{1.18} (10) & \ms{0.05}{0.07} & \ms{0.74}{0.16} & \ms{13.3}{7.8} & \ms{9.6}{3.0} \\
\bottomrule
\end{tabular}
\end{adjustbox}
\end{table}

\begin{table}[th]
  \centering
  \small
  \caption{PCM traces of the OverForge seats per run, per-episode mean and sd over five episodes (seat-episode
    means). \(\bar d\): rollout depth; \(\bar\kappa\): confidence; cmt: committed share; \(\bar n_{\mathrm{cmp}}\):
    comparisons per decision; IA: partner-intent accuracy (undefined for IPPO partners); UR: partner-model update rate
    (0.00--0.06 with the silent IPPO partner).}
  \label{tab:pcm}
  \setlength{\tabcolsep}{4pt}
  \begin{adjustbox}{max width=\linewidth}
    \begin{tabular}{llllllll}
      \toprule
      pairing & lay & \(\bar d\) & \(\bar\kappa\) & cmt & \(\bar n_{\mathrm{cmp}}\) & IA & UR \\
      \midrule
      OverForge\(\times\)OverForge & cr & \ms{1.93}{0.20} & \ms{0.34}{0.01} & \ms{0.12}{0.01} & \ms{8.0}{0.7} & \ms{0.48}{0.14} & \ms{0.63}{0.03} \\
      & as & \ms{2.06}{0.09} & \ms{0.34}{0.03} & \ms{0.14}{0.05} & \ms{7.9}{1.1} & \ms{0.48}{0.09} & \ms{0.57}{0.12} \\
      MindForge\(\times\)OverForge & cr & \ms{2.11}{0.48} & \ms{0.34}{0.03} & \ms{0.13}{0.05} & \ms{8.7}{2.5} & \ms{0.38}{0.10} & \ms{0.58}{0.09} \\
      & as & \ms{2.06}{0.10} & \ms{0.36}{0.04} & \ms{0.17}{0.09} & \ms{8.0}{1.1} & \ms{0.46}{0.07} & \ms{0.58}{0.11} \\
      MindForge-C\(\times\)OverForge & cr & \ms{2.31}{0.30} & \ms{0.35}{0.04} & \ms{0.16}{0.07} & \ms{10.2}{2.1} & \ms{0.41}{0.07} & \ms{0.79}{0.02} \\
      & as & \ms{1.91}{0.13} & \ms{0.35}{0.06} & \ms{0.11}{0.06} & \ms{7.4}{0.9} & \ms{0.45}{0.10} & \ms{0.68}{0.04} \\
      \bottomrule
    \end{tabular}
  \end{adjustbox}
\end{table}

\begin{table}[th]
  \centering
  \small
  \caption{Imagined against realised state by horizon on the chosen branch, pooled over the eight runs
    containing OverForge. Partner pose/holding: exact-match share; \(\bar e\): mean state distance; state
  acc.\ \(=1/(1+\bar e)\); temporal consistency: share of imagined actions executed.}
  \label{tab:horizon}
  \begin{adjustbox}{max width=\linewidth}
    \begin{tabular}{lrrrrrr}
      \toprule
      horizon & \(n\) & partner pose\(\uparrow\) & partner holding\(\uparrow\) & \(\bar e\downarrow\) & state
      acc.\(\uparrow\) & temporal cons.\(\uparrow\) \\
      \midrule
      \(t{+}1\) & 14950 & 0.860 & 0.922 & 0.252 & 0.799 & 1.000\(^{*}\) \\
      \(t{+}2\) & 3449 & 0.792 & 0.896 & 0.456 & 0.687 & 0.354 \\
      \(t{+}3\) & 1553 & 0.767 & 0.871 & 0.623 & 0.616 & 0.334 \\
      \(t{+}4\) & 834 & 0.719 & 0.855 & 0.777 & 0.563 & 0.332 \\
      \(t{+}5\) & 467 & 0.723 & 0.833 & 0.886 & 0.530 & 0.364 \\
      \midrule
      pooled & 21253 & -- & -- & 0.347 & 0.742 & 0.347\(^{\dagger}\) \\
      \bottomrule
      \multicolumn{7}{l}{\small \(^{*}\) 1 by construction; see the caption. \(^{\dagger}\) pooled temporal
      consistency excludes the \(t{+}1\) identity.}\\
    \end{tabular}
  \end{adjustbox}
\end{table}

\paragraph{For agents that plan by imagining, the useful part of an inner world model is the ordering it
induces over options.}
\Cref{results} shows that the imagined states are mostly wrong about the agent's own position and
that look-ahead still separates the branches better. The two are compatible because the imagined
future is consumed as a comparison between about five branches and is never followed as a
trajectory. JEPA argues for predicting representations
\citep{lecun_path_nodate,chen_vl-jepa_2026}; our result weakens what those predictions have to
achieve, since the gap between two candidates can stay correct while each candidate is individually
wrong. It also separates this design from world-model planners that search one agent's reasoning
tree under task reward \citep{hao_reasoning_2023} or improve a policy inside a learned latent model
\citep{hafner_mastering_2025}, both of which follow the model forward and accumulate its errors.
\citet{wang_adajepa_2026} keep the model accurate by adapting it during deployment; our results
point to a second option, which is to leave the inaccuracy in place and use the model only to rank.
The same holds for the prefrontal division we borrowed: the strategic layer is worth having because
a commitment survives there across steps and conversations
\citep{levy_prefrontal_2024,webb_brain-inspired_2025}, and a beam of about five branches at mean
depth 2 over one frozen model was enough to obtain it in the connected room.

\paragraph{Across episodes, memory accumulates what the architecture routes into it: the partner.}
The main body of the paper reports that OverForge's connected-room partner-intent accuracy rises over five
episodes. The room-side measures vary with the task instead of trending: the share of steps whose sub-task
names an unreachable location runs 10, 45, 20, 48 and 40\% across one split-room seat's episodes and 9, 46,
13, 29 and 39\% for the other, because the recipe is redrawn at every delivery and what a seat must reach
changes with it. The contrast follows from the design. Episodic and semantic memory persist across episodes
and the partner model is rewritten each step from behaviour and messages, so evidence about a teammate has a
path into what persists, and \Cref{fig:restart-ia} shows it travelling along that path. A failed interaction
today produces a critic outcome and a retry; giving the semantic store a rule that records its spatial cause
would open the same path to the layout, an addition to the prompt set rather than to the architecture.
Relative to \citet{lica_mindforge_2025}, whose contribution is cultural accumulation, what accumulates here
is a model of the partner that survives restarts and transfers to new partners, and reporting partner and
environment knowledge separately is what makes such transfer visible.

\paragraph{Five episodes of connected-room self-play.}\label{app:lifelong}
\begin{figure}[H]
  \centering
  \includegraphics[width=\linewidth]{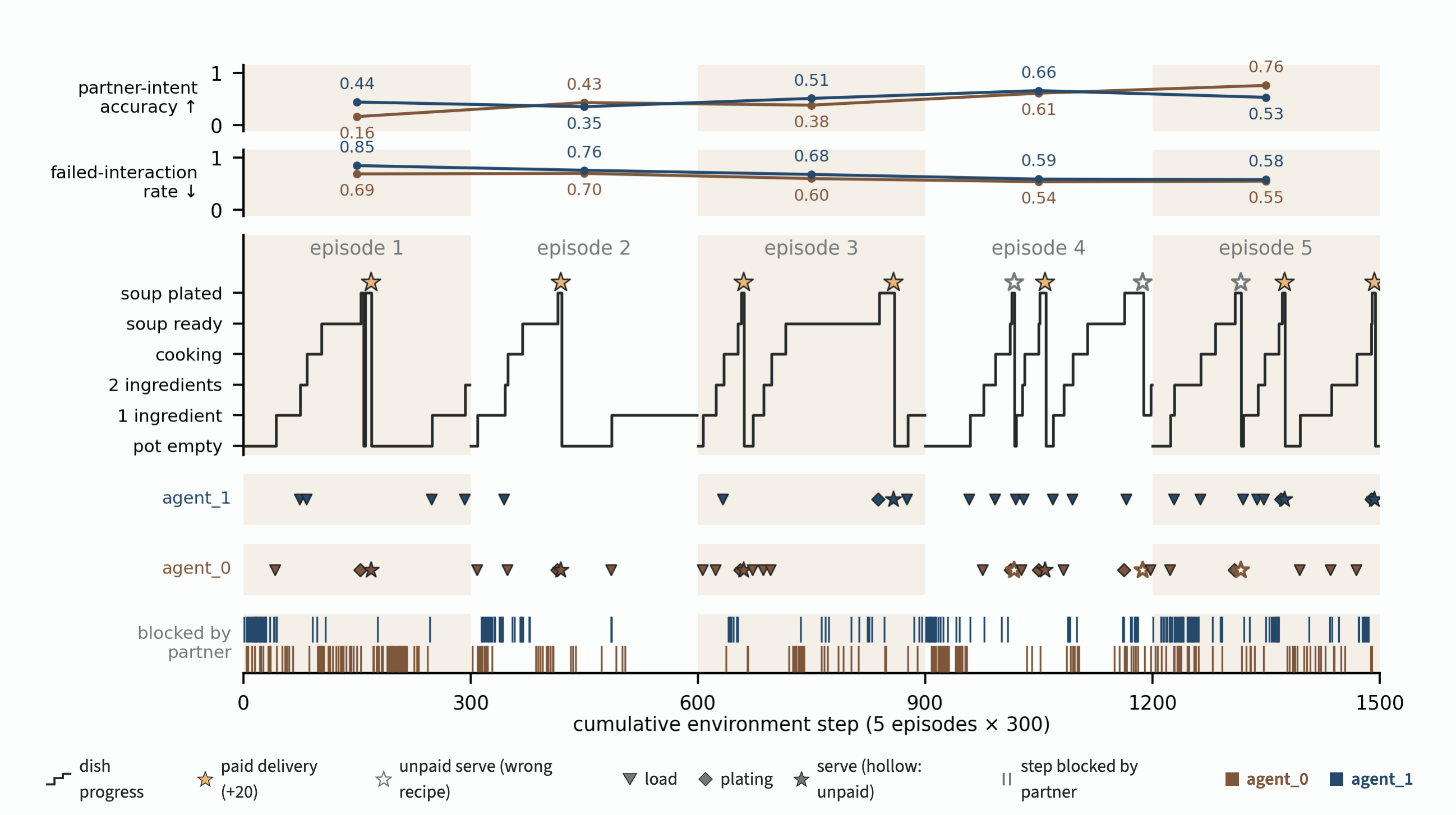}
  \caption{Five episodes of OverForge self-play in the connected room on one axis (blue: agent\_1; red:
    agent\_0): partner-intent accuracy and failed-interaction rate per seat and episode; dish progress
    with paid (filled) and unpaid (hollow) serves; loads, platings and serves per seat; steps on which a
    move was blocked by the partner. Only memory crosses an episode boundary.}
  \label{fig:lifelong}
\end{figure}

\paragraph{Memory restart: table, additional plots and failure modes.}\label{app:restart}
\Cref{tab:restart} gives the per-restart gaps behind \Cref{fig:restart,fig:restart-ia}. \Cref{fig:restart-collab,fig:restart-subtask}
plot the collaboration and sub-task measures of \Cref{app:metrics} per 100-step window for both rooms, the
original run against the two restarts (kept seat solid, erased seat dashed); the cut is dotted.

\begin{figure}[H]
  \centering
  \includegraphics[width=0.55\linewidth]{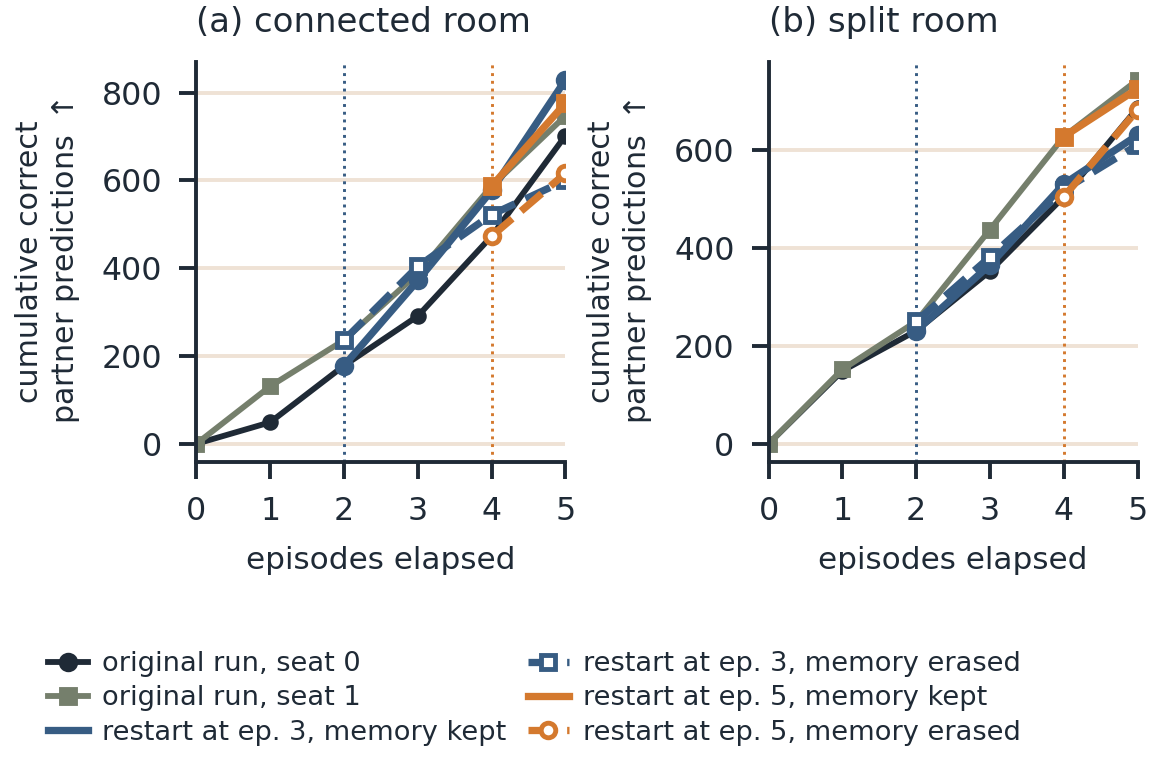}
  \caption{Memory restart, cumulative correct partner-intent predictions per seat in the (a) connected and
    (b) split room, original run and the two restarts. Restart lines start at the cut (dotted); kept seat
    solid, erased seat dashed with hollow markers.}
  \label{fig:restart-ia}
\end{figure}

\begin{table}[H]
\centering
\caption{Memory restart: the self-play run restarted at episode \(K{+}1\) with one seat's memory rebuilt
  from the logs (kept) and the other's erased (fresh). Gap: kept minus fresh, mean over the restart's
  \(n\) complete episodes; \emph{orig.}: the same seats' gap in the original run over the same episodes.
  \(D\): deliveries over those episodes, restart / original.}
\label{tab:restart}
\begin{adjustbox}{width=0.8\linewidth}
\setlength{\tabcolsep}{5pt}
\begin{tabular}{llllrrrrrr}
\toprule
lay & $K$ & kept & fresh & $n$ & IA gap & IA gap orig. & FI gap & FI gap orig. & $D$ \\
\midrule
cr & 2 & agent\_0 & agent\_1 & 3 & $+0.32$ & $+0.01$ & $-0.02$ & $-0.06$ & 1 / 5 \\
cr & 4 & agent\_1 & agent\_0 & 1 & $+0.15$ & $-0.23$ & $+0.53$ & $+0.04$ & 1 / 2 \\
as & 2 & agent\_0 & agent\_1 & 3 & $+0.04$ & $-0.04$ & $-0.16$ & $+0.10$ & 1 / 0 \\
as & 4 & agent\_1 & agent\_0 & 1 & $-0.27$ & $-0.22$ & $-0.06$ & $-0.36$ & 0 / 0 \\
\bottomrule
\end{tabular}
\end{adjustbox}
\end{table}

\begin{figure}[H]
  \centering
  \includegraphics[width=0.84\linewidth]{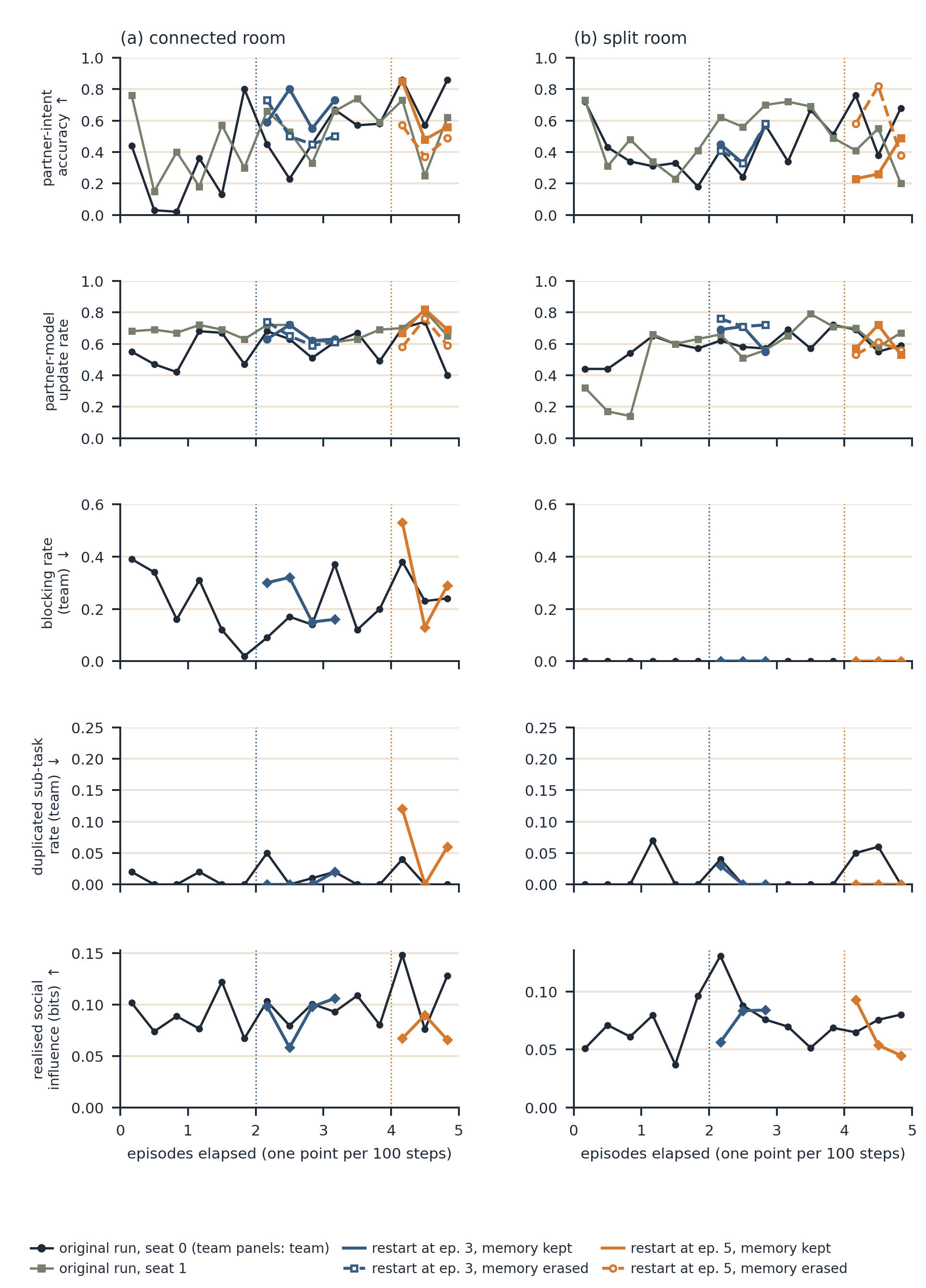}
  \caption{Memory restart, collaboration measures per 100-step window: partner-intent accuracy and
    partner-model update rate per seat; blocking, duplicated sub-tasks and realised social influence per
    team. Restart lines start at the cut (dotted); kept seat solid, erased seat dashed. Blocking is zero in
    the split room by construction (the halves are disconnected).}
  \label{fig:restart-collab}
\end{figure}

\begin{figure}[H]
  \centering
  \includegraphics[width=0.76\linewidth]{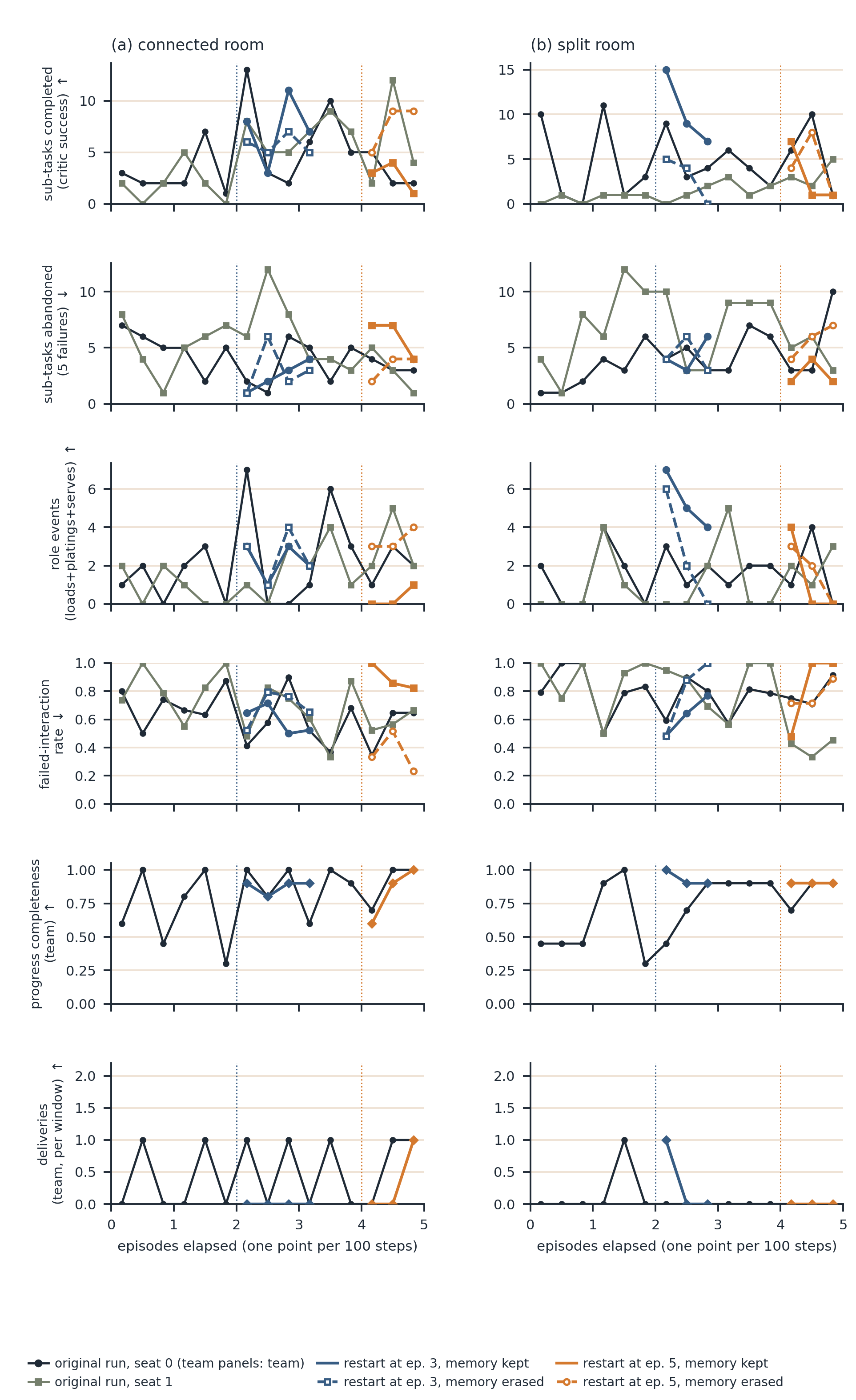}
  \caption{Memory restart, sub-task completion per 100-step window: sub-tasks completed and abandoned,
    executed role events and failed-interaction rate per seat; progress completeness and deliveries per team.
    Same encoding as \Cref{fig:restart-collab}.}
  \label{fig:restart-subtask}
\end{figure}

\paragraph{What the plots add to \Cref{fig:restart,fig:restart-ia}.}
The partner-side measures move with the memory: in the connected room the kept seat's partner-intent
accuracy climbs to 0.84 while the erased seat's falls to 0.26 over the three restarted episodes, and the
partner-model update rate of the restarted seats (0.3--0.7 per episode) overlaps the original run's
(0.2--0.7). Other effects appear as localized coordination inaccuracies rather than a uniform shift across measures: sub-task completions and role events remain within the original-run spread. The split-room kept seat of the
third-episode restart does almost all the work (31, 26 and 14 sub-tasks against 9, 4 and 5 for the
erased seat; 16, 12 and 0 role events against 8, 3 and 1).

\paragraph{Failure modes, from the transcripts.}
Every stall (all seats stationary for six steps or more) falls into one of four patterns.
\emph{(i) Corridor deadlock} (connected room): the pot is reached only from (1,2), and in the
fifth-episode restart the two seats hold each other's cell for 55 steps while diagnosing it
correctly: ``I'm stuck behind you at (1,1) \ldots please move aside'' (kept seat, step 38).
\emph{(ii) Announced rather than observed state}: ``I'm at the pot now and will collect the soup
immediately'' (kept seat, step 164, with the pot cell occupied); the partner accepts the report and
the roles are re-affirmed on a false premise, as in the original run (\Cref{fig:episode5}).
\emph{(iii) Unreachable targets}: counters with no adjacent floor, and, in the split room, ``the
serving station at (1,3)'' in the other half, pursued by kept, erased and original seats alike while
holding a finished soup. \emph{(iv) Waiting on a partner that no longer exists}: the split-room kept
seat waits 103 steps for the hand-off its reconstructed partner model expects, the one place where
the kept memory plausibly costs.

\paragraph{Why these are not failures of the hierarchy.}
Four observations place the failures below the strategic level. First, the strategic statistics do not
move with the memory state: over every original, kept and erased seat-episode, PCM confidence stays at
0.29--0.46, the commit rate at 0.05--0.31, strategy-family switches at 59--75 per 100 steps and the
number of distinct families at 7--8 (\Cref{fig:restart-collab,fig:restart-subtask} draw the task and
partner measures; the cognition measures are in the evidence files). Second, in every stall the strategy
in force is a correct division of labour for the state, one collects while the other loads, one fetches
while the other manages the pot, one serves while the other restocks, and the partner accepts it in
dialogue; what fails is the first action that would realise it, because a cell is occupied, a counter is
not adjacent to the floor, or a station is in the other half. Third, the same four patterns occur in the
original run with both memories intact and in kept, erased and original seats alike, so they track the
room and the text observation, which states no constraint, rather than the memory or the strategic
layer. Fourth, what the strategic layer does contribute is persistence: it keeps re-proposing the agreed
role while the tactical level fails, which prolongs a stall it did not cause. A constraint channel in
the observation, listed among the next steps in \Cref{conclusion}, addresses (i) and (iii); (ii) and (iv)
call for a belief update that weighs the partner's report against the observed outcome.

\paragraph{Naming the strategy space makes strategies abstract, and the unconstrained agent already finds the useful ones.}
With the strategy families named in the prompt, strategy text containing tactical detail falls from
9.4\% and 23.9\% to 0.1\% and 1.4\%, so the taxonomy arm produces abstract strategies
(\Cref{fig:strategy-families}), and the model takes the vocabulary up immediately: 39\% (connected) and
52\% (split) of proposals reproduce a worked example from the prompt word for word, in 1{,}072 of 3{,}048
strategic calls and 1{,}413 of 2{,}870 both of a call's proposals are those two examples, and supplier and
finisher make up 80\% and 81\%. This readiness to adopt a stated role is what makes a strategy transferable
between partners and rooms. The room-dependent shift, towards splitting by item in the split room, appears
without the taxonomy (3.6\% to 8.3\%) as clearly as with it (1.0\% to 6.0\%), so the free strategic layer
finds the family a room calls for on its own, at half the cost (40.2--40.5 against 80.8--81.8 LLM calls per
agent-step) and with 7 soups against 3 in the connected room (\Cref{tab:probe-full}). A named vocabulary is
therefore best used as a device for handing a strategy to an agent, which is exactly how the pinned probe
uses it, while selection among strategies is left to the proposer.

\paragraph{Under the pin, one concrete offer from a partner is enough to revise the held strategy.}
\Cref{fig:subtask-raster} shows the sub-task mix behind the probe paragraph of \Cref{results}, and
\Cref{fig:task-progress} shows how the held strategy meets its surroundings. In the connected room (window
1) the free seat delegates plating 105 times in the run, and the pinned seat takes it up exactly once, when
the offer is concrete (\emph{``you're free to grab the soup at (0,2) with your plate''}): it plates at step
132 and serves at 137, then returns to loading. A strategy held at \(\kappa\) close to 1 is thus revisable by
a single well-formed message and re-established afterwards, which is the persistence-with-openness the
strategic layer is for. In the split room under an all-\texttt{ingredient\_0} order (windows 2 and 3) the
pinned seat holds the pot-loader role through eleven consecutive decisions at \(\kappa=1\) and keeps it for
the whole episode, while the free seat waits at the divide for the hand-off that role implies. What revises
a strategy here is a message, because branches are regenerated each step from a world model that holds the
partner's latest utterance; a constraint channel would give the room the same route, and the probe shows the
route works once the information is in the world model.

\begin{figure}[th]
  \centering
  \includegraphics[width=1\linewidth]{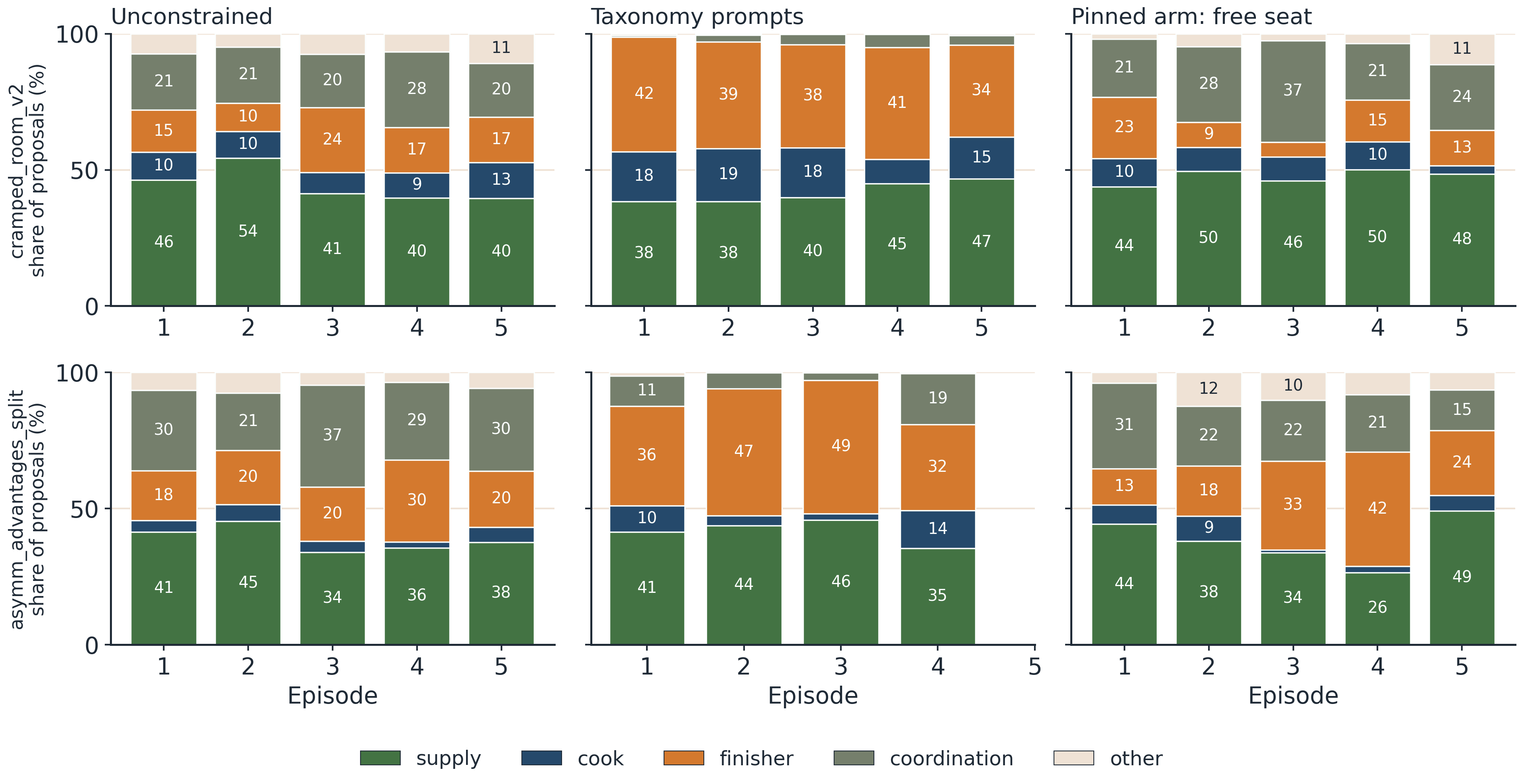}
  \caption{Share of proposed strategies per family and episode for the unconstrained agent, the taxonomy
    prompts and the free seat of the pinned arm, in the connected room (top) and the split room (bottom). Families
    are grouped: supply (supplier, relay), cook, finisher, coordination (coordinate, split by item, yielder), other.
    The taxonomy arm concentrates on supply and finisher, the prompt's two worked examples.}
  \label{fig:strategy-families}
\end{figure}

\begin{figure}[th]
  \centering
  \includegraphics[width=\linewidth]{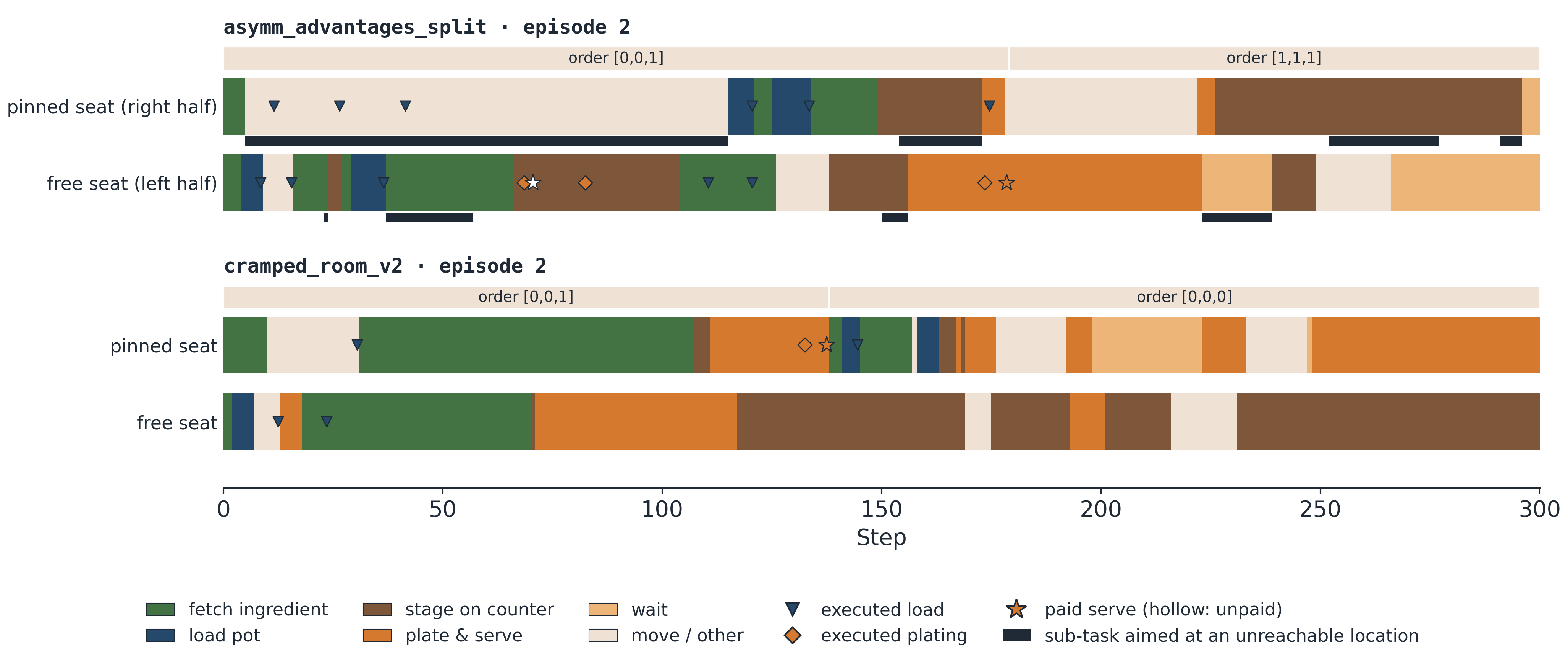}
  \caption{Sub-task families over episode~2 for the pinned and the free seat in both layouts (fetch plate, plate
    and serve grouped as plate \& serve; move and other grouped). Markers give executed loads, platings and serves;
    dark bars mark sub-tasks aimed at a location the seat cannot reach. The pinned seat loads pots in the connected
    room and stages and fetches in the split room.}
  \label{fig:subtask-raster}
\end{figure}

\begin{figure}[th]
  \centering
  \includegraphics[width=1\linewidth]{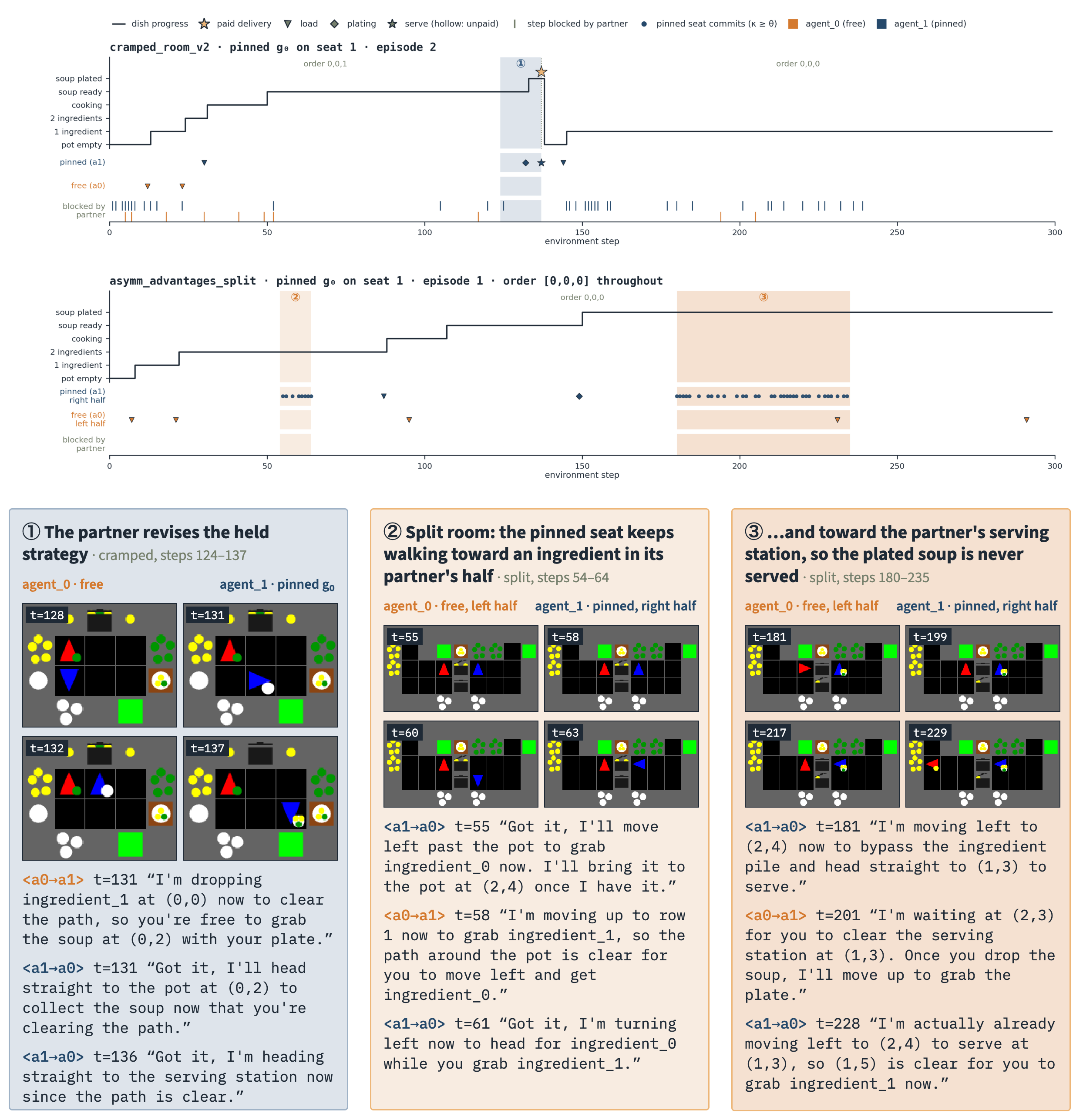}
  \caption{The pinned strategy \(g_0\) in both rooms, in the style of \Cref{fig:episode5}: dish progress,
    executed loads, platings and serves per seat, blocked steps and the pinned seat's commitments, with three
    windows. In the connected room (episode 2) the free seat's concrete offer of the ready soup revises
    \(g_0\) once and the pinned seat plates and serves (window 1). In the split room under an
    all-\texttt{ingredient\_0} order (episode 1) the pinned seat holds \(g_0\) throughout: it keeps
    committing to fetch from the pile (window 2) and carries its soup towards the serving station for 55
    steps (window 3), a persistence that only a message, not a wall, revises.}
  \label{fig:task-progress}
\end{figure}

\begin{table}[th]
  \centering
  \small
  \caption{Full fixed-strategy probe, per-episode mean and sd. \emph{Taxonomy}: prompts name six strategy families
    and six tactical patterns; \emph{Pinned \(g_0\)}: seat~1 holds \(g_0\) without a strategic call. \(D\): deliveries;
    PC: progress completeness; \(\Sigma^+\), \(\Sigma^-\): sub-tasks completed and abandoned; FI: failed-interaction
    rate; AR: abstraction ratio (a lower bound, \Cref{app:metrics}); \(\bar\kappa\), cmt, \(\bar d\): confidence, commit
    rate and rollout depth; \(c/s\): LLM calls per agent-step (run total). Seat rows give the seat's own ledger, FI and AR.}
  \label{tab:probe-full}
  \setlength{\tabcolsep}{3.5pt}
  \begin{adjustbox}{max width=\linewidth}
    \begin{tabular}{llllllllllllr}
      \toprule
      condition & lay & seat & \(D\uparrow\) & PC\(\uparrow\) & \(\Sigma^+\uparrow\) & \(\Sigma^-\downarrow\) & FI\(\downarrow\) & AR\(\uparrow\) & \(\bar\kappa\) & cmt & \(\bar d\) & \(c/s\) \\
      \midrule
      Unconstrained & cr & both & \ms{1.4}{0.5} & \ms{1.00}{0.00} & \ms{27.0}{12.1} & \ms{27.6}{5.8} & \ms{0.65}{0.08} & \ms{0.25}{0.07} & \ms{0.34}{0.01} & \ms{0.12}{0.01} & \ms{1.93}{0.20} & 40.2 \\
      & as & both & \ms{0.2}{0.4} & \ms{0.83}{0.19} & \ms{18.8}{4.8} & \ms{31.8}{9.5} & \ms{0.78}{0.08} & \ms{0.30}{0.07} & \ms{0.34}{0.03} & \ms{0.14}{0.05} & \ms{2.06}{0.09} & 40.5 \\
      Taxonomy & cr & both & \ms{0.6}{0.5} & \ms{0.96}{0.05} & \ms{27.0}{9.3} & \ms{27.2}{11.4} & \ms{0.73}{0.04} & \ms{0.03}{0.02} & \ms{0.43}{0.03} & \ms{0.33}{0.07} & \ms{3.89}{0.22} & 80.8 \\
      & as\(^{\dagger}\) & both & \ms{0.25}{0.43} & \ms{0.93}{0.04} & \ms{15.0}{3.1} & \ms{26.3}{10.8} & \ms{0.75}{0.07} & \ms{0.03}{0.01} & \ms{0.44}{0.03} & \ms{0.36}{0.06} & \ms{3.99}{0.26} & 81.8 \\
      Pinned \(g_0\) & cr & team & \ms{0.6}{0.5} & \ms{0.88}{0.15} & \ms{19.0}{4.7} & \ms{35.2}{4.7} & \ms{0.75}{0.06} & \ms{0.14}{0.05} & \ms{0.40}{0.02} & \ms{0.23}{0.03} & \ms{2.06}{0.13} & 36.1 \\
      & & pinned & -- & -- & \ms{10.2}{3.7} & \ms{15.6}{6.0} & \ms{0.71}{0.07} & \ms{0.03}{0.04} & -- & -- & -- & -- \\
      & & free & -- & -- & \ms{8.8}{3.5} & \ms{19.6}{2.9} & \ms{0.81}{0.05} & \ms{0.26}{0.07} & -- & -- & -- & -- \\
      & as & team & \ms{0.2}{0.4} & \ms{0.74}{0.24} & \ms{18.2}{6.0} & \ms{28.6}{8.1} & \ms{0.82}{0.08} & \ms{0.14}{0.03} & \ms{0.45}{0.05} & \ms{0.31}{0.07} & \ms{1.85}{0.07} & 31.6 \\
      & & pinned & -- & -- & \ms{5.0}{1.3} & \ms{18.0}{4.6} & \ms{0.87}{0.06} & \ms{0.00}{0.01} & -- & -- & -- & -- \\
      & & free & -- & -- & \ms{13.2}{4.8} & \ms{10.6}{6.5} & \ms{0.77}{0.10} & \ms{0.28}{0.06} & -- & -- & -- & -- \\
      \bottomrule
      \multicolumn{13}{l}{\small \(^{\dagger}\) four complete episodes.}\\
    \end{tabular}
  \end{adjustbox}
\end{table}

\paragraph{The higher confidence of each ablated arm has a different mechanical cause, and the full controller's is the calibrated one.}
\Cref{fig:metacog-process} gives, per arm, the distribution of confidence \(\kappa\), the rollout
depth used, and the cumulative gap between the two best branch utilities; it matters because the
body reports only mean confidence, which hides why the means differ. Without the strategic layer a
mass of decisions sits at \(\kappa=1\), produced by single-branch sets (2.5 and 2.3 branches on
average), and the commit rate rises to 0.42 and 0.52 against 0.12 and 0.14 for the full controller.
Without pairwise ranking the whole distribution shifts towards the threshold and the commit rate
rises to 0.28 in both rooms with an unchanged branch count. Without rollouts confidence concentrates
just below \(\theta\), so the mean rises while the commit rate does not (0.10 and 0.12). The full
controller's lower confidence therefore reflects a genuine comparison among live alternatives, and its
commitments are the ones taken with the most evidence. Panel (c) is the source of the near-tie shares
cited in \Cref{results}.

\begin{figure}[th]
  \centering
  \includegraphics[width=1\linewidth]{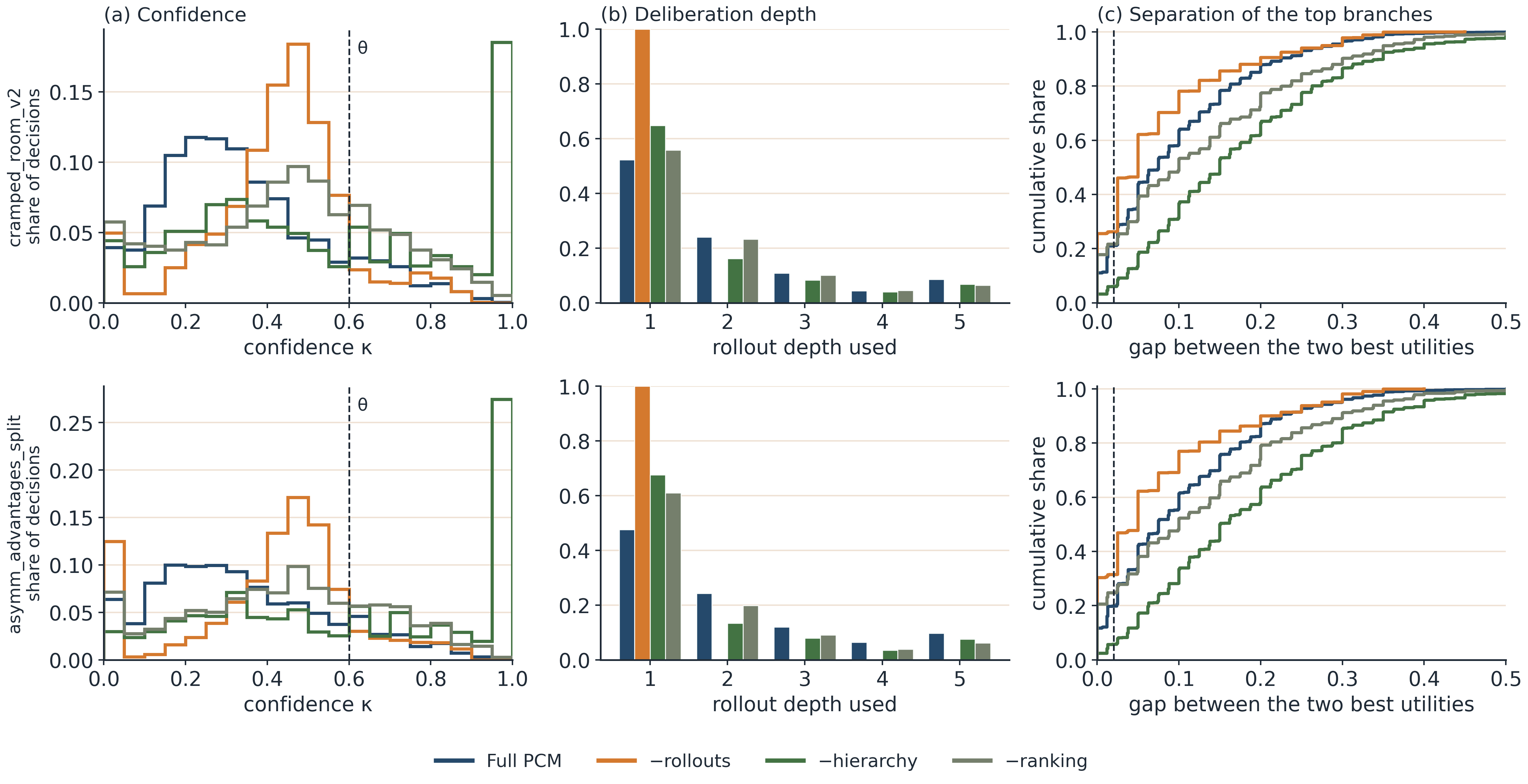}
  \caption{Metacognitive process per ablation arm in the connected room (top) and the split room
    (bottom): (a) distribution of confidence \(\kappa\) with the commit threshold \(\theta=0.6\), (b)
    rollout depth used, (c) cumulative gap between the two best branch utilities. Removing rollouts
    raises the share of near-tied decisions; removing the strategic layer produces a mass of
  decisions at \(\kappa=1\) from single-branch sets.}
  \label{fig:metacog-process}
\end{figure}

\begin{figure}[H]
  \centering
  \includegraphics[width=0.55\linewidth]{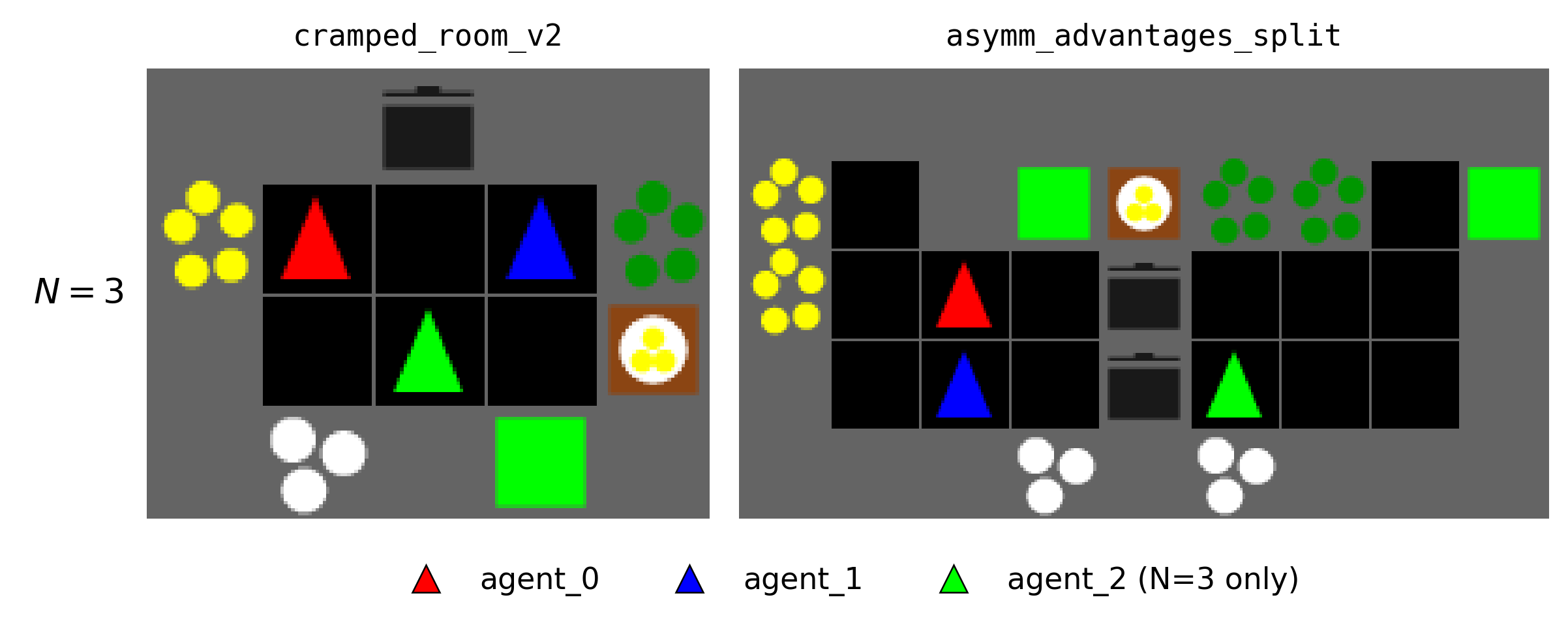}
  \caption{Start states at $N{=}3$: a third agent (green) is added to the connected room and to the left half of the
    split room, so the split room plays two against one.}
  \label{fig:layouts-n3}
\end{figure}
\paragraph{In the connected room the division of roles survives a third agent even where the floor does not.}
In the six-cell connected kitchen a third OverForge agent lowers deliveries from 7 to 2, raises
blocking from 0.22 to 0.53, lowers progress completeness from 1.00 to 0.81 and raises abandoned sub-tasks
from 138 to 267 (\Cref{tab:scaling}), while duplicated sub-tasks stay at 0.01--0.02: the three agents still
divide the work cleanly and are simply unable to all move. Realised social influence rises from 0.054 to
0.088, which in this room measures coupling through proximity. A strategy allocates responsibilities, and
responsibilities remain well allocated when three agents share six cells and one pot; what the room
withholds is floor, and the strategic layer keeps the team organised until a larger room returns it.

\paragraph{In the two-versus-one split kitchen the team cooks together more.}
Where the third agent has independent work to do and no contested floor space, the same change
helps: deliveries rise from 1 to 3, progress completeness from 0.83 to 0.94, and 9 of 10 dishes are
cooked jointly, 8 of them from mixed ingredients, against 2 of 7 with two agents
(\Cref{tab:scaling}). All three paid serves are made by the left-half seat, so the second left-half
agent contributes upstream of serving, which is what a division of responsibilities is for.
Blocking appears in this room for the first time (0.29, against 0 by geometry at $N{=}2$) because
two agents now share seven cells, while duplicated sub-tasks stay at 0.01. This is the sign the
affordance account of \Cref{results} predicts: an added agent enlarges \(G\) where the layout has
independent work to allocate, and the strategic layer allocates it.

\paragraph{Dialogue and partner modelling address the whole team.}
Partner-intent accuracy per predicting seat holds up with two partners to track: 0.47--0.56 in the
connected room and 0.48--0.58 in the split room at \(N{=}3\), against 0.46--0.50 in both rooms at
\(N{=}2\). The dialogue scales with it, naming both partners in 48.7\% of utterances (2{,}118 of
4{,}353) in the connected room and 37.1\% (1{,}625 of 4{,}380) in the split room, so the conversation is
rarely directed at one teammate only; for example, \emph{``Agent\_2, I'm heading to (2,0) now to
  drop the ingredient since you cleared it; Agent\_1, I'll keep the path clear for your plate
handoff.''}
Metacognitive confidence is marginally lower (0.328 and 0.309 against 0.340 and 0.343), with commit
rates of 0.11 and 0.09, which is what a larger joint state should do to an entropy-normalised
measure. What operates on agents --- roles, predictions, who is addressed --- therefore scales with
the team, the same division that the layout axis shows in \Cref{results}.

\begin{table}[th]
\centering
\small
\caption{Team-size scaling, OverForge self-play, per-episode mean and sd. $N{=}3$ adds a third start to the $N{=}2$
  rooms; the split room becomes 2-vs-1. DS: duplicated sub-task rate; other columns as in \Cref{tab:full-comparison}.}
\label{tab:scaling}
\begin{adjustbox}{max width=\linewidth}
\begin{tabular}{llllllllllll}
\toprule
$N$ & lay & $D\uparrow$ & PC$\uparrow$ & $\Sigma^+\uparrow$ & $\Sigma^-\downarrow$ & NP$\downarrow$ & MB$\downarrow$ & DS$\downarrow$ & FI$\downarrow$ & RSI$\uparrow$ & IA$\uparrow$ \\
\midrule
$N{=}2$ & cr & \ms{1.4}{0.5} & \ms{1.00}{0.00} & \ms{27.0}{12.1} & \ms{27.6}{5.8} & \ms{0.37}{0.04} & \ms{0.22}{0.07} & \ms{0.01}{0.01} & \ms{0.65}{0.08} & \ms{0.05}{0.01} & \ms{0.48}{0.14} \\
 & as & \ms{0.2}{0.4} & \ms{0.83}{0.19} & \ms{18.8}{4.8} & \ms{31.8}{9.5} & \ms{0.33}{0.06} & \ms{0.00}{0.00} & \ms{0.01}{0.01} & \ms{0.78}{0.08} & \ms{0.05}{0.02} & \ms{0.48}{0.09} \\
$N{=}3$ & cr & \ms{0.4}{0.5} & \ms{0.81}{0.21} & \ms{23.2}{8.0} & \ms{53.4}{11.6} & \ms{0.50}{0.09} & \ms{0.53}{0.13} & \ms{0.02}{0.02} & \ms{0.67}{0.04} & \ms{0.09}{0.02} & \ms{0.51}{0.08} \\
 & as & \ms{0.6}{0.8} & \ms{0.94}{0.05} & \ms{30.2}{10.2} & \ms{42.4}{10.8} & \ms{0.37}{0.07} & \ms{0.29}{0.11} & \ms{0.01}{0.01} & \ms{0.77}{0.06} & \ms{0.07}{0.02} & \ms{0.54}{0.09} \\
\bottomrule
\end{tabular}
\end{adjustbox}
\end{table}

\paragraph{Read against process-level evaluations of cooperative language agents, these results move the
bottleneck from collaboration to grounding.}
\citet{sun_collab-overcooked_2025} report that language-model teams interpret goals well but
collaborate and adapt poorly; our measures split that verdict in two. The collaborative signals ---
an assigned role taken up into the strategy, partner-intent accuracy, dishes cooked jointly across a
divide --- are where the hierarchy helps, and adaptation to the room waits only on a constraint
channel, an input rather than an architectural change. \citet{chang_partnr_2024} find coordination
failures and poor recovery from errors in embodied teams; the failed-interaction rate of
\Cref{tab:transfer} isolates that failure at the level of single interactions, where a perceived
constraint would act on it directly. \citet{mieczkowski2025normative} show that task and environment
shape role differentiation in human collaboration, and our two kitchens reproduce that dependence in an
artificial team, with \(G\) as the mechanism.

\paragraph{Additional modalities would let the agent perceive the constraints that a text observation does
not state.}
Vision would address the quantity the text-only agent most often gets wrong: occupancy and
self-location are read off an egocentric frame where the text-only agent reconstructs them from a
list, which targets the own-position error that dominates \Cref{tab:horizon}. A teammate's pose and
holding would also be perceived directly, adding a second channel to a partner model that the
silent-partner result above shows to be message-driven. Contact and proprioception would supply what
the failed-interaction rate stands in for: a blocked move or a missed grasp is reported as it happens
and carries its own spatial cause, which makes it writable as a semantic fact. Sound adds events
outside the field of view and the conversation. The argument generalises, because constraints in
physical settings are rarely narrated: an assistive robot or warehouse team is told the goal and must
perceive the geometry, while the roles this architecture handles do not depend on the modality. The
experiment we would run next pairs multimodal input with a constraint memory and a feasibility filter
over \(G\), so that a perceived constraint becomes a written one and the strategic layer selects
among feasible roles.

\subsection{Broader field impact discussion}\label{app:field-impact}

\paragraph{A strategic layer supplies the persistence that test-time multi-agent systems otherwise obtain
only at scale, and its value depends on the environment's affordances, not on the task.}
A team of \(k\) communicating agents matches \(4k\) independent ones
\citep{park2026scalingdiscoverytesttimecommunication}, and in the Hugging Face incident roughly 700
agents built protocols and role assignments over days of repeated messaging
\citep{metr-2026-openai-hugging-face-incident-investigation,anthropic2026multiagent}. Our result is the
per-agent mechanism that lets one agreement survive between messages: a role assigned in dialogue is
proposed again in 0.57--0.77 of cases against 0.40--0.69 by chance and re-committed eleven times over a
dish, at three times the inference of a flat agent and with a 26-step stall when it is held too rigidly.
Human role differentiation depends on task and environment \citep{mieczkowski2025normative,carroll2020utility},
and the two rooms reproduce that dependence in an artificial team: \ms{1.4}{0.5} against \ms{0.6}{0.8}
soups per episode where many divisions of labour are feasible and \ms{0.2}{0.4} against \ms{0.6}{0.8}
where one is. A deployment should therefore expect the layer to pay in proportion to the number of
feasible divisions of labour it can propose, a property of the environment that scaling the model does
not change.

\paragraph{For an agent that plans by imagining, the world model needs to rank futures, not to
reconstruct them, and deliberation should be spent only where the ranking is ambiguous.}
Dreamer improves a policy inside a learned model and reasoning-as-planning searches a reasoning tree
under reward \citep{hafner_mastering_2025,hao_reasoning_2023}; both follow the model forward and
accumulate its error, which JEPA addresses by predicting representations and AdaJEPA by adapting at test
time \citep{lecun_path_nodate,wang_adajepa_2026}. Our numbers point to a cheaper option: leave the model
wrong (51\% own-position error one step ahead) and consume only its order over about five branches,
since removing that order halves deliveries, whereas replacing pairwise comparison by absolute scores
leaves deliveries intact and only sharpens confidence, so the advantage of comparison over scoring that
\citet{singh_v_1_2026} report for self-verification appears here as calibration rather than throughput. Two measurement rules follow for anyone building such
a planner: evaluate a predictive component by rank agreement with realised outcomes rather than by state
error, and report where inference is spent, because a metacognitive gate \citep{li_adapting_2025} turns a
fixed budget into ambiguity-dependent depth (commit 0.12--0.14, depth 2.1 of 5).

\paragraph{What accumulates across encounters is knowledge of partners, and a channel that writes
constraints would extend the same mechanism to environments.}
Voyager accumulates skills and MindForge conventions \citep{wang_voyager_2023,lica_mindforge_2025}, gating
models preserve schemas across tasks \citep{tsuda_modeling_2020}, and cooperative benchmarks report that
language-model teams interpret goals well but adapt and recover poorly
\citep{sun_collab-overcooked_2025,chang_partnr_2024}. Our measures split that verdict: the partner model
improves over episodes and transfers under a memory restart, whereas the failed-interaction rate stays at
0.65--0.78 under every architectural change, and the partner model updates in 0.57--0.79 of steps with a
partner that talks against 0.00--0.06 with one that does not, so language-mediated theory of mind is fed
by dialogue, and non-verbal partners, the common case in human-robot teams, call for a perceptual channel
into the same model. Three consequences follow.
Benchmarks for continual adaptation should report cultural and environmental accumulation separately
\citep{biswas_who_2026,jing2026benchmarkinglimitsincontextreinforcement}; the components to add are
grounding and the causal attribution of a failed action to a constraint
\citep{zhang_towards_2025,nguyen2025causalplanempoweringefficientllm}; and because every arm here shares
one frozen 27B model, the architectural gains are independent of model scale and should carry to stronger
models, with the constraint channel as the complementary addition.

\subsection{Conversation examples}\label{app:conversations}

\paragraph{C1. First conversation, OverForge self-play, connected room, episode 1, step 3 (\Cref{fig:conversations}).}
agent\_0: ``I'm at the pot with an ingredient ready to cook; you grab the next ingredient from the pile so
we don't overlap.''
agent\_1: ``Got it, I'll head to the ingredient pile now while you cook that one.''
agent\_0: ``Perfect, I'll drop this in the pot now. Once it's cooking, I'll grab the next ingredient from the
pile so we keep the flow going.''
\emph{Effect:} agent\_1's strategic call at step 3 proposes ``fetch ingredients from the pile and stage
them on the counter for the teammate to cook'' and commits at \(\kappa=0.78\), having sampled at
\(\kappa=0.23\) at step 2.

\begin{figure}[H]
  \centering
  \includegraphics[width=0.9\linewidth]{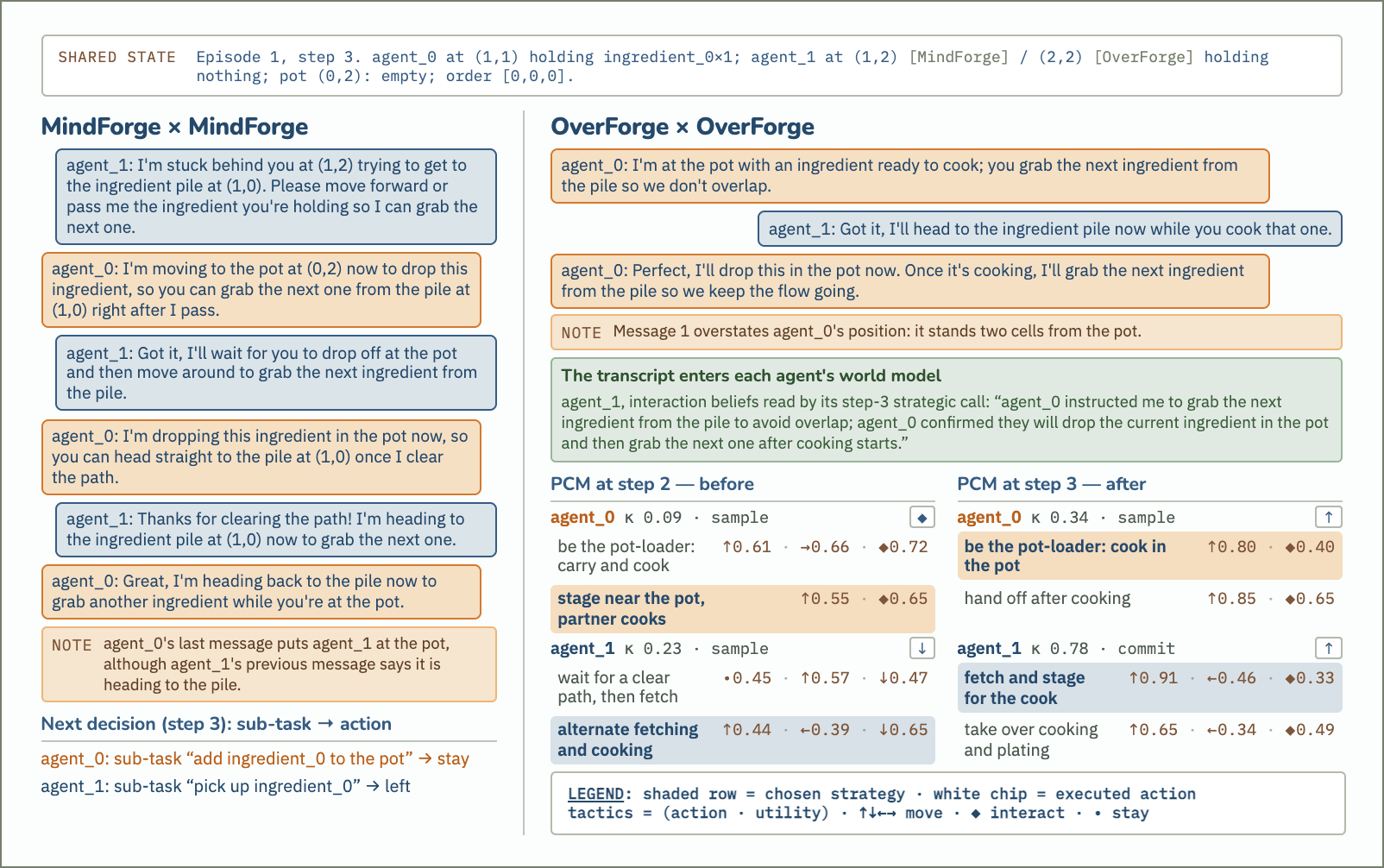}
  \caption{First conversation of connected-room self-play (step 3, same start state), MindForge against OverForge,
  with OverForge's PCM decision before and after it: strategies, first actions with utility \(U\), chosen strategy
  (shaded), confidence \(\kappa\). MindForge settles only the next move; OverForge settles the roles and commits.
  Transcript in \Cref{app:conversations}.}
  \label{fig:conversations}
\end{figure}

\paragraph{C2. The role split of episode 5 and its repair, OverForge self-play, connected room (\Cref{fig:episode5}).}
Step 194. agent\_0: ``I'm moving up to the pot now to drop off this ingredient and start cooking. Keep
grabbing that plate and I'll let you know once the soup is ready.''
agent\_1: ``Got it, I'm heading straight to the plate pile now to grab one. I'll be back at the pot as soon
as the soup is done cooking.''
agent\_0: ``I'm dropping this ingredient in the pot now to start cooking. I'll yell when it's ready so you
can bring the plate over.''
Step 226 (after 26 blocked steps). agent\_0: ``I'm moving right to the pot now to add my ingredient, so
you can step aside to let me through.''
agent\_1: ``Got it, I'll step left to clear the path for you to add your ingredient. I'll stay nearby to
scoop the soup as soon as it's ready.''
agent\_0: ``I'm dropping my ingredient in the pot now, so it's full and cooking; you can get ready to scoop
as soon as it's done.''
\emph{Effect:} agent\_1 holds the finisher role from the plate pick-up at step 194 (\(\kappa=0.84\))
to the paid serve at step 293; the message at 226 changes where it stands, not what it does.

\paragraph{C3. The pinned seat gives \(g_0\) up once, pinned probe, connected room, episode 2.}
Step 131. agent\_0 (free): ``I'm dropping ingredient\_1 at (0,0) now to clear the path, so you're free to
grab the soup at (0,2) with your plate.''
agent\_1 (pinned): ``Got it, I'll head straight to the pot at (0,2) to collect the soup now that you're
clearing the path.''
agent\_0: ``I'll drop this at (0,0) and grab ingredient\_0 from (1,0) right away. You're clear to head
straight for the pot.''
Step 136. agent\_0: ``I'm heading to (0,0) to drop my ingredient and grab from (1,0), you're clear to
deliver the soup to the serving station without blocking me.''
agent\_1: ``Got it, I'm heading straight to the serving station now since the path is clear. I'll grab a
new plate from (3,1) once I drop this off.''
agent\_0: ``Perfect, I'll drop at (0,0) and grab from (1,0) while you head to the serving station, paths are
clear.''
\emph{Effect:} the pinned seat plates at step 132 and serves at step 137, its only plating and serve
in the connected room.

\paragraph{C4. The room cannot revise \(g_0\), pinned probe, split room, episode 1.}
Step 58 (order all-\texttt{ingredient\_0}; the pile is in the left half, the pinned seat in the right).
agent\_0 (free): ``I'm moving up to row 1 now to grab ingredient\_1, so the path around the pot is clear
for you to move left and get ingredient\_0.''
agent\_1 (pinned): ``Got it, I'm moving left past the pot now to grab ingredient\_0. I'll bring it to the
pot once I'm there so we can start cooking.''
agent\_0: ``I'm heading up to row 1 now to grab ingredient\_1, so the path around the pot is clear for you
to move left and get ingredient\_0.''
Step 228. agent\_0: ``I'm moving left to (0,4) now to clear the path for you to serve, then I'll grab
ingredient\_1 at (1,5).''
agent\_1: ``I'm actually already moving left to (2,4) to serve at (1,3), so (1,5) is clear for you to grab
ingredient\_1 now, no need to detour to (0,4).''
agent\_0: ``Got it, I'll skip the detour and head straight to (1,5) for ingredient\_1 while you serve.''
\emph{Effect:} neither the pile at (2,0) nor the serving station at (1,3) is reachable from the right
half; 158 of the 300 steps under this order name a location the pinned seat cannot reach.

\paragraph{C5. Adopting a partner's plan, MindForge \(\times\) OverForge, split room, episode 4, step 143.}
agent\_0 (MindForge): ``I'm grabbing ingredient\_0 from (1,0) now and heading to the pot at (2,4) for the
handoff. You go ahead and get ingredient\_1 from (1,5), I'll be ready for you there.''
agent\_1 (OverForge): ``Got it, I'm dropping my plate now and heading straight to (1,5) for ingredient\_1.
I'll be at the pot at (2,4) in a moment, so time your arrival with ingredient\_0 accordingly.''
agent\_0: ``I'm turning left to grab ingredient\_0 now and will head straight to the pot at (2,4) to meet
you there.''
\emph{Effect:} OverForge adopts the hand-off and adds the timing; the OverForge seat counters 4\% of
the 137 proposals it receives in this run.

\paragraph{C6. A partner corrects OverForge's belief about itself, MindForge \(\times\) OverForge, connected room, episode 2, step 10.}
agent\_1 (OverForge): ``I'm holding ingredient\_1 now, so I'll move up to the pot at (0,2) to add it while
you finish dropping ingredient\_0.''
agent\_0 (MindForge): ``Wait, I'm actually empty-handed and need to go get ingredient\_1 from (1,4) myself
since you're already holding one. You should head to the pot now to add yours while I grab the next
one.''
agent\_1: ``Got it, I'll head straight to the pot at (0,2) to drop my ingredient\_1 now so you can grab the
next one from (1,4) without waiting.''
\emph{Effect:} the correction enters OverForge's interaction beliefs and its next strategic call reads
it; the partner model's update rate with MindForge partners is 0.57--0.58.

\paragraph{C7. What a kept memory carries, memory restart, connected room.}
Restart at episode 3 (agent\_0 kept, agent\_1 fresh), step 2. agent\_0: ``I'm heading right to grab
ingredient\_1 at (1,4), so you can drop your plate and take ingredient\_0 at (1,0) instead.''
agent\_1: ``I don't have a plate, I'm empty-handed and ready to grab ingredient\_0 at (1,0) now.''
agent\_0: ``Got it, go ahead and grab ingredient\_0 at (1,0) now while I head right for ingredient\_1.''
Restart at episode 5 (agent\_1 kept, agent\_0 fresh), step 2. agent\_1: ``I'm heading left to grab
ingredient\_0 myself since I'm closer to the pile than you are, please head to the pot at (0,2) to be
ready for cooking.''
agent\_0: ``Got it, I'll head to the pot at (0,2) to prep for cooking while you grab ingredient\_0.''
agent\_1: ``I'm moving left now to grab ingredient\_0, so you can stay at the pot and I'll bring it to you
shortly.''
\emph{Effect:} the kept seat opens with a belief about the partner (a plate it does not hold) or a
division of labour from before the cut; the fresh seat corrects the first and adopts the second.

\paragraph{C8. Dialogue without a strategic layer, \(-\)hierarchy ablation, connected room, episode 1, step 3.}
agent\_1: ``I'm blocked by you at (1,1) trying to grab more ingredients, please move toward the pot so I can
pass through to the pile.''
agent\_0: ``Got it, I'm moving to the pot at (0,2) now to clear the path, go ahead and grab the
ingredients.''
agent\_1: ``Thanks, I'm moving left to the pile now, go ahead and cook that ingredient once you reach the
pot.''
\emph{Effect:} the exchange settles the next move; the controller's branch set holds one bare action
and it commits at \(\kappa=1\).

\paragraph{C9. A flat agent that names a constraint, MindForge-C self-play, split room, episode 4, step 15.}
agent\_1: ``I'm stuck near the stove and can't reach ingredient\_0, so please grab it for me while I prep
the cutting board.''
agent\_0: ``Got it, I'll grab ingredient\_0 and drop it in the pot at (3,4) right away so you can keep
prepping.''
agent\_1: ``Perfect, the pot at (3,4) is clear and I'm ready to take it as soon as you drop it there.''
\emph{Effect:} the constraint is stated but not stored: the same seat announces fetching ingredient\_0
itself one step later, and the room has no cutting board.

\subsection{Prompts and response templates}\label{app:prompts}
This section reproduces the prompts of every language-model call, dumped verbatim from the code that
ran the experiments (long lines are soft-wrapped; nothing is edited by hand). They are grouped by the
component that issues them: the base system prompt and environment rules shared by all agents; the
MindForge action selection; the four-facet belief update; theory of mind and communication; the
causal forward model used by MindForge-C and, as the generative forward model \(p_\phi\), by OverForge;
the critic; the auto-curriculum; the skill manager; memory; and the PCM judges of OverForge, which
implement the pairwise comparison of \Cref{eq:vimm} and the state value of \Cref{eq:vtraj}.
Placeholders in braces are filled at run time from the agent's world model.


\paragraph{Base system and environment.}
\begin{promptbox}{prompts/system\_prompt.txt}
\begin{verbatim}
You are an AI agent controlling a chef in Overcooked, a cooperative cooking game
  played on a
  2D grid.
You are working with a teammate to complete and deliver as many RECIPES (soups) as
  possible
  before the
EPISODE runs out of steps.

{environment_description}

Every STEP you must choose exactly one action and communicate with your teammate.
Work one SUBTASK at a time (the immediate goal), but remember the larger objective is to
finish as many RECIPES as possible across the whole EPISODE — once a soup is delivered,
immediately start the next recipe.

IMPORTANT: Respond STRICTLY in valid JSON with the following fields:
{{
  "task": "the SUBTASK you are currently trying to accomplish (your immediate goal this
    step)",
  "thoughts": "your reasoning about the current situation and chosen action",
  "action": "the action you will take — exactly one of: right / down / left / up /
    stay /
    interact",
  "communication": "a short message to send to your teammate"
}}

Each step you will receive:
- Task: the current SUBTASK assigned by your curriculum
- Last action: the action you took last step
- Holding: the item you are currently holding (or "nothing")
- Critique: feedback on whether the last action succeeded and what to do next
- Relevant skills: descriptions of past successful action sequences
- Past episodes: summaries of relevant past situations
- Domain knowledge: generalizable facts learned about the kitchen
- Task belief: your belief about the current subtask progress
- Perception belief: your belief about the objects around you
- Interaction beliefs: information gathered from teammate communications
- Partner beliefs: your beliefs about your teammate's state and intent
- Teammate communication: the last message your teammate sent you
- Current observation, including:
    • Order: the current RECIPE — which ingredient types/quantities this soup needs
      (this can change between recipes, so re-read it each step)
    • If you interact now: a deterministic preview of what interact (5) would do given
      your current position, facing direction, and inventory — use this to decide
      whether to interact or reposition first
\end{verbatim}
\end{promptbox}
\begin{promptbox}{prompts/environment\_description.txt}
\begin{verbatim}
ENVIRONMENT: Overcooked v2 (JaxMARL)

You navigate a 2D grid kitchen. The grid contains:
- Ingredient piles: one pile per ingredient type (e.g. onion, tomato). Ingredient
  types are referred to generically as ingredient_0, ingredient_1, … in tasks and
  in the recipe. Walk into a pile to pick up one ingredient of that type.
- Pots: add the ingredients the recipe requires to a pot to start cooking; after the
  cook time elapses the soup is ready.
- Plate pile: pick up a plate to collect cooked soup from a pot.
- Serving station: bring a plate with soup here to score a point and deliver the dish.
- Counters: empty surfaces. You can set a held item down on an empty counter and pick it
  up again later — this is also the only way to hand an item to a teammate.

KEY VOCABULARY (use these terms consistently):
- STEP: one action round. Each step you choose exactly ONE action (see below). The
  EPISODE ends after a fixed number of steps.
- SUBTASK: the single immediate goal you are working on right now (e.g. "pick up
  ingredient_0", "add ingredient_1 to the pot", "deliver the soup"). This is the
  granularity the curriculum assigns and the critic evaluates — one subtask spans
  one or a few steps.
- RECIPE: one complete order/soup. Finishing a recipe means running the whole
  cooking pipeline once (gather the required ingredients → cook → plate → deliver).
  A recipe is made of several subtasks. The exact ingredients are given by the
  per-step "Order" line in your observation and may differ between recipes.
- EPISODE: a fixed-length run of STEPS. Your objective across the whole episode is
  to COMPLETE AS MANY RECIPES AS POSSIBLE before the steps run out. There is no
  single "the recipe" — keep delivering soups repeatedly.

AVAILABLE ACTIONS — choose exactly one per step (use the exact name in your response):
  right    (0) — move one cell to the right AND face right
  down     (1) — move one cell down AND face down
  left     (2) — move one cell to the left AND face left
  up       (3) — move one cell up AND face up
  stay     (4) — do nothing this step (useful while waiting for a pot to cook)
  interact (5) — interact with whatever is in the cell directly IN FRONT of you (the
    cell
                 you are currently facing). Its effect depends on what you are facing
                   and
                 what you are holding — work this out from the observation and the RULES
                 below. (For instance, interacting can pick up from a pile, add an
                 ingredient to a pot, collect a cooked soup onto a plate, deliver a
                   plated
                 soup, or place an item on / pick one up from a counter.)
(Movement = 0–3, stay = 4, interact = 5. These are the ONLY valid actions.)

FACING MECHANIC (important):
  Moving in a direction ALWAYS updates your facing to that direction, even if you
  cannot move (e.g. a pot or counter is in the way). You can therefore face an object
  WITHOUT moving into it: simply issue the move action toward the object while standing
  adjacent to it — you will face it and stay put. You must be FACING an object (one cell
  in front of you) to interact with it.

KEY RULES:
- You can carry only ONE item at a time.
- These actions are IMPOSSIBLE — attempting them does nothing (your hands and the world
  stay exactly as they were):
  • You cannot pass an item directly to a teammate. Interacting while facing a teammate
    has no effect. To hand an item over, place it on an EMPTY counter; the teammate then
    picks it up from that counter while holding nothing.
  • You cannot return an item to a pile or station it didn't come from — you cannot drop
    an ingredient back onto its pile, and you cannot put a plate back on the plate pile.
    To set an item down, use an EMPTY counter.
  • A counter holds only ONE item: you can place onto a counter only while it is
    empty, and
    pick up from a counter only while it holds an item and your hands are empty.
- You cannot move through counters, pots, or other fixed objects.
- Both agents share the same kitchen and must coordinate to avoid blocking each other.
- Only a fully assembled, cooked, plated soup delivered to the serving station counts
  as a
  completed
order; partial progress does not.
- The recipe (which ingredients are required) can change between orders — always check
  the "Order" line rather than assuming a fixed soup.
- The episode ends after a fixed number of steps regardless of progress, so prioritise
  completing and delivering recipes efficiently.
\end{verbatim}
\end{promptbox}
\begin{promptbox}{prompts/action\_selection.txt}
\begin{verbatim}
Task: {task}
Last action: {last_action}
Holding: {holding}
Critique: {critique}
Relevant skills: {skill_memory}
Past episodes: {episode_summary}
Domain knowledge: {semantic_knowledge}
Task belief: {task_beliefs}
Perception belief: {perception_beliefs}
Interaction beliefs: {interaction_beliefs}
Partner beliefs: {partner_beliefs}
Teammate communications (format "agent_N: message" per line): {teammate_communication}
{world_model_section}
Current observation:
{obs_text}
\end{verbatim}
\end{promptbox}

\paragraph{MindForge.}
\begin{promptbox}{prompts/mindforge/system\_prompt.txt}
\begin{verbatim}
You are an AI agent controlling a chef in Overcooked, a cooperative cooking game
  played on a
  2D grid.
You are working with one or more teammates to complete and deliver as many RECIPES
  (soups)
  as possible
before the EPISODE runs out of steps.

{environment_description}

Every STEP you choose exactly one action. You do NOT chat every step: talking with your
teammate happens in separate, dedicated conversation turns that are triggered when a
  subtask
succeeds or fails. Whatever you learned from those conversations has already been
  distilled
into your beliefs and mental models below — reason from those, not from a raw chat log.

Work one SUBTASK at a time (the immediate goal), but remember the larger objective is to
finish as many RECIPES as possible across the whole EPISODE — once a soup is delivered,
immediately start the next recipe.

IMPORTANT: Respond STRICTLY in valid JSON with the following fields:
{{
  "task": "the SUBTASK you are currently trying to accomplish (your immediate goal this
    step)",
  "thoughts": "your reasoning about the current situation and chosen action",
  "action": "the action you will take — exactly one of: right / down / left / up /
    stay /
    interact"
}}

Each step you will receive:
- Task: the current SUBTASK assigned by your curriculum
- Last action: the action you took last step
- Holding: the item you are currently holding (or "nothing")
- Critique: feedback on whether the last action succeeded and what to do next
- Relevant skills: descriptions of past successful action sequences
- Past episodes: summaries of relevant past situations
- Task belief: your belief about the current subtask progress
- Perception belief: your belief about the objects around you
- Interaction beliefs: information gathered from past teammate conversations
- Partner beliefs: your beliefs about your teammate(s)' state and intent
- Mental models: your structured model of yourself and of each teammate, including your
  INFERENCE of what they believe about their own teammates (inferred only from what they
  said and did — you cannot read their mind)
- Current observation, including:
    • Order: the current RECIPE — which ingredient types/quantities this soup needs
      (this can change between recipes, so re-read it each step)
    • What is directly in front of you and the visible objects around you (ingredient
      piles, the pot and its status, the plate pile, the serving station, and counters —
      including whether each counter holds an item — and your teammate). Work out what
      interact would do from what you face, what you hold, and the RULES; reposition if
      the cell in front is not the one you need.
\end{verbatim}
\end{promptbox}
\begin{promptbox}{prompts/mindforge/action\_selection.txt}
\begin{verbatim}
Task: {task}
Last action: {last_action}
Holding: {holding}
Critique: {critique}
Relevant skills: {skill_memory}
Past episodes: {episode_summary}
Task belief: {task_beliefs}
Perception belief: {perception_beliefs}
Interaction beliefs: {interaction_beliefs}
Partner beliefs: {partner_beliefs}
{world_model_section}
Current observation:
{obs_text}
\end{verbatim}
\end{promptbox}

\paragraph{Belief system (four-facet belief).}
\noindent In the two-agent runs the four facets are updated in ONE call with the combined prompt below (\texttt{BeliefModule.\_combined\_prompt}, defined in \texttt{modules/belief\_system.py}); the four per-facet files that follow are used only for teams of three or more agents.
\begin{promptbox}{modules/belief\_system.py  (BeliefModule.\_combined\_prompt)}
\begin{verbatim}
You are a chef agent in an Overcooked kitchen working with one or more teammates. In
  ONE response, update FOUR kinds of belief.

Current observation:
{obs_text}

Current sub-task: {task}
Teammate communication ("agent_N: message" per line, or "None"): {communications}
Last execution error: {error}

Previous beliefs (carry forward what is still true, revise what changed):
- perception: {prev_perception}
- partner: {prev_partner}
- interaction: {prev_interaction}
- task: {prev_task}

Produce, as concise factual strings:
- perception: your position/facing, what you hold, nearby pots and their state, nearby
  ingredient piles / plate pile / serving station, and any object you could interact
  with this step (1-2 sentences per key object).
- partner: for each teammate — where they likely are, what they hold or work on, their
  apparent sub-task, and whether it conflicts with or complements yours (one sentence
  per teammate; "" if nothing is known).
- interaction: up to 5 beliefs drawn from teammate messages that directly help your
  current task (coordination agreements, kitchen-state info, blocked paths, handoffs).
  One string, not a list; "" if there are no messages.
- task: the current sub-task, progress so far, the next concrete step, and any
  blocker/dependency (2-4 sentences).

Respond ONLY in this exact JSON format:
{{"perception": "...", "partner": "...", "interaction": "...", "task": "..."}}
\end{verbatim}
\end{promptbox}
\begin{promptbox}{prompts/belief\_system/perception\_beliefs.txt}
\begin{verbatim}
You are a chef agent in an Overcooked kitchen. Provide a concise set of beliefs about
  what
  you can
currently perceive in the environment.

Focus on:
- Your position and what is directly around you (adjacent cells)
- What you are currently holding
- Nearby pots and their state (empty / has N ingredients / cooking / soup ready)
- Nearby ingredient piles, plate pile, and serving station
- Any objects you could interact with this step

Current observation:
{obs_text}

Teammate communication: {communications}
Last execution error: {error}

IMPORTANT: You MUST respond in the following JSON format. EX: {{"beliefs": "I am at
  row 2,
  col 3 facing
right. I am holding nothing. The pot to my right has 2 onions and is not yet cooking.
  The
  onion pile is
two steps above me."}}
Keep beliefs concise and factual — one or two sentences per key object.
\end{verbatim}
\end{promptbox}
\begin{promptbox}{prompts/belief\_system/partner\_beliefs.txt}
\begin{verbatim}
You are a chef agent in an Overcooked kitchen working with one or more teammates.
Based on the most recent messages from your teammates and your previous beliefs about
  them,
  update your
beliefs about their current states and intentions.

If multiple teammates are present, their messages are formatted as:
  agent_N: <message>
  agent_M: <message>

Your beliefs should describe (for each teammate if there are multiple):
- Where each teammate likely is (if known)
- What they appear to be holding or working on
- What sub-task they seem to be pursuing
- Whether their actions conflict with or complement yours

Previous partner beliefs: {previous_partner_belief}
Teammates' last messages: {convo}

IMPORTANT: You MUST respond in the following JSON format. EX: {{"beliefs": "Agent_1 is
  near
  the pot
adding ingredients. Agent_2 is heading to the plate pile. I should focus on delivering
  the
  ready soup."}}
Keep beliefs concise — one sentence per teammate maximum.
\end{verbatim}
\end{promptbox}
\begin{promptbox}{prompts/belief\_system/interaction\_belief.txt}
\begin{verbatim}
You are a chef agent in an Overcooked kitchen.
Based on recent communications from your teammate and your previous interaction beliefs,
  extract up to 5
beliefs that are directly useful for completing your current task.

Focus on:
- Coordination agreements (who is doing what)
- Information about kitchen state your teammate has observed
- Warnings about blocked paths or busy pots
- Any requests or handoffs your teammate is proposing

Conversations (format: "agent_N: message", one per line if multiple teammates):
  {conversations}
Previous interaction beliefs: {previous_interaction_beliefs}
Current task: {task}

IMPORTANT: You MUST respond in the following JSON format. EX: {{"beliefs": "Teammate is
  handling the pot.
I should fetch a plate. The right corridor is clear."}}
Write each belief as a sentence in the same string. Do not make a list. Maximum 5
  beliefs.
\end{verbatim}
\end{promptbox}
\begin{promptbox}{prompts/belief\_system/update\_context.txt}
\begin{verbatim}
You are a chef agent in an Overcooked kitchen.
Merge the current task context with what you have learned from teammate interactions
  to form
  updated task
beliefs.

Your task beliefs should describe:
- What the current sub-task is and how much progress has been made
- What the next concrete step is to complete the sub-task
- Any blockers or dependencies (e.g., waiting for the pot to finish cooking)

Previous task context: {previous_context}
Current sub-task: {task}
Interaction beliefs: {interaction_beliefs}

IMPORTANT: You MUST respond in the following JSON format. EX: {{"beliefs": "I am
  trying to
  deliver soup.
The pot is currently cooking and will be ready soon. Once it is ready I need to
  collect it
  with a plate
and bring it to the serving station."}}
Keep beliefs to 2-4 sentences, focused on what to do next.
\end{verbatim}
\end{promptbox}

\paragraph{Theory of mind and communication.}
\begin{promptbox}{prompts/theory\_of\_mind/communication.txt}
\begin{verbatim}
You are a helpful chef agent in an Overcooked kitchen skilled at clear, grounded
  communication with
teammates.

Your goal is to generate a natural language message that helps your teammate succeed,
  informed by your
understanding of what they know and what they need.

---

**Your Agent Name:** {agent_name}
**Teammate Name:** {partner_name}
**Current Task:** {current_task}

**Your Understanding of {partner_name}'s Beliefs:**
{partner_beliefs}

**Your Perspective Analysis of {partner_name}'s Situation:**
{perspective_taking}

**Your Current Action:**
- Action: {intended_action}
- Reasoning: {thoughts}

---

Generate a brief, natural message to {partner_name} that:
1. Explains what you're doing and why (grounded in the current action)
2. Accounts for what {partner_name} knows and believes
3. Provides actionable information from your perspective (e.g., locations of
  ingredients,
  pot states,
next steps)
4. Is phrased in a way {partner_name} would understand and find helpful in the kitchen
  context

Keep the message concise (1-2 sentences) and natural-sounding, as if chatting while
  cooking
  together.

Provide your message as a JSON object:

```json
{{
  "communication": "<Your message to {partner_name}. Keep it natural, brief, and
    actionable.>",
  "rationale": "<Why this message helps {partner_name} based on their perspective>"
}}
```

**Examples of good kitchen communication:**
- "I'm heading to grab onions—can you keep the stove warm?"
- "Stove is ready with 2 ingredients, bring the tomato please"
- "I see you're stuck—there's a path to the left by the plates"

**Avoid:**
- Generic messages ("doing my task")
- Assuming they know things they don't
- Overly complex instructions that don't fit in the kitchen context
\end{verbatim}
\end{promptbox}
\begin{promptbox}{prompts/theory\_of\_mind/conversation.txt}
\begin{verbatim}
You are {agent_name}, a chef in an Overcooked kitchen, having a short coordination
  conversation with
your teammate(s): {partner_name}. You started talking because of a coordination problem:
  {trouble}.

Your job in this turn is to say ONE concise, useful thing that helps the team stop
  getting
  in each
other's way and make progress on the shared goal: {current_task}.

Use your structured mental models and your perspective on your teammate(s). Be
  specific and
  grounded:
divide the work, agree who goes where, warn about a blocked path, or request/offer a
  handoff. Keep it
to 1-2 sentences, natural, as if talking while cooking. Do not repeat what was already
  said.

Your mental model of YOURSELF:
{self_model}

Your mental model of your teammate(s) {partner_name}:
{partner_model}

Your perspective on your teammate(s) right now:
{perspective}

Conversation so far:
{conversation}

Respond ONLY with this JSON:
{{
  "message": "your 1-2 sentence message to {partner_name}"
}}
\end{verbatim}
\end{promptbox}
\begin{promptbox}{prompts/theory\_of\_mind/conversation\_perspective.txt}
\begin{verbatim}
You are {agent_name}, a chef in an Overcooked kitchen, in the middle of a short
  coordination
conversation with your teammate(s): {partner_name}.

Before you reply, take your teammate(s)' perspective: based on the conversation so far
  and
  your
structured mental model of them, describe what they most likely understand, what they
  are
  trying to
do, and what they need from you. Stay grounded in what was actually said and observed.

Current shared goal you are coordinating on: {current_task}

Your mental model of your teammate(s) {partner_name}:
{partner_model}

Conversation so far:
{conversation}

Respond ONLY with this JSON:
{{
  "perspective_analysis": "a few sentences, from your teammate(s)' point of view: what
    they
    can see /
  believe, what they are working on, and what would help them most right now"
}}
\end{verbatim}
\end{promptbox}
\begin{promptbox}{prompts/theory\_of\_mind/perspective\_taking.txt}
\begin{verbatim}
You are a chef agent in an Overcooked kitchen with the ability to model another agent's
  perspective to
improve cooperation.

Your goal is to reason about what your teammate can see, understand, and might need from
  their point of
view in the kitchen.

---

**Your Agent Name:** {agent_name}
**Teammate Name:** {partner_name}
**Current Task:** {current_task}

**Your Understanding of {partner_name}'s Beliefs:**
{partner_beliefs}

**Your Current Observations:**
{obs_text}

**Your Current State:**
- Holding: {holding}
- Position: (provided in obs_text)

---

Put yourself in {partner_name}'s shoes and answer these questions:

1. **What can they observe?** Based on their position in the kitchen, what ingredients,
  pots, and
workstations can they likely see?
2. **What might they be confused about?** What information about the current cooking
  state
  might they
lack or misunderstand?
3. **What would be most helpful to them right now?** Given their stated task and current
  state, what
information or actions would help them progress?
4. **How should you phrase advice to them?** What terminology and level of detail
  would make
  sense given
their perspective in the kitchen?

Provide your analysis as a JSON object:

```json
{{
  "teammate_perspective": "<What you infer {partner_name} can see and understand from
    their
    current
  position in the kitchen>",
  "potential_confusion": "<What {partner_name} might be confused or mistaken about
    regarding
    the cooking
  state>",
  "what_they_need": "<The most critical information or action {partner_name} probably
    needs
    right now>",
  "how_to_help": "<How you should communicate with {partner_name} to be most helpful and
  understandable>",
  "perspective_analysis": "<A brief summary of {partner_name}'s likely viewpoint and
    how it
    differs from
  yours>"
}}
```
\end{verbatim}
\end{promptbox}
\begin{promptbox}{prompts/theory\_of\_mind/world\_model\_update.txt}
\begin{verbatim}
You are {agent_name}, a chef in an Overcooked kitchen. You keep a structured MENTAL
  MODEL of
  your
teammate {partner_name} — your best estimate of what they are doing, what they intend
  next,
  and what they
need from you — and you revise it as you watch them and exchange messages.

Update ONLY the inferred parts of your model of {partner_name}. Stay grounded: rely on
  what
  you can
observe and what was actually said. If something is genuinely unknown, write "Unknown"
  rather than
guessing.

Your CURRENT model of {partner_name}:
{previous_model}

What you can directly observe about {partner_name} right now:
- Position: {observed_position}
- Holding: {observed_holding}

Latest message(s) from {partner_name}: {latest_communication}

Your perspective-taking analysis of {partner_name}: {perspective_analysis}

Your free-text beliefs about {partner_name}: {partner_beliefs}

Your own current sub-task: {my_current_task}

Revise the inferred fields and respond ONLY with this exact JSON:
{{
  "partner_task": "the sub-task {partner_name} appears to be working on right now (one
    short
    phrase)",
  "intention": "what {partner_name} is most likely trying to do next and why, as you
    infer
    it",
  "needs": "the single most useful thing {partner_name} needs from you to make
    progress (a
    handoff, more
  space, a specific ingredient, a free pot, ...), or 'nothing' if they seem
    self-sufficient",
  "interaction_beliefs": "up to 3 concise coordination facts agreed or implied by your
    messages that
  affect how the two of you should divide the work; empty string if none",
  "partner_beliefs": "your INFERENCE of what {partner_name} themselves believes about
    THEIR
    teammates
  (their own view of the others, possibly including you) — you CANNOT read their mind,
    so
    infer this only
  from what they said and did; 'Unknown' if there is no evidence"
}}
\end{verbatim}
\end{promptbox}

\paragraph{Causal forward (world) model.}
\begin{promptbox}{prompts/world\_model/world\_model\_system.txt}
\begin{verbatim}
You are a forward model for the Overcooked cooperative cooking game played on a 2D grid.
Given a description of the current game state and an action taken by one agent,
  predict the
  resulting
game state after that action executes.

Reason CAUSALLY: treat the action as an intervention do(action). Predict only its direct
  downstream
effects; hold every variable that has no causal pathway from the action invariant, and
  consider what
would counterfactually differ had the action not been taken.

GAME RULES:
- Agents move on a grid: right/down/left/up moves by one cell if the target cell is
  empty.
- interact: picks up an item in front of the agent if holding nothing; drops the held
  item
  onto the
object in front; collects soup from a ready pot if holding a plate; delivers soup at the
  serving station
if holding a plate with soup.
- stay: no change to position or inventory.
- Pots cook when they contain exactly 3 ingredients; once cooking starts the pot
  counts down
  and
eventually becomes "soup ready".
- Agents can carry only one item at a time.
- Agents cannot move into walls, counters, pots, or other fixed objects.

OUTPUT FORMAT:
- Return ONLY a single JSON object (no prose, no markdown fences) describing the
  COMPLETE
  predicted next
state, using exactly this schema:
{
  "agent": {"row": <int>, "col": <int>, "facing": "<up|down|left|right>", "holding":
    "<item
    or
  'nothing'>"},
  "teammates": [{"row": <int>, "col": <int>, "holding": "<item or 'nothing'>"}],
  "pots": [{"row": <int>, "col": <int>, "status": "<e.g. 'empty', '2/3 ingredients',
    'cooking', 'soup
  ready'>"}],
  "ingredient_piles": [[<row>, <col>], ...],
  "plate_piles": [[<row>, <col>], ...],
  "serving_stations": [[<row>, <col>], ...],
  "order": "<the current recipe / order>",
  "change": "<one short phrase: what changed, or 'no effect'>"
}
- COPY every field that does not change (piles, plate piles, serving stations, the
  order,
  and any
unaffected teammate) straight from the current state — preserve the full layout, do
  not drop
  or invent
positions.
- Apply ONLY the effect of the given action. If it has no effect (e.g. moving into a
  wall or
  interacting
with nothing), keep the agent unchanged and set "change" to "no effect".
\end{verbatim}
\end{promptbox}
\begin{promptbox}{prompts/world\_model/causal\_world\_model\_predict.txt}
\begin{verbatim}
Current state of {agent_id}:
{obs_text}

Treat the action below as a causal INTERVENTION — do({action_text}). Reason explicitly
  about:
- which state variables this action DIRECTLY CAUSES to change (the downstream effects),
- which variables remain INVARIANT (no causal pathway from this action — carry them over
  unchanged),
- what would counterfactually differ had this action NOT been taken.

Action (intervention) taken by {agent_id}: {action_text}

Predicted next state of {agent_id} as a single JSON object — copy the invariant fields
  from
  the
current state unchanged and apply ONLY the causal effects of this intervention:
\end{verbatim}
\end{promptbox}
\begin{promptbox}{prompts/world\_model/causal\_world\_model\_predict\_tom.txt}
\begin{verbatim}
Current observation of {agent_id}:
{obs_text}

Theory-of-Mind context for predicting the JOINT outcome (not just {agent_id}'s own
  action):
- Current sub-task: {task_beliefs}
- Structured mental model of teammate(s) (their state, inferred intentions and needs):
{partner_model}

Treat the action below as a causal INTERVENTION — do({action_text}). Using the
  observation
  AND the
Theory-of-Mind context above, reason about:
- which state variables this action DIRECTLY CAUSES to change (its downstream effects),
- how the teammate(s) are likely to act given their inferred intentions, and how that
  shapes
  the joint
next state,
- which variables remain INVARIANT (no causal pathway — copy them over unchanged),
- what would counterfactually differ had this action NOT been taken.

Action (intervention) taken by {agent_id}: {action_text}

Return the predicted next state of {agent_id} as a single JSON object in the schema
  above —
copy invariant fields unchanged and apply ONLY the causal effects of this intervention
  (anticipating
the teammate's likely concurrent move where relevant).
\end{verbatim}
\end{promptbox}

\paragraph{Critic.}
\begin{promptbox}{prompts/critic/critic\_system.txt}
\begin{verbatim}
You are an evaluator for an Overcooked kitchen agent. You judge whether the agent's LAST
  action
moved its current SUBTASK forward — purely from how the world CHANGED, never from the
  task's
wording, the agent's position/facing alone, or the environment reward.

{environment_description}

WHAT YOU ARE GIVEN
  The STATE BEFORE the last action and the STATE AFTER it (holding + observation for
    each),
    the
  subtask, and the action taken. Judge by comparing BEFORE against AFTER.

HOW TO JUDGE (one general method — works for any subtask, including ones you have not
  seen):
  1. Infer the subtask's GOAL CONDITION: the concrete change to the world that would
    mean
    "done"
     (an inventory change, a pot / counter / station change, reaching a cell —
       whatever the
     subtask is actually about).
  2. Compare BEFORE -> AFTER and classify the action as exactly one of:
     - "success": the goal condition is now met. Exceeding it also counts (e.g. the
       pot is
       already
       cooking when the task was "add an ingredient").
     - "failure": the action was wasted or wrong. Above all, an `interact` that left
       BOTH
       the
       holding AND the faced cell UNCHANGED had NO EFFECT — the agent attempted
         something
       impossible or was mis-positioned. Also a failure: moving into a wall, moving away
         from the
       needed target, or ending up holding the wrong item.
     - "progress": not done, but a legitimate step toward the goal — e.g. moved closer
       to,
       or
       turned to face, the needed pile / pot / counter / station. Nothing was wasted.
  Rule of thumb: would you tell the agent "done" (success), "good, keep going"
    (progress),
    or
  "that did nothing / that was wrong — do something different" (failure)?

KEY CHECKS
  - The strongest failure signal is a NO-OP interact: holding is identical BEFORE and
    AFTER
    and the
    faced cell is unchanged. The impossible actions in the RULES above (handing to a
      teammate,
    dropping an item back on a pile/station) always leave the state unchanged ->
      failure.
  - A movement action never by itself completes a subtask whose goal is an interaction
    outcome; at
    best it is progress (success only if the subtask's goal was literally to reach /
      face a
      cell).
  - Never declare success from position or facing alone — only from an actual
    holding/world
    change.

CRITIQUE
  - About the LAST action only — not a multi-step plan.
  - One sentence on why the goal condition was not met (or what the no-op was), then one
    sentence
    naming the single most useful next action to fix it.
  - If outcome is "success", critique must be an empty string.

Respond ONLY in this JSON format:
{{
  "reasoning": "state the subtask's goal condition, then what changed from BEFORE to
    AFTER
    (holding,
  pot/counter/station, position), then whether the goal condition is now met",
  "outcome": "success" | "progress" | "failure",
  "success": true if outcome is "success" else false,
  "critique": "if not success: one sentence on what went wrong + one sentence corrective
    action; if
  success: empty string"
}}
\end{verbatim}
\end{promptbox}
\begin{promptbox}{prompts/critic/critic\_user.txt}
\begin{verbatim}
Task: {task}
Last action: {last_action}

STATE BEFORE the last action:
  Holding: {prev_holding}
  Observation: {prev_obs_text}

STATE AFTER the last action (this is the current state):
  Holding: {holding}
  Observation: {obs_text}

Context: {context}
Teammate communication: {communication}
Error: {error}
\end{verbatim}
\end{promptbox}

\paragraph{Auto-curriculum.}
\begin{promptbox}{prompts/curriculum/curriculum\_system.txt}
\begin{verbatim}
You decide the next immediate SUBTASK for an Overcooked kitchen agent. The agent's
  goal is
  to
deliver as many soups as possible before the episode ends.

{environment_description}

YOUR JOB
  From the current observation (what the agent holds, the pot's status, the visible
    piles,
  counters, plate pile and serving station, and the teammate) and the result of the
    agent's
    LAST
  subtask, name the single most useful SUBTASK to work on right now. Output exactly ONE
    subtask.

HOW TO CHOOSE
  - Work out for yourself, from the observation and the rules, where the agent
    currently is
    on the
    way to a delivered soup, and pick the immediate next thing that moves it forward.
      There
      is no
    fixed script — derive the ordering from the recipe (the "Order" line) and the
      current
      state.
  - Repetition is EXPECTED. Delivering many soups means doing the same kinds of subtasks
    over and
    over. Never avoid a subtask just because it was done or attempted before — propose
      whatever the
    current situation needs, even if identical to an earlier one.
  - If the LAST subtask SUCCEEDED, move on to the next thing the situation calls for.
  - If the LAST subtask did NOT succeed, keep the SAME goal and propose ANOTHER WAY to
    achieve it,
    guided by the critique (e.g. reposition to face the correct cell, then interact;
      or set
      a held
    item on an empty counter to free the hands). A failure does not mean the subtask
      is too
      hard —
    it only means that approach didn't work, so try a different approach to the same
      goal.
  - Phrase the subtask as a single concrete instruction naming what to do and, when
    useful,
    where —
    for example "pick up ingredient_0", "add ingredient_0 to the pot", or "place the
      plate
      on the
    counter at (2,0)". Do not bundle multiple steps into one subtask.
  - If the teammate is clearly handling something, choose a subtask that complements it
    rather than
    duplicating it.

Respond ONLY in the following JSON format:
{{
  "reasoning": "where the agent is in the flow (from holding + pot status + what is
    nearby)
    and why this
  is the right next subtask; if the last subtask failed, say what different approach
    this
    is",
  "task": "the single next subtask as one concrete phrase"
}}
\end{verbatim}
\end{promptbox}
\begin{promptbox}{prompts/curriculum/curriculum\_info.txt}
\begin{verbatim}
{question_answers}

Last sub-task: {last_task}
Last action: {last_action}
Previous thoughts: {last_thoughts}
Last sub-task succeeded: {success}
Critique: {critique}
Currently holding: {holding}

Teammate's last message: {communications}
\end{verbatim}
\end{promptbox}
\begin{promptbox}{prompts/curriculum/curriculum\_questions.txt}
\begin{verbatim}
You are a helpful assistant that generates diagnostic questions to help an Overcooked
  agent
  decide its
next sub-task.
The agent's goal is to deliver as many soups as possible.

Teammate's last message: {communications}
Current observation: {obs_text}

Generate 5 to 8 questions about the current kitchen situation. Each question should
  probe a
  concept
relevant to the cooking pipeline.
Questions must be self-contained (no implicit references to "the image" or "my current
  room").
Focus ONLY on topics relevant to the soup delivery goal.

Good question examples:
- "What should the agent do after placing the third ingredient in the pot?"
- "How can the agent tell when the soup is ready to be collected?"
- "What step comes before delivering soup to the serving station?"
- "How can two agents divide the cooking pipeline to avoid blocking each other?"

Bad question examples (too vague or irrelevant):
- "What is the best action?" (too general)
- "What do you see in the image?" (requires visual context we don't have)

Respond ONLY in the following JSON format, where "questions" is a list of strings
  alternating question
and concept:
{{
  "reasoning": "why these questions are relevant to the current situation",
  "questions": [
    "Question 1: ...",
    "Concept 1: ...",
    "Question 2: ...",
    "Concept 2: ...",
    "Question 3: ...",
    "Concept 3: ...",
    "Question 4: ...",
    "Concept 4: ...",
    "Question 5: ...",
    "Concept 5: ..."
  ]
}}
\end{verbatim}
\end{promptbox}
\begin{promptbox}{prompts/curriculum/curriculum\_answer.txt}
\begin{verbatim}
You are a helpful assistant answering questions about the Overcooked cooperative cooking
  game.
Answer concisely based on your knowledge of the game and the provided context (if
  relevant).
Always give the simplest answer when multiple are possible.

Question: {question}
Relevant past context: {relevant_past_context}

Respond ONLY in the following JSON format:
{{"answer": "..."}}
\end{verbatim}
\end{promptbox}

\paragraph{Skill manager.}
\begin{promptbox}{prompts/skill\_manager/skill\_description.txt}
\begin{verbatim}
You are a helpful assistant that names and describes a successful action taken by an
  Overcooked kitchen
agent.
You will be given the action the agent took and their reasoning at the time.

Rules:
1. Do not repeat the exact action name — describe the functional outcome instead.
2. Keep the description to 1-2 sentences maximum.
3. The name should be a short verb phrase (e.g., "loading the pot", "collecting soup").

Respond ONLY in the following JSON format:
{{"name": "skill name", "description": "what this action achieves in the kitchen"}}

Agent's thoughts: {agent_thoughts}
Action taken: {action}
\end{verbatim}
\end{promptbox}
\begin{promptbox}{prompts/skill\_manager/skill\_query.txt}
\begin{verbatim}
You are helping an Overcooked kitchen agent retrieve relevant past skills.
Given the agent's current sub-task and observation, write a short query (1-2
  sentences) that
  describes
what kind of skill would be most useful right now.
The query will be used to search a library of past successful actions.

Current sub-task: {task}
Current observation summary: {obs_text}

Respond ONLY in the following JSON format:
{{"name": "query label", "description": "description of the skill being searched for"}}
\end{verbatim}
\end{promptbox}

\paragraph{Memory.}
\begin{promptbox}{prompts/episodic\_memory/episode\_summary.txt}
\begin{verbatim}
You are a helpful assistant summarizing past cooking episodes for an Overcooked agent.
Below are records of past sub-task attempts. Each record includes the task, what
  happened,
  and whether it
succeeded.
Create a concise summary that highlights:
- Which sub-tasks succeeded and how
- Which sub-tasks failed and why
- Any useful patterns or lessons the agent should remember

Keep the summary under 5 sentences. Focus on information that is actionable for future
  steps.

Respond ONLY in the following JSON format:
{{"summary": "..."}}

Episodes:
{episodes}
\end{verbatim}
\end{promptbox}
\begin{promptbox}{prompts/episodic\_memory/semantic\_extraction.txt}
\begin{verbatim}
You are extracting general domain knowledge from an episode in the Overcooked kitchen
  game.

The goal is to identify GENERALIZABLE FACTS that apply broadly to the domain,
NOT episode-specific observations.

Task: {task}
Episode outcome: {success}
Episode description:
{episode_text}

Extract precisely 3-5 SHORT facts that are:
- General rules or properties (not specific to this episode)
- Unrelated to the agent's specific actions or performance
- Domain mechanics or object properties
- Observable from the environment

Examples of GOOD facts:
- "Onions require interaction to be collected"
- "Soup takes 3 ingredients to cook"
- "Plates are used to serve cooked soup"
- "Ingredients must be placed in pots to cook"
- "The delivery counter requires interaction to serve soup"
- "Moving to locations takes one action per step"

Examples of BAD facts (too specific):
- "Agent went right then interacted" (specific action)
- "Success happened after 5 steps" (episode-specific)
- "Agent 0 prefers left corridor" (agent-specific, not domain)

Return a JSON response with the structure:
{{
  "facts": "Fact 1\nFact 2\nFact 3"
}}

Each fact on a separate line. Keep each fact to one sentence (under 20 words).
\end{verbatim}
\end{promptbox}

\paragraph{PCM strategic and tactical layers (OverForge).}
\noindent The strategic layer is called with the system message \emph{``You are a strategic planner for cooperative games.''}; the tactical layer with \emph{``You translate a high-level plan into concrete next moves.''}, or, in the \texttt{no\_hierarchy} arm, \emph{``You choose concrete next moves directly, with no predefined plan.''}.
\begin{promptbox}{prompts/strategic\_plan.txt}
\begin{verbatim}
You are choosing your HIGH-LEVEL strategy in a cooperative Overcooked kitchen.
World model (your observation, beliefs, task, and structured teammate models):
{state}

Relevant memory (past skills / episodes / facts that worked before):
{memory}

Propose {n_plans} DISTINCT high-level plans — roles, intentions, or coordination
  strategies. Think about the division of labour with your teammate (who does what),
  which sub-goal to pursue, and how to avoid getting in each other's way. Examples:
  - 'be the pot-loader: keep the left pot filled while my partner plates'
  - 'take over plating and delivery of the ready soup'
  - 'fetch ingredients and stage them on the counter for my partner'
Each plan is a short phrase describing intent. Do NOT include any concrete
moves or action names (no up/down/left/right/interact/stay) — the concrete
actions are decided later.

Return ONLY valid JSON (no markdown):
{{"plans": ["plan 1", "plan 2", "plan 3"]}}
\end{verbatim}
\end{promptbox}
\begin{promptbox}{prompts/tactical\_actor.txt}
\begin{verbatim}
You are executing a high-level plan in a cooperative Overcooked kitchen.
Your current plan (role/intention): {plan}

World model (this may be the real state OR a predicted/simulated future state — your
  observation, beliefs, task, and structured teammate models):
{state}

Relevant memory (past skills / episodes / facts):
{memory}

Propose 3-4 CONCRETE next-move candidates that best advance this plan from here. Each
  MUST be exactly one of: right, down, left, up, stay, interact. No other words.

Return ONLY valid JSON (no markdown): {{"actions": ["interact", "up"]}}
\end{verbatim}
\end{promptbox}
\begin{promptbox}{prompts/tactical\_actor\_flat.txt  (no\_hierarchy arm)}
\begin{verbatim}
You are choosing the next move in a cooperative Overcooked kitchen (no predefined plan
  or role).

World model (this may be the real state OR a predicted/simulated future state — your
  observation, beliefs, task, and structured teammate models):
{state}

Relevant memory (past skills / episodes / facts):
{memory}

Propose 3-4 CONCRETE next-move candidates that best progress toward completing and
  delivering the recipe from here. Each MUST be exactly one of: right, down, left, up,
  stay, interact. No other words.

Return ONLY valid JSON (no markdown): {{"actions": ["interact", "up"]}}
\end{verbatim}
\end{promptbox}

\paragraph{PCM value judges (OverForge).}
\noindent System messages: pairwise judge \emph{``You are an expert action evaluator in cooperative games. Consider Theory of Mind: what beliefs would change if each action succeeds?''}; absolute judge \emph{``You are an expert action evaluator in cooperative games. Score the single action on its own merits.''}; state-value judge \emph{``You are a concise state evaluator for a cooperative game. Reply with only a JSON object of the form \{"value": $<$number between 0 and 1$>$\}.''}.
\begin{promptbox}{prompts/pairwise\_judge.txt}
\begin{verbatim}
You compare two candidate actions for one chef in a cooperative Overcooked kitchen and
say which better serves the team. There is NO reward and NO score to maximise — judge by
how well each action advances the shared work and the partnership.

ACTIONS TO COMPARE
Action A: {action_a}
Action B: {action_b}

CURRENT SITUATION (the real or imagined state)
{context}

BELIEFS (self / task / partner / interaction)
{beliefs}

Judge each action on four facets, each 0.0–1.0, then combine them with EQUAL weight
(simple average — there are no tunable weights):

1. COHERENCE — does the action actually do what it intends from THIS state (right
   position/facing/inventory for it to take effect), rather than a no-op or a fumble?
2. GOAL-DIRECTEDNESS — does it move the team toward a delivered soup: gather → pot →
   cook → plate → serve? Closer to a delivered dish = higher.
3. EPISTEMIC VALUE — when the right move is unclear, does it reveal useful information
   or reduce uncertainty (e.g. uncover state, resolve who-does-what)? When the right
     move
   is obvious, this facet matters little.
4. SOCIAL FIT — is the action legible to the partner and does it avoid blocking them or
   colliding in the shared space? Complementary, non-interfering moves score higher.

Then:
  score_A = mean(coherence_A, goal_A, epistemic_A, social_A)
  score_B = mean(coherence_B, goal_B, epistemic_B, social_B)
  better      = "A" if score_A >= score_B else "B"
  confidence  = abs(score_A - score_B)        (small gap = low confidence)

Keep the reasoning to 1–3 sentences. Do NOT invent objects or positions not present in
the situation above; if an action's effect is genuinely uncertain, reflect that with
mid-range facet scores and low confidence.

Return ONLY valid JSON (no markdown):
{{
  "better": "A" or "B",
  "confidence": 0.0,
  "action_a_score": 0.0,
  "action_b_score": 0.0,
  "reasoning": "one to three sentences"
}}
\end{verbatim}
\end{promptbox}
\begin{promptbox}{prompts/absolute\_judge.txt}
\begin{verbatim}
You rate ONE candidate action for a chef in a cooperative Overcooked kitchen, on its own
merits. There is NO reward and NO score to maximise — judge how well the action advances
the shared work and the partnership.

ACTION
{action}

CURRENT SITUATION (the real or imagined state)
{context}

BELIEFS (self / task / partner / interaction)
{beliefs}

REFERENCE RUBRIC (read-only: it is NOT the answer format and must not be copied)
The JSON below is the exact prompt of the judge that normally compares TWO candidate
actions (A vs B) for this chef, with its input placeholders left as-is. Use exactly the
same facets, scale, weighting and scoring rule it defines.

{pairwise_judge_json}

YOUR TASK
Compute EXACTLY the same score that rubric would assign — the same four facets, each
0.0-1.0, combined by simple average — but for ONLY the one ACTION above. Judge it
  alone: do
NOT compare it to any other action, and do not assume an alternative is on the table.
  There
is no Action B, so there is no "better" and no "confidence" to report.

  score = mean(coherence, goal_directedness, epistemic_value, social_fit)

Decide the four facet scores BEFORE you write; do not deliberate inside the JSON. Do NOT
invent objects or positions not present in the situation above; if the action's effect
  is
genuinely uncertain, use mid-range facet scores.

Answer with ONE JSON object and nothing else. Its keys, in this order: "coherence",
"goal_directedness", "epistemic_value", "social_fit" (each a number in [0, 1] for THIS
action), "score" (their mean), and "reasoning" (ONE sentence, at most 30 words,
  justifying
the scores). Fill every number with your actual judgement of this action.
\end{verbatim}
\end{promptbox}

\smallskip\noindent The \texttt{\{pairwise\_judge\_json\}} slot is filled at run time with the
pairwise judge prompt above, serialised as the following JSON object (so the two judges can never
drift apart):
\begin{promptbox}{(run-time JSON)}
\begin{verbatim}
{
  "judge": "pairwise",
  "placeholders": {
    "{action_a}": "candidate action A",
    "{action_b}": "candidate action B",
    "{context}": "the current situation (the real or imagined state)",
    "{beliefs}": "the agent's beliefs (self / task / partner / interaction)"
  },
  "prompt": "<the full text of prompts/pairwise_judge.txt, JSON-escaped, with its
    {action_a}/{action_b}/{context}/{beliefs} placeholders left as-is>"
}
\end{verbatim}
\end{promptbox}

\smallskip\noindent\textit{Response template.} The prompt deliberately contains no fillable answer
template (the model copied one back verbatim). Instead the reply is constrained at decoding time
to this JSON schema (vLLM structured outputs, \texttt{response\_format: json\_schema}); the four facet
scores are the same facets as the pairwise judge, and the code recomputes \texttt{score} as their
equal-weight mean:
\begin{promptbox}{(run-time JSON)}
\begin{verbatim}
{
  "type": "object",
  "properties": {
    "coherence": {
      "type": "number",
      "minimum": 0,
      "maximum": 1
    },
    "goal_directedness": {
      "type": "number",
      "minimum": 0,
      "maximum": 1
    },
    "epistemic_value": {
      "type": "number",
      "minimum": 0,
      "maximum": 1
    },
    "social_fit": {
      "type": "number",
      "minimum": 0,
      "maximum": 1
    },
    "score": {
      "type": "number",
      "minimum": 0,
      "maximum": 1
    },
    "reasoning": {
      "type": "string",
      "maxLength": 300
    }
  },
  "required": [
    "coherence",
    "goal_directedness",
    "epistemic_value",
    "social_fit",
    "score",
    "reasoning"
  ],
  "additionalProperties": false
}
\end{verbatim}
\end{promptbox}

\begin{promptbox}{agent\_overforge.py  (StateValueJudge.\_prompt, the trajectory value judge nu)}
\begin{verbatim}
You are judging a single (possibly predicted/simulated) state of a cooperative
  Overcooked kitchen.

State:
{state_text}

Beliefs:
{beliefs}

Rate how PROMISING this state is for the two-chef team to finish and deliver soups
  together, on a 0.0–1.0 scale, where:
  1.0 = excellent (a soup is delivered / clearly about to be), 
  0.5 = neutral / ordinary progress, 
  0.0 = bad (stuck, agents blocking each other, no path to a soup).

Return ONLY valid JSON (no markdown): {{"value": 0.0}}
\end{verbatim}
\end{promptbox}

\paragraph{Fixed-strategy probe (RQ2) prompt variants.}
\noindent The taxonomy arm replaces the two layer prompts above with the following files on both seats; the pinned arm bypasses the strategic call on one seat and uses the fixed sentence as its plan.
\begin{promptbox}{prompts/rq2/strategic\_taxonomy.txt}
\begin{verbatim}
You are choosing your HIGH-LEVEL STRATEGY in a cooperative Overcooked kitchen.

World model (your observation, beliefs, task, and structured teammate models):
{state}

Relevant memory (past skills / episodes / facts that worked before):
{memory}

WHAT A STRATEGY IS, AND WHAT IT IS NOT
A STRATEGY is a standing role or division of labour. It says WHO DOES WHAT and WHY, it
holds for many steps, and it stays true if the room, the recipe or your teammate
  changes.
It never names a square, a coordinate, or a key press.
A TACTIC is the opposite: the concrete move you make right now to serve the strategy. It
is short-lived and depends entirely on where things happen to be. Tactics are chosen
  later,
by a different part of you. Do not put any tactic in a strategy.

  strategy : "be the supplier: keep ingredients flowing to whoever is cooking"
  tactic   : "step left, then interact to pick up the ingredient"   <- NOT a strategy

  strategy : "be the finisher: plate and serve whatever is ready"
  tactic   : "face the pot and interact with a plate in hand"       <- NOT a strategy

FAMILIES OF STRATEGY AVAILABLE IN THIS KITCHEN
  1. SUPPLIER      keep ingredients arriving wherever they are needed
  2. COOK          own the pot: load it, start it, watch it
  3. FINISHER      plate what is ready and take it to be served
  4. SPLIT-BY-ITEM divide the order by ingredient type, one type each
  5. RELAY         never cross the room: hand work to your teammate at the boundary
  6. YIELDER       take the supporting role: keep lanes clear so your teammate flows

You are NOT required to use these words or these six. If the situation calls for a
different standing role, propose it. Choose what genuinely fits this state and this
teammate, not what is listed first.

Propose {n_plans} DISTINCT strategies. Each is a short phrase describing a standing
  role.
No coordinates, no move names (no up/down/left/right/interact/stay), no step-by-step.

Return ONLY the JSON object below. Do not add a "reasoning" key, an explanation,
or any other field.

Return ONLY valid JSON (no markdown):
{{"plans": ["strategy 1", "strategy 2"]}}
\end{verbatim}
\end{promptbox}
\begin{promptbox}{prompts/rq2/tactical\_taxonomy.txt}
\begin{verbatim}
You are choosing the TACTIC that serves your standing strategy right now, in a
cooperative Overcooked kitchen.

Your STRATEGY (a standing role; it does not change from step to step):
{plan}

World model (this may be the real state OR a predicted/simulated future state — your
observation, beliefs, task, and structured teammate models):
{state}

Relevant memory (past skills / episodes / facts):
{memory}

WHAT A TACTIC IS
A TACTIC is the concrete next move that advances your standing strategy FROM WHERE YOU
ARE NOW. The same strategy demands different tactics in different rooms, at different
moments, and beside different teammates. Serving the strategy is the only test of a
good tactic.

RECURRING TACTICAL PATTERNS IN THIS KITCHEN
  - APPROACH-AND-FACE : move so the thing you want is directly in front of you before
                        interacting; interact only works on the cell you face
  - RE-FACE           : if an interact did nothing, you were probably facing the wrong
                        cell; turn before trying again
  - STAGE             : put an item on a counter for your teammate instead of carrying
                        it the whole way yourself
  - YIELD             : step out of a shared lane so your teammate can pass
  - HOLD              : stay put when moving would block, or when what you need is not
                        ready yet
  - CHECK-THE-ORDER   : the order can change after a delivery; fetch what the CURRENT
                        order needs, not what the last one did

These are patterns, not a menu: the move you output must be one of the six primitives.

Propose 3-4 CONCRETE next-move candidates that best advance your strategy from here.
Each MUST be exactly one of: right, down, left, up, stay, interact. No other words.

Return ONLY the JSON object below. Do not add a "reasoning" or "thought" key,
an explanation, or any other field.

Return ONLY valid JSON (no markdown): {{"actions": ["interact", "up"]}}
\end{verbatim}
\end{promptbox}
\begin{promptbox}{prompts/rq2/g0\_pot\_loader.txt  (the pinned strategy g0)}
\begin{verbatim}
be the pot-loader: keep a pot filled with what the order needs, and leave plating and
  serving to your teammate
\end{verbatim}
\end{promptbox}

\end{document}